\documentclass[sn-mathphys]{sn-jnl}

\usepackage{cleveref}

\usepackage{acronym}
\usepackage{siunitx}
\usepackage{acronym}
\usepackage{listings}
\usepackage{graphicx}
\usepackage{wrapfig}
\usepackage{booktabs}
\usepackage{caption}
\usepackage{subcaption}
\usepackage{indentfirst}
\usepackage{tabularx}
\usepackage{adjustbox}

\newacro{CNN}[CNN]{Convolutional Neural Network}
\newcommand{\cnn}{\ac{CNN} }

\newacro{RNN}[RNN]{Recurrent Neural Network}
\newcommand{\rnn}{\ac{RNN} }

\newacro{MLP}[MLP]{Multi Layer Perceptron}
\newcommand{\mlp}{\ac{MLP} }

\acrodef{DL}[DL]{Deep Learning}   
\newcommand{\DL}{\ac{DL} }

\acrodef{RL}[RL]{Reinforcement Learning}   
\newcommand{\RL}{\ac{RL} }

\newacro{RCNN}[RCNN]{Region$-$based Convolutional Neural Network}    
\newcommand{\rcnn}{\ac{RCNN} }

\newacro{SSD}[SSD]{Single$-$Shot Detector}
\newcommand{\ssd}{\ac{SSD} }

\acrodef{SORT}[SORT]{Simple Online and Realtime Tracking}   
\newcommand{\SORT}{\ac{SORT} }

\acrodef{SOT}[SOT]{Single Object Tracking}
\newcommand{\SoT}{\ac{SOT} }

\acrodef{MOT}[MOT]{Multi Object Tracking} 
\newcommand{\MoT}{\ac{MOT} }

\acrodef{RGB}[RGB]{Red Green Blue}

\acrodef{SCT}[SCT]{Single Camera Tracking}
\newcommand{\SCT}{\ac{SCT} }

\acrodef{SCTs}[SCTs]{Single Camera Trackings}

\acrodef{MCTs}[MCTs]{Multi Camera Tracks}
\newcommand{\MCT}{\ac{MCTs} }

\acrodef{MTMCT}[MTMCT]{Multi$-$Target Multi$-$Camera Tracking}   
\newcommand{\MTMCT}{\ac{MTMCT} }

\acrodef{MT}[MT]{Mostly Tracked}

\acrodef{FN}[FN]{False Negatives}   
\newcommand{\FN}{\ac{FN} }

\acrodef{FP}[FP]{False Positives}

\acrodef{MOTA}[MOTA]{Multi$-$Object Tracking Accuracy}   
\newcommand{\MOTA}{\ac{MOTA} }

\acrodef{MOTP}[MOTP]{Multi$-$Object Tracking Precision}

\newacro{LSTM}[LSTM]{Long Short Term Memory}

\acrodef{IoT}{Internet of Things}

\acrodef{KF}[KF]{Kalman Filter}   
\newcommand{\KF}{\ac{KF} }

\makeatletter
\newcommand\inlinecomment{\begingroup\@makeother\#\@inlinecomment}
\newcommand\@inlinecomment[2]{
    \todo[%
        inline,
        color=#1,
    ]{#1:  #2}
    \endgroup
}
\newcommand\comment{\begingroup\@makeother\#\@comment}
\newcommand\@comment[2]{
    \todo[
        color=#1,
    ]{#1:  #2}
    \endgroup
}
\makeatother

\jyear{2023}

\theoremstyle{thmstyleone}

\theoremstyle{thmstyletwo}

\theoremstyle{thmstylethree}

\begin{document}

\title[Semi-Automated Computer Vision based Tracking]{Semi-Automated Computer Vision based Tracking of Multiple Industrial Entities -- A Framework and Dataset Creation Approach}

\author*{\fnm{Jérôme} \sur{Rutinowski*}}\email{jerome.rutinowski@tu-dortmund.de}
\equalcont{These authors contributed equally to this work.}

\author{\fnm{Hazem} \sur{Youssef}}
\equalcont{These authors contributed equally to this work.}

\author{\fnm{Sven} \sur{Franke}}

\author{\fnm{Irfan Fachrudin} \sur{Priyanta}}

\author{\fnm{Frederik} \sur{Polachowski}}

\author{\fnm{Moritz} \sur{Roidl}}

\author{\fnm{Christopher} \sur{Reining}}

\affil{\orgdiv{Chair of Material Handling and Warehousing}, \orgname{TU Dortmund University}, \country{Germany}}

\abstract{This contribution presents the TOMIE framework (Tracking Of Multiple Industrial Entities), a framework for the continuous tracking of industrial entities (e.g., pallets, crates, barrels) over a network of, in this example, six RGB cameras.
This framework, makes use of multiple sensors, data pipelines and data annotation procedures, and is described in detail in this contribution.
With the vision of a fully automated tracking system for industrial entities in mind, it enables researchers to efficiently capture high quality data in an industrial setting.

Using this framework, an image dataset, the TOMIE dataset, is created, which at the same time is used to gauge the framework's validity.
This dataset contains annotation files for 112,860 frames and 640,936 entity instances that are captured from a set of six cameras that perceive a large indoor space.
This dataset out-scales comparable datasets by a factor of four and is made up of scenarios, drawn from industrial applications from the sector of warehousing.
Three tracking algorithms, namely ByteTrack, Bot-Sort and SiamMOT are applied to this dataset, serving as a proof-of-concept and providing tracking results that are comparable to the state of the art.}

\keywords{Warehousing, Computer Vision, Object Detection, Classification}

\maketitle
\section{Introduction}\label{sec1}

The continuous, real-time tracking of entities of interest plays a crucial role in industrial settings from production facilities to warehouses \cite{franko_reliable_2020}.
In light of future challenges, automated, vision-based tracking of industrial entities helps increase process transparency \cite{anuj2017multiple}. 
The application potentials for the industry are manifold. 
With emerging needs in digitization and automation, industrial entities need to be continuously tracked in real time to increase the adaptability of logistics systems with different conditions in terms of layouts, conveyors, etc.
The available, however still unused information of these entities could be leveraged to automate and efficiently design the subsequent interfaces in the process. 

\begin{figure}[htbp]
    \centering  
        \includegraphics[width=0.8\linewidth]{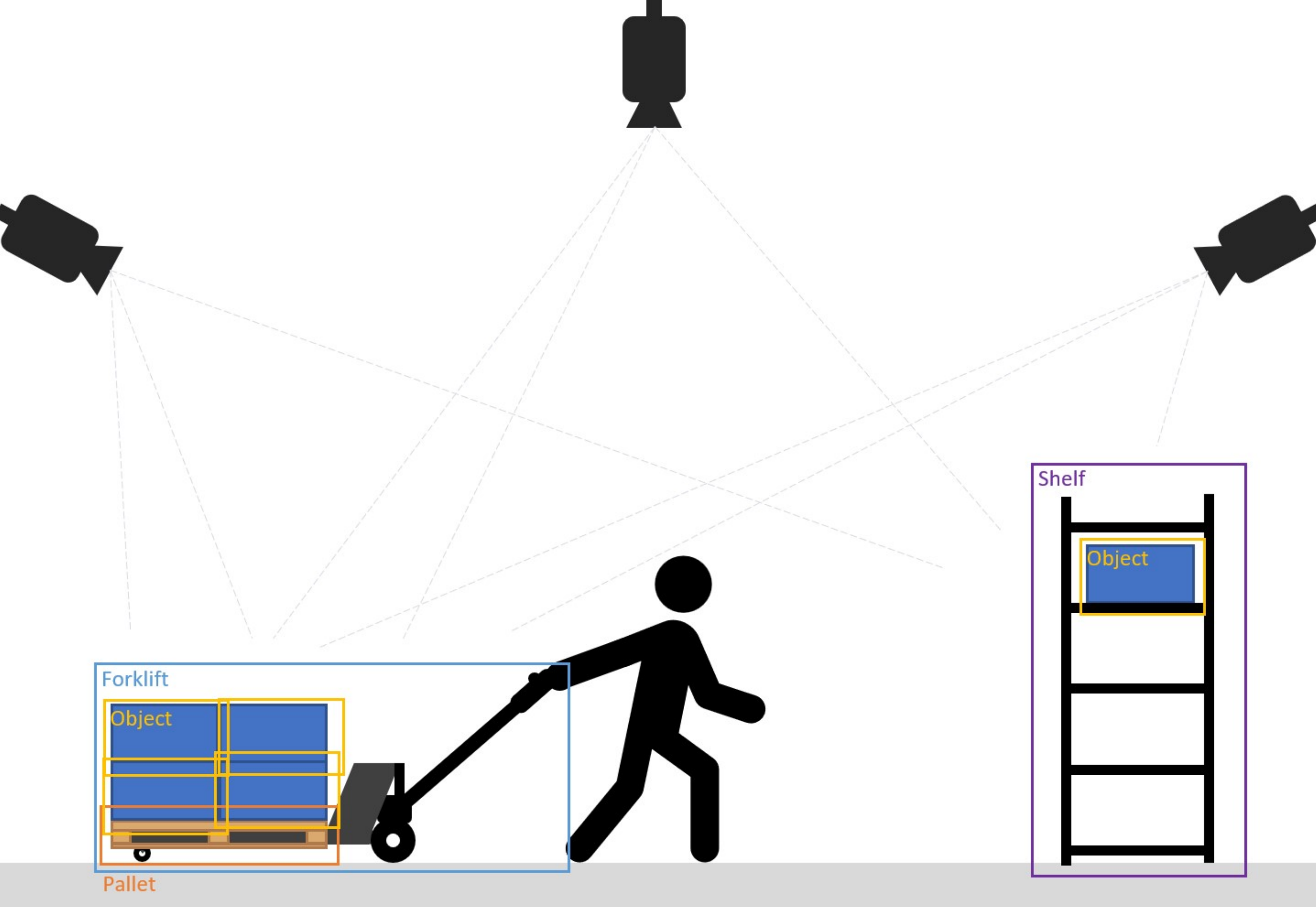}
        \caption{Visualization depicting the RGB camera based tracking of multiple industrial entities in a warehousing environment.}
        \label{fig:vision}
\end{figure}

The adoption of object tracking in the industry would facilitate the creation of a future-proof, scalable, and flexible infrastructure for monitoring processes. 
Process steps that rely on manual object identification or scanning equipment could then be eliminated and replaced by comparatively inexpensive cameras that function in an environment-agnostic manner. 
Given these requirements, our vision of a fully automated tracking of multiple entities in the industry can be articulated as follows: 
In an industrial environment, such as a warehouse, all entities should be continuously tracked, classified and identified in real time.
As a consequence, their location, 6D pose and identity are known at all times.
This remains the case when multiple entities are present at once, might be in motion and might occlude one another.
The sensors used for this purpose are comparatively inexpensive, do not need to meet a narrow set of criteria, do not need to be mounted in a very specific manner, and are easily obtainable.
An example for one such sensor could be an RGB camera with a standard lens and resolution.
The information that is inferred from the sensor data is used to monitor and optimize and to increase the transparency of existing processes (e.g., in the form of a  digital twin).
Thanks to some of this information, novel processes might emerge.
A visualization of this vision, put into practice in a warehousing scenario, might look like, can be seen in  Fig. \ref{fig:vision}. 

Besides the task of object tracking \cite{anuj2017multiple}, research in the field of tracking concerning humans has also been performed \cite{liu_recent_2023,zhan_ray3d_2022,wang_deep_2021}.
We define the difference between (industrial) entities and (human) subjects in the sense that objects have simple and predictable movement patterns with only a brief and limited motion profile, if any. 
On the other hand, subjects, like humans, possess dynamic structures that are prone to self-occlusion, along with unpredictable and unrepeatable movement patterns. 
In our work, we only refer to object tracking, hence we take only industrial entities into account. 

\clearpage

\subsection{Problem Statement}
As to put the herein described vision of a fully automated tracking system into practice, the following challenges have to be addressed:
Realistic scenarios, demonstrating the movement of industrial entities throughout a common industrial environment, have to be chosen and planned.
For this purpose, a viable data foundation, in the sense of entities that are commonly used in industrial settings, moved in a way in which they would be moved in the latter, has to be established.
Out of these scenarios, a dataset has to be created.
This dataset needs to contain annotated recordings, that can be used as trustworthy, ground truth training data for a computer vision algorithm.
A set of such algorithms has to be selected and applied to the recorded data, and subsequently be compared to one another based on pre-defined evaluation metrics.
Describing all these challenges, however, reveals the challenge that is at the core of this undertaking -- the lack of a recording framework, that enables researchers to efficiently record and (semi-)automatically annotate data.
 
\subsection{Goal of the Contribution}

The goals of this contribution are the following:
We aim to provide a framework for the continuous tracking of industrial entities over a network of cameras.
The provision of such a framework for the research community is motivated by the increase in efficiency and reduction of laborious annotation work entailed by it.
We will describe the process of creating this framework in detail.
Further, we aim to create a dataset with high quality ground truth data, that can be used as a benchmark for subsequent research.
This dataset will comprise multiple scenarios, that we will establish and describe in this contribution and that closely resemble industrial scenarios.
We subsequently aim to apply a set of algorithms to the dataset, as to provide a proof-of-concept for our framework.

\subsection{Structure and Methodological Approach}
The next sections are structured as follows: Section \ref{sec2} will outline and contextualize the related work on computer vision of tracking entities. This is followed by an explanation of the conducted experiments and used methodology in section \ref{sec3}. Section \ref{sec4} shows the corresponding results. Finally, in section \ref{sec5}, the results are summarized, discussed, and an outlook is given on what further research in tracking industrial entities can look like. All in all, we want to provide a transparent approach on how state-of-the-art object tracking can be used as a benchmark for others. Our framework realises tracking with a concrete approach that is also applicable for the industry and is a key element for practical application.

\section{Related Work}\label{sec2}
Computer vision based tracking is a research field that has gained attention in recent years. 
The rapid developments of this field of study lead to the emergence of numerous multi-object tracking algorithms and frameworks as well as datasets.
Therefore, this chapter briefly presents the relevant literature related to camera based object tracking techniques and frameworks, existing computer vision datasets, and methods of dataset creation. We also discuss existing tracking approaches in different application domains.

\subsection{Camera based Object Tracking}
The points of interest for camera based object tracking are the detection of particular entities and the estimation of their movement trajectories while maintaining a distinct identification for each item within the camera view.
In the current state of the art, \MoT is one of the computer vision based tracking concepts that has been widely implemented in diverse application fields.
The tendency of using \MoT can be seen for various algorithms and benchmarks and can be applied to single camera or multi camera systems.
However, the deployment of \MoT applications considers some real-world challenges, i.e., occlusion of  entities over long time periods and the task of re-identification after the occlusion.    
Therefore, in this subsection, we review both systems to provide insights into their respective drawbacks and advantages.  

\subsubsection{Single Camera Systems}
The single camera system is a fundamental system architecture for the development of tracking algorithms.
Ciaparrone et al. \cite{Ciaparrone_2019} conducted a survey emphasizing the usage of \DL in \MoT for 2D data using the \SCT technique. The survey specifies that most \MoT algorithms that are developed to be used with a single camera have four steps/stages in common: detection, feature extraction / motion prediction, affinity, and association. The implied aim is to implement \DL at every stage and to evaluate the given algorithms as a whole on a \emph{MOTChallenge} dataset \cite{dendorfer2021motchallenge}. The datasets mostly consist of benchmarks for pedestrian tracking. Deep learning is mostly used for the first two stages, while only a few contributions implement \DL approaches for affinity and association. 

From this survey \cite{Ciaparrone_2019}, the authors emphasize three important parameters to deploy \MoT algorithms: (i) the detection quality, (ii) \cnn for feature extraction, and (iii) \SoT trackers.
In terms of detection quality, appropriate detectors must be thoroughly selected to reduce the number of \FN in the \MOTA score. Currently, the best performing \DL based detector is Faster \rcnn from \cite{yu_poi_2016}. In contrast, \ssd performs worse, as presented in \cite{Hilke_2018, Zhao2018MultiObjectTW}. However, \ssd was almost able to work in real-time \mbox{($4.5$ FPS)}, including the detection step. 

For the feature extraction stage \cite{Ciaparrone_2019}, the best-performing method, GoogLeNet \cite{GoogLeNet_2015}, is applied to the datasets of MOT15 \cite{MOT15}, MOT16 and MOT17 \cite{MOT16}. Approaches that do not use appearance (whether they are deep or conventional methods) typically perform worse. Visual features alone, however, are insufficient to compute affinity; many of the better-performing algorithms additionally include other characteristics, particularly motion features. The integration of \SoT to the private \MoT detectors along with \DL is considered to generate well-performing online trackers.

Authors of \cite{Tang2017MultiplePT, Chen2017OnlineMT, Ma2018CustomizedMT} have investigated a \DL approach for affinity using the MOT16 \cite{MOT16} dataset. Both works of \cite{Ma2018CustomizedMT, Tang2017MultiplePT} demonstrate the reliable similarity measures to support person re-identification after occlusions and are able to reach the highest \MOTA score of $49.3$$\%$. 
The survey also mentions that few have used \DL to enhance the association process from the classical association, like the Hungarian algorithm, such as \rnn \cite{RNN_assoc_2018}, deep \mlp \cite{Hilke_2018}, and \RL \cite{Ren_2018}. 
However, the usage of \DL as to directly guide the association algorithm and to perform tracking is still at its starting stage. 

The \SORT \cite{SORT_2016} algorithm is regarded as the foundation for the online and real-time application of MOT. This approach implements \KF as the basic prediction of the tracklet bounding box between frames and the constant-velocity model as the motion model. One of the  limitations of \SORT is that it accumulates error estimation of the entity position over time due to obstacles or non-linear motion.
To overcome this issue, the BoT-SORT tracker \cite{Aharon22} was developed by combining the benefits of camera-motion correction, motion and appearance information, and a more precise Kalman filter state vector.
In addition, this tracker provides a novel, straightforward, and compelling technique of Intersection over Union (IoU) and re-identification through cosine-distance fusion, in order to obtain stronger correlations between detections and tracklets.
The authors \cite{Aharon22} further integrate BoT-SORT into the novel Byte-Track \cite{zhang_bytetrack_2022}, which uses the backbone of the high-performance detector YOLOX.
Both BoT-SORT and Byte-Tracker are evaluated using the datasets from the MOT17 and MOT20 challenges.
The trackers outperform all current trackers in the MOTChallenge, with the results from MOT17 test set, which are $80.2$ IDF1 (the ratio of correct detections to the average number of ground truth and calculated detections), $65.0$ HOTA (Higher Order Tracking Accuracy), and $80.5$ MOTA. 

\subsubsection{Multi Camera Systems}

As shown in the aforementioned survey results, \SCT shows a promising solution to handle \MoT tasks. Nevertheless, \SCT covers a finite view of a single camera which leads to inadequate detection, tracking, and re-identification robustness,  due to occlusions over longer time spans \cite{Yoon2015BayesianMT, Tiwari2017ARO, Analysis_DL_2021}. To resolve the occlusion problem, a \MTMCT approach is proposed by several contributions \cite{Bredereck_2012, Wang2013IntelligentMV, zhang2015camera, An_Occlussion_2021}. \MTMCT defines the combination of the different perspectives from multiple networked cameras to detect and track entities. Common \MTMCT pipelines begin with the \SCT step based on \emph{tracking-by-detection} from each camera \cite{Analysis_DL_2021, Wang2013IntelligentMV, zhang2015camera, An_Occlussion_2021}. The tracklets from the detection step are then generated as the input to get \SCT for each entity in each camera. All tracked targets of each single camera or SCT  are furthermore associated by using the camera clustering approach \cite{Improving_Multi_Cam_2022, An_Occlussion_2021, liu2021city}. The final output of \MTMCT is Multi Camera Tracks (MCTs) in a high dimensional space which are obtained by the clustering step.

Zhang et al. \cite{zhang2015camera} introduce a challenging benchmark for \MoT on pedestrians that is comprised of two main modules: intra- and inter-camera tracking. Their dataset is recorded from non-overlapping video recordings from six to eight cameras with a resolution of $640$~×~$480$. Intra-camera tracking generates tracklets for each individual camera that utilize the \SCT algorithm. SCT's output is then forwarded to the inter-camera module where the data association takes place in the \MTMCT system. 
Tracking Length (TL), Crossing fragments (XFrag), and Crossing ID-switches (XIDS) are three possible evaluation metrics. For scenarios two to six, TL results (percentage of the correctly tracked object) are varying from $70\%$ to $80\%$, XFrag results (number of times for a linked pair of tracks) are ranging from $29$ to $42$ links, XIDS results demonstrate from $23$ to $44$ tracks that lack a link to the ground truth trajectories.

A survey about intelligent multi-camera video surveillance is carried out by the authors of \cite{Wang2013IntelligentMV}. Their work introduces key technologies: multi-camera calibration, computation of camera networks topology, multi-camera tracking, object re-identification, and multi-camera activity analysis. The survey looks at ways of estimating 3D camera calibration, including intrinsic and extrinsic parameters, common ground plane, automatic calibration, and two cameras with substantial overlap.  The survey also emphasizes the topology of multi-camera networks which explains the handover of objects and computation of the topology. There are multiple methods for topology computation, including; correspondence-based, correspondence-free, and topology inferred by non-overlapping camera networks. The section goes fairly in-depth into the ideas of inter-camera tracking based on multi-camera calibration, inter-camera tracking with appearance
cues, and solving correspondence views across multiple cameras.   

Specker et al. \cite{An_Occlussion_2021} define an occlusion-aware \MTMCT approach for vehicle tracking and re-identification that enhances both \SCT and \MCT operation. Furthermore, the authors adopt the global
feature learning model from \cite{he2020multi} to handle vehicle re-identification. To improve the resulting accuracy, a multiple re-identification network is applied. 
The \SCT setup introduces an occlusion handling strategy and additional modules for filtering faulty detections. These steps can be achieved by using temporal information from tracks. 
The \MCT setup uses a novel pipeline that includes a scene model, filtering of tracks, re-identification distance calculation, and hierarchical clustering. 
The hierarchical cross-camera clustering based on vehicle re-identification features is adapted from works of \cite{hsu2019multi, kohl2020mta} to merge the multi-camera tracks by leveraging topological and temporal constraints of the tracks of each camera in the network. 
The authors \cite{An_Occlussion_2021} propose that in order to decrease the negative influence of overlapping vehicles, one should improve re-identification by
excluding boxes in the background or with occlusion.

\subsection{Computer Vison Datasets}
Successful deployment of DL-based computer vision applications relies on relevant and high quality datasets \cite{mayershofer2020loco}.
Nowadays, datasets are aimed to encompass diverse and specific use cases and current trends tend to be dominated by outdoor applications, i.e., \emph{MOTChallenge} (MOT15 \cite{MOT15}, MOT16, MOT17 \cite{MOT16}, MOT20 \cite{MOT20}), KITTI \cite{geiger2013vision}, MS COCO \cite{lin_microsoft_2014} (Common Objects in Context).
The \emph{MOTChallenge} dataset is a popular framework containing a large collection of multiple people-tracking datasets in dense pedestrian scenarios and the evaluation benchmark for various tracker algorithms.

The \emph{MOTChallenge} uses different metrics to evaluate the performance of \MoT methods.
Standard evaluation metrics include multi object tracking accuracy (MOTA) \cite{Bernardin08}, higher order tracking accuracy (HOTA) \cite{Luiten20}, Identity $F_1$ Score (IDF1) \cite{Ristani16}, and Identity switches (IDs) \cite{zhang_bytetrack_2022}.
Metrics differ in their consideration of the causes of errors.
The IDs metric counts the number of swapped object identities during tracking.
The MOTA metrics combines three sources of errors and is defined as follows:
\begin{equation}
    \text{MOTA} = 1 - \frac{\sum_t (\text{FP}_t + \text{FN}_t + \text{IDs}_t)}{\sum_t \text{GT}_t},
\end{equation}
where $t$ is the current frame and $GT$ is the total number of visible objects \cite{MOT16}.

Alongside the TP, FP, FN, and TP measures, the HOTA metric considers the classification of associations.
Given a TP $c$, the set of True Positiv Associations (TPAs) is the set of TPs with the same ground truth and predicted identities as $c$\cite{Luiten20}.
The HOTA metric with a localization threshold $\alpha$ is defined as:
\begin{align}
    &\text{HOTA}_\alpha = \sqrt{ \frac{ \sum_{c \in \{TP\}} A(c)} {\text{TP} + \text{FN} + \text{FP}} }, \\
    \text{with }&A(c) = \frac{\mid\text{TPA(c)}\mid} { \mid \text{TPA(c)} \mid + \mid\text{FNA(c)}\mid + \mid\text{FPA(c)}\mid}.
\end{align}

The IDF1 Score considers the assignment of objects to their ground truth identities and is defined as:
\begin{equation}
    \text{IDF}_1 = \frac{2 TP_{ID}} {2TP_{ID} + FP_{ID} + FN_{ID}}.
\end{equation}

An autonomous-driving related dataset is demonstrated in KITTI \cite{geiger2013vision}, that specifies various traffic scenarios.
The published dataset contains six hours of video from the cameras and sensor measurements which are captured at \mbox{$10$~-~$100$} Hz readings.
Moreover, MS COCO \cite{lin_microsoft_2014} (Common Objects in Context) datasets contribute to providing daily life scenes with over $80$ object classes and $200,000$ labeled images.
Despite large datasets, MS COCO does not cover industry-related computer vision applications.
MVTec ITODD \cite{drost2017introducing} accommodates realistic industrial setups for 3D object detection and pose estimation.
The dataset consists of $28$ asset classes that are sorted in more than $800$ scenes and labeled using approximately $3,500$ rigid 3D transformations as the ground truth \cite{drost2017introducing}, i.e., engine parts, metal plates, bearings, injection pumps, etc.
Luo et al. \cite{luo2019benchmark} present a benchmark dataset for industrial tools (ITD) to identify  different types of tools at the level of usage.
This dataset is aimed to accurately forecast how a robot would interact with various industry settings.
ITD includes more than $11,000$ hand-labeled RGB images in eight tool categories with $24$ general industrial tools in total as well as their multi-perspective views of every tool.
Regardless of various scenario views, this dataset only focuses on small industrial tools such as safety goggles, wrenches, screw drivers, etc.

Synthetic-based industrial object datasets are, e.g., created in the research work of \cite{de2022dataset, abou2022synthetic}. 
The authors of \cite{de2022dataset} develop both real-world and synthetic data of industrial metal or reflective objects that are arranged as multi-view RGB images with 6D object pose labels.
The real-world objects dataset contains $600$ scenes with $31,200$ RGB images and the synthetic data provides $42,600$ synthetic scenes containing $553,800$ images.
The twin resemblance of synthetic and real-world datasets including a controlled environment facilitates simulation-to-real-world research.
In this manner, computer vision based simulations with scalable scenarios are able to be conducted.
Akar et al. \cite{abou2022synthetic} propose synthetic datasets of industrial objects for object detection applications.
The datasets are generated as $200,000$ photo-realistic generated images with precise bounding box annotations that are categorized as $8$ industrial objects in $32$ scenarios.
The warehouse environment model as well as the datasets are rendered using NVIDIA Omniverse.
The goal of synthetic datasets is to automatically generate datasets for real-world multiple object detectors from genuine camera feeds.

The Logistics Objects in Context (LOCO) \cite{mayershofer2020loco} dataset presents an indoor environment dataset for warehousing logistics. However, the LOCO dataset does not contain timestamps for the recorded image streams which renders it unsuitable for object-tracking algorithms.
This type of logistics or industry related dataset is rare to encounter in research \cite{s20154083, Rutinowski, Rutinowskia}.
The authors \cite{mayershofer2020loco} intend to accelerate computer vision based research for logistics by emphasizing the creation of objects and scenes of warehousing entities and privacy protection of image acquisition.
The LOCO dataset has $39,101$ images comprising $151,428$ annotated logistics entities such as pallets, pallet trucks, and forklifts.

\subsection{Dataset Creation Methods}
The creation of industry related datasets is the topic of this subsection.
Obtaining and marking such datasets in an industrial environment can be difficult due to factors such as it being time-consuming, susceptible to human mistakes, and constrained by various privacy and security regulations \cite{abou2022synthetic,mayershofer2020loco, de2022dataset}.
Therefore, using a semi- or fully-automated pipeline for the dataset creation should be considered.
All setups of the related industrial dataset papers are summarized in Table \ref{:comparison_dataset}.
\begin{table}[b!]
\centering
\caption{Comparison of industrial based datasets creation setups.}
\begin{adjustbox}{width=\textwidth}
\begin{tabular}{lllrl}
    \hline
    Dataset &    Acquisition Tool &    Camera Type &    Resolution [px] &    Evaluation \\
    \hline
    MVTec ITODD \cite{drost2017introducing}&
    \begin{tabular}[c]{@{}l@{}}3 Cameras\\ 3D Camera\end{tabular} &
    \begin{tabular}[c]{@{}l@{}}Grayscale\\ Stereo\end{tabular} & 8 MP &
    \begin{tabular}[c]{@{}l@{}}PP3D\\ PP3D-E\\ PP3D-E-2D\\ S2D\\ RANSAC\end{tabular} \\
    \begin{tabular}[c]{@{}l@{}}Industrial Metal \\ Objects \cite{de2022dataset}\end{tabular} &
    \begin{tabular}[c]{@{}l@{}}JAI GO-5000-PGE\\ mvBlueFOX3\\ RealSense L515\\ RealSense D415\\ Rico Theta S\end{tabular} &
    \begin{tabular}[c]{@{}l@{}}RGB\\ Grayscale\\ RGB, LiDAR\\ RGB, IR Stereo\\ $360^o$ Camera\end{tabular} &
    \begin{tabular}[c]{@{}l@{}}2560 x 2048\\ 4064 x 3044\\ 1920 x 1080\\ 1920 x 1080\end{tabular} & MSSD \\
    ITD \cite{luo2019benchmark}&
    Kinect 2.0 &
    RGBD &
    1024 x 575 &
    \begin{tabular}[c]{@{}l@{}}FR-CNN\\ R-FCN\\ YOLOv3\\ SSD\end{tabular} \\
    LOCO \cite{mayershofer2020loco} &
    \begin{tabular}[c]{@{}l@{}}MS Kinect v2\\ Intel Realsense D435\\ SJCAM SJ-4000MS \\ LifeCam HD-3000\\ Logitech C310\end{tabular} &
    \begin{tabular}[c]{@{}l@{}}RGBD\\ RGBD\\ RGB\\ RGB\\ RGB\end{tabular} &
    \begin{tabular}[c]{@{}r@{}}1920 x 1080\\ 1920 x 1080\\ 1920 x 1080\\ 1280 x 800\\ 1280 x 800\end{tabular} &
    \begin{tabular}[c]{@{}l@{}}YOLOv4608\\ YOLOv4tiny\\ FR-CNN\end{tabular} \\
    \begin{tabular}[c]{@{}l@{}}Synthetic Object \\ Dataset \cite{abou2022synthetic}\end{tabular} &
    NVIDIA Omniverse &
    \begin{tabular}[c]{@{}l@{}}Renderer\\ Software\end{tabular} &
    720 &
    \begin{tabular}[c]{@{}l@{}}FR-CNN\\ SSD\end{tabular}\\
    \hline
\end{tabular}
\end{adjustbox}
\label{:comparison_dataset}
\end{table}
Semi-manual annotation for the 3D images of the industrial objects is adapted in MVTec ITODD \cite{drost2017introducing}.
For each object, three types of scenes are captured: (i) those with only one instance of the object and no extra items, (ii) those with multiple instances of the object and no extra items, and (iii) those with both multiple instances of the object and additional clutter. 
The individual scene is recorded once using a 3D industrial camera, and twice using grayscale cameras: one scene with a randomly projected pattern and another one without a random pattern.
Both grayscale and 3D cameras are located on top of the shelf setup and calibrated previously with regard to their relative position to the object.
The recorded object is positioned on a calibrated turn's movements under the cameras that allow the multiple scenes to be captured automatically.
In this manner, the ground truth of 3D object poses are transferred directly for every rotation.
Instead of using a rounding box as the correctness measure, the authors \cite{drost2017introducing} implement 3D pose based evaluation.
The datasets are evaluated using 3D pose based methods: Shape-Based 3D Matching (S2D), Point-Pair Voting (PP3D), Point-Pair Voting with 3D edges (PP3D-E), Point-Pair Voting with 3D edges and 2D refinement (S2D), and RANSAC.
Although S2D outperforms other methods when estimating the image results, a majority of the results are false positives. PP3D-E performs the prediction well with a top-1 detection rate of $68$\% with the given threshold of $5\%$ but the running time is higher (by $0.1$ s) which must be improved for the industrial use.

The Industrial tool dataset (ITD) \cite{luo2019benchmark} is gathered utilizing a Kinect $2.0$ sensor that can generate $30$ RGBD frames per second, featuring a resolution of $1024$~×~$575$ px, as well as $512$~×~$424$ px depth frames. 
To collect the data, the tools are positioned within a distance range of $1$~m~-~$5$~m from the camera. 
The tools are placed in their typical positions and industrial settings, while the camera is positioned at the same point of view as that of the worker's eyes. 
The worker walks smoothly around the target tool while maintaining a consistent focus on it.
The labeling process is conducted manually by experts. 
Each worker is tasked with identifying the name of the tool, the category it belongs to, and its potential usage. 
The task requires a total of approximately $200$~h to complete.
The performed evaluations demonstrate that cluttered backgrounds and inconsistent ambient lighting impact tool detection. 
Moreover, the performance suffers from the worker's motion-induced visual blur.
To achieve the industrial requirements, the refinement of detection methods is necessary.

The dataset for industrial metal objects, described in \cite{de2022dataset}, is recorded in two parts -- real-world and synthetic data.
An industrial grasping robot, the Fanuc M20ia, is equipped with the data acquisition setup listed in Table \ref{:comparison_dataset} (except the $360^o$ camera) to record multi-view images of various scenes in the real world.
The real-world scene is captured by each camera from $13$ different viewpoints to obtain 6D poses of each object.
Six different metal objects with different lighting setups are also considered during the recording.
In addition, the objects are recorded in three different types of carriers: metal plates, small bins, and cardboard boxes.
The labeling of 6D poses from object models is carried out semi-manually using a proprietary tool.
The synthetic datasets are generated by mimicking real-world scenes, i.e., poses, lighting, models, textures on Unity for which the virtual environment uses a HDRI environment map.
This map is constructed by the captured images from a $360^o$ camera using different types of exposures.
Finally, all real-world and virtual scenes are generated as the dataset containing subfolders for each camera IDs and an individual subfolders corresponding to each CAD model of the respective object.
To evaluate the labeling performance, de Roovere et al. calculate the pose errors using Maximum Symmetry-Aware Surface Distance (MSSD). 

A full synthetic dataset for warehousing environments is rendered in NVIDIA Omniverse based on the Universal Scene Description (USD) method \cite{abou2022synthetic}.
Akar et al. employ Material AI tools to transform the captured images from real-world cameras and material scanners into realistic virtual models.
The scene recording setups are emulated as authentic factory representations that have many assets and instances. 
For each scene recording, the randomized locations and rotations are assigned to the camera in order to capture the scene's randomness from diverse perspectives.
Subsequently, synthetic image generation is initiated to automatically and accurately annotate the images in each scene up to the pixel level.
FRCNN ResNet50 surpasses SSD DL model in terms of detecting stillages, transport robots, dollies, and pallets with the Average Precision (AP) metric at $0.5$ are $69.90$$\%$, $89.93$$\%$, and $48.60$$\%$, respectively.
The recordings of the LOCO \cite{mayershofer2020loco} dataset are captured using different types of cameras with diverse fields of view and resolutions in a real warehousing environment.
The cameras are set up on a mobile unit with a special arm, thus enabling the re-adjustment of the camera's point of view.
The mobile unit moves around the warehouse while changing the cameras' perspectives. The captured images are recorded and stored with a 1 Hz frequency.
The LOCO annotator uses the backbone of the COCO annotator with additional features, such as an automated bounding box tool and new hotkeys.
To ensure the privacy of the warehouse workers in the dataset, Mayershofer et. al. utilize a neural network to automatically perform pixelization of all detected faces during the annotation phase.
The evaluated models exhibit a lower performance compared to the COCO benchmark, with an mAP at $0.5$ $\approx~$20$\% - $40$\%$ on the LOCO benchmark.

\section{Methodology}\label{sec3}
Due to the existing deficiency in the publicly available object tracking datasets in the logistics and industrial domains, we collect a custom dataset and annotate it in a semi-automated fashion. The following section describes our dataset recording procedure, our dataset structure, and the annotation process.
The word \emph{entity} is used in this work to refer to the recorded objects. This excludes commonly used references in the literature such as object pose estimation, object tracking, and object detection.

\subsection{Planning and Execution of the Dataset Recording} \label{sec:scenarios} 
\begin{figure}[ht]
    \centering
    \begin{subfigure}{0.2\linewidth}
        \includegraphics[width=\linewidth]{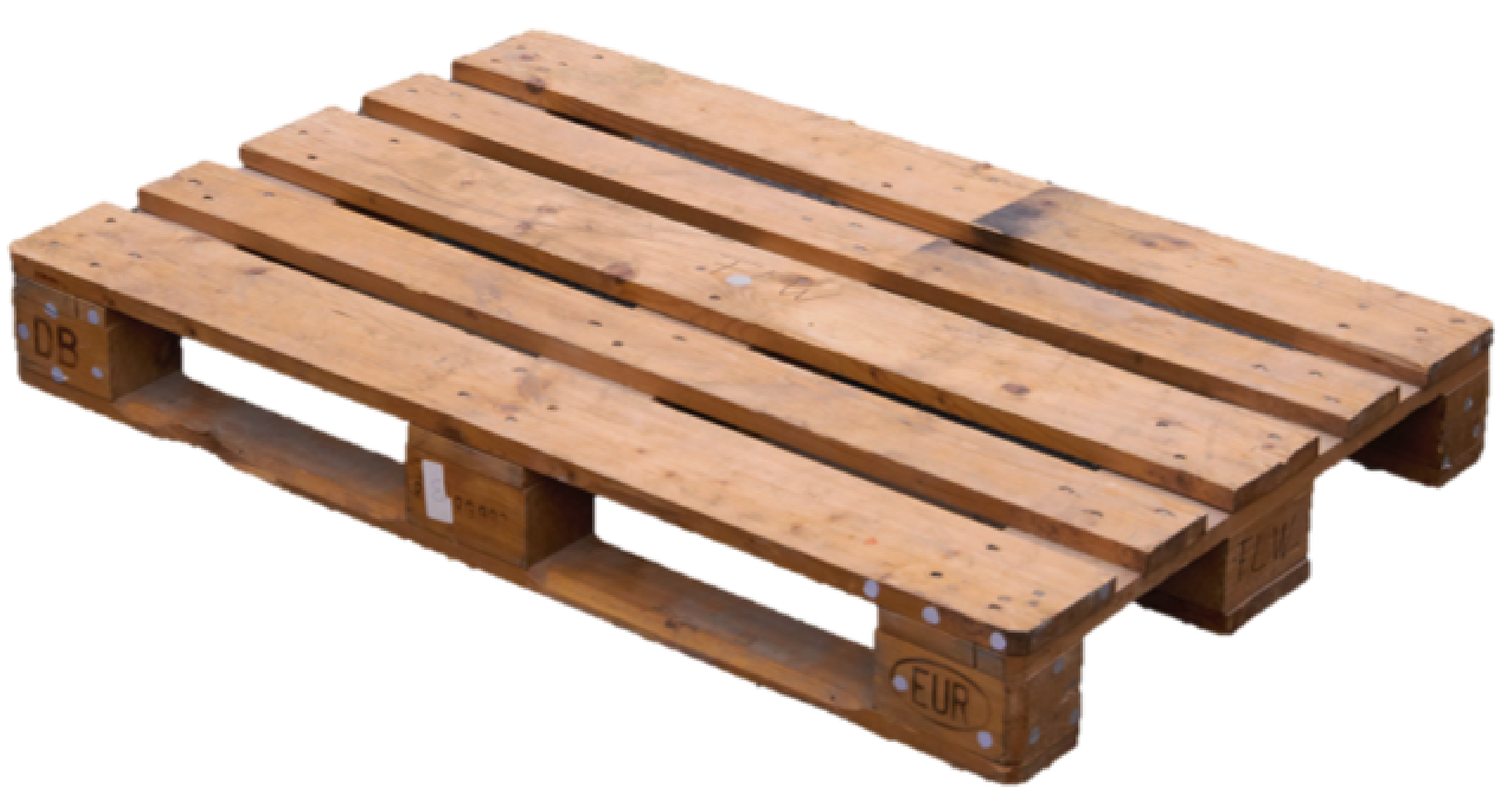}
        \caption{}
    \end{subfigure}
        \begin{subfigure}{0.2\linewidth}
        \includegraphics[width=\linewidth]{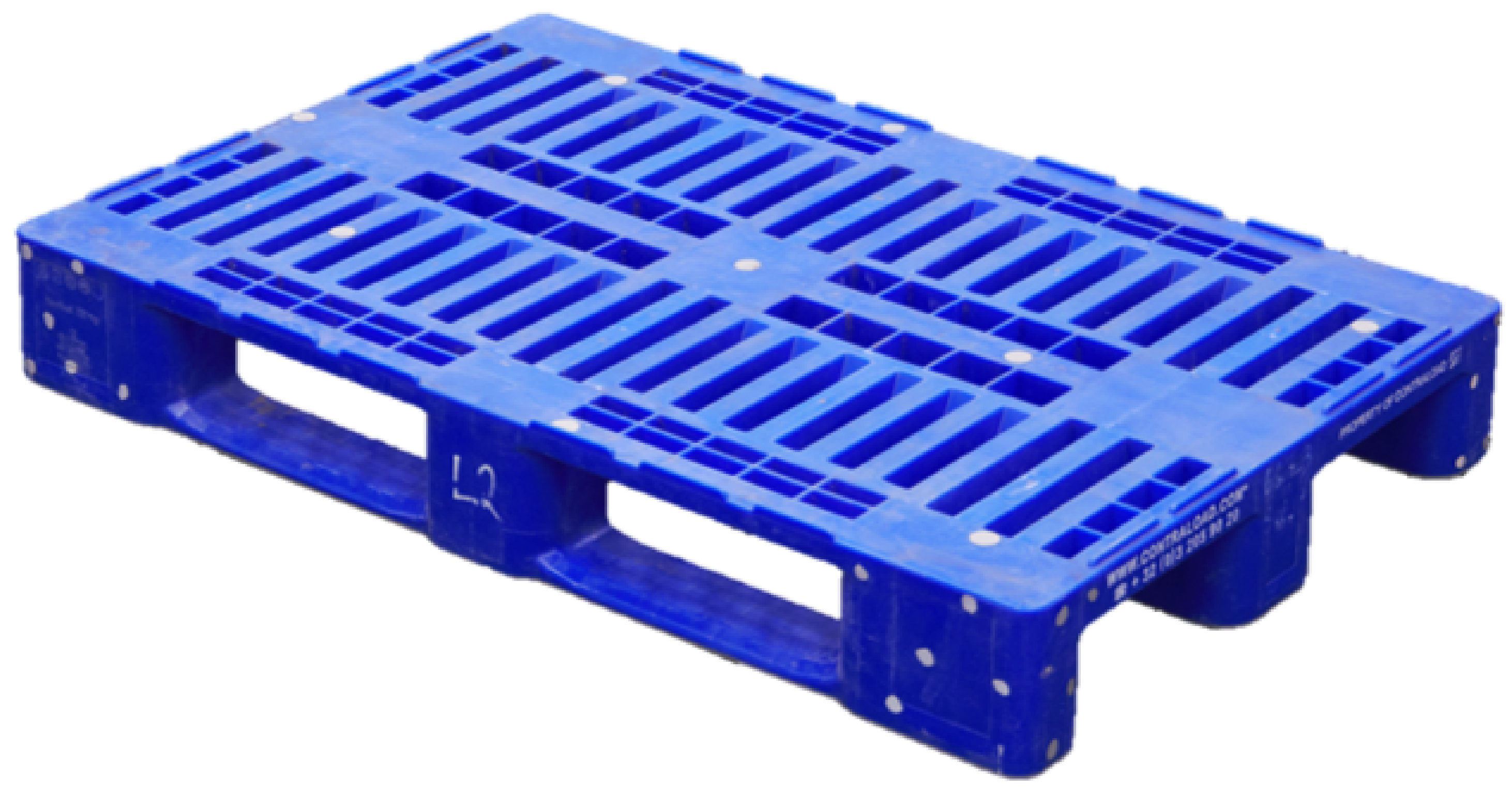}
        \caption{}
    \end{subfigure}
    \begin{subfigure}{0.2\linewidth}
        \includegraphics[width=\linewidth]{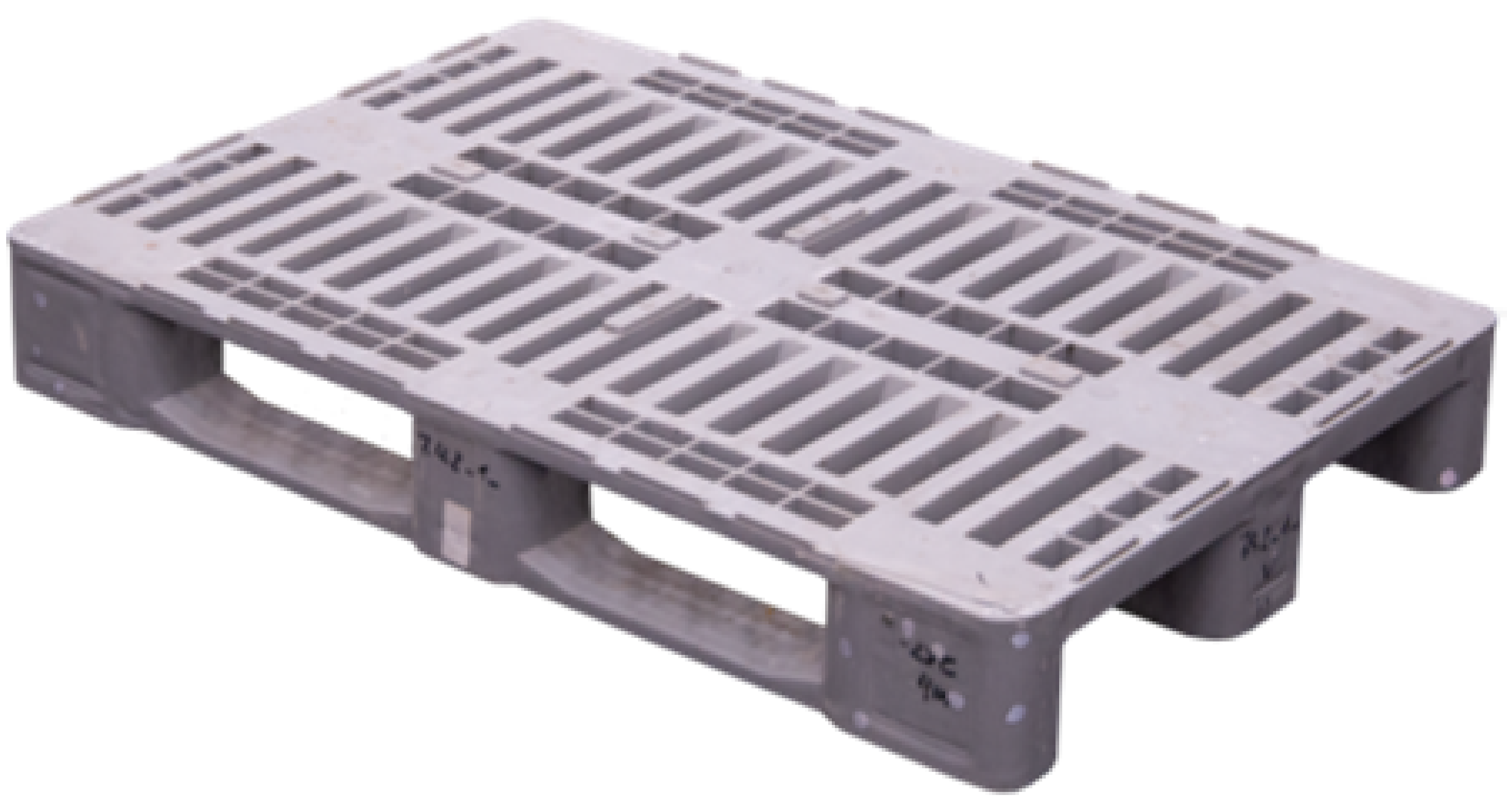}
        \caption{}
    \end{subfigure}
    \begin{subfigure}{0.2\linewidth}
        \includegraphics[width=\linewidth]{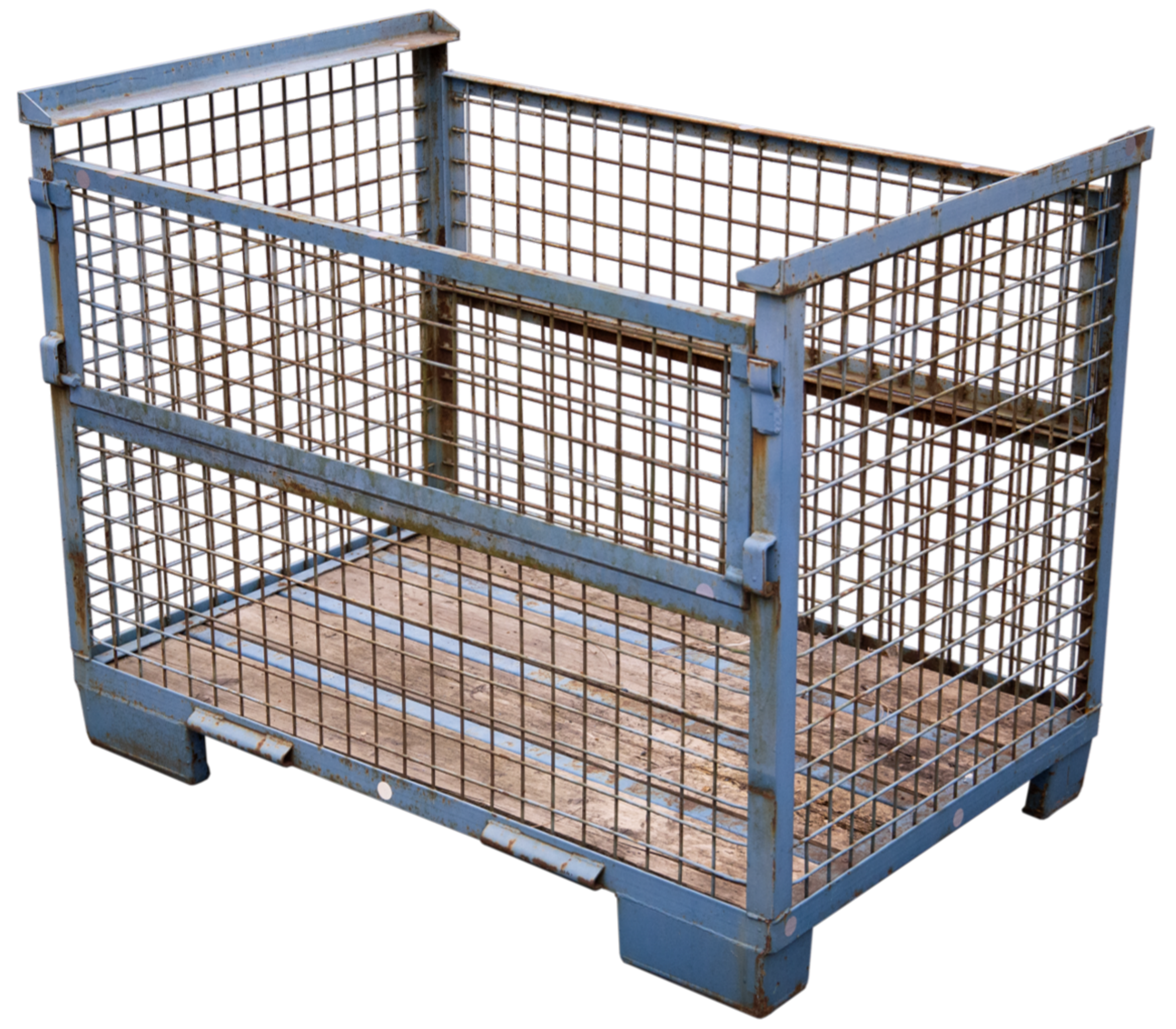}
        \caption{}
    \end{subfigure}
    \begin{subfigure}{0.2\linewidth}
        \includegraphics[width=\linewidth]{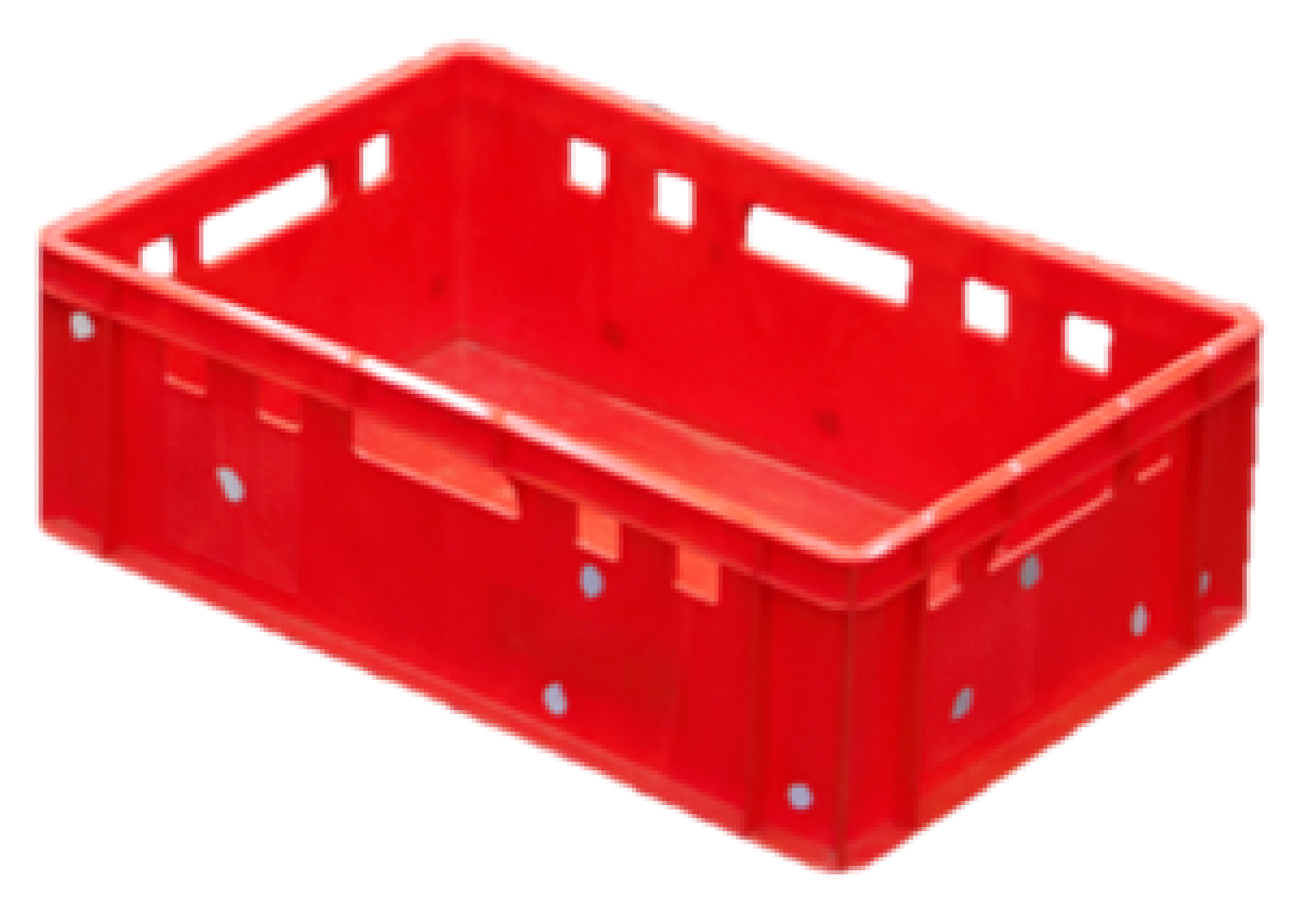}
        \caption{}
    \end{subfigure}
    \begin{subfigure}{0.2\linewidth}
        \includegraphics[width=\linewidth]{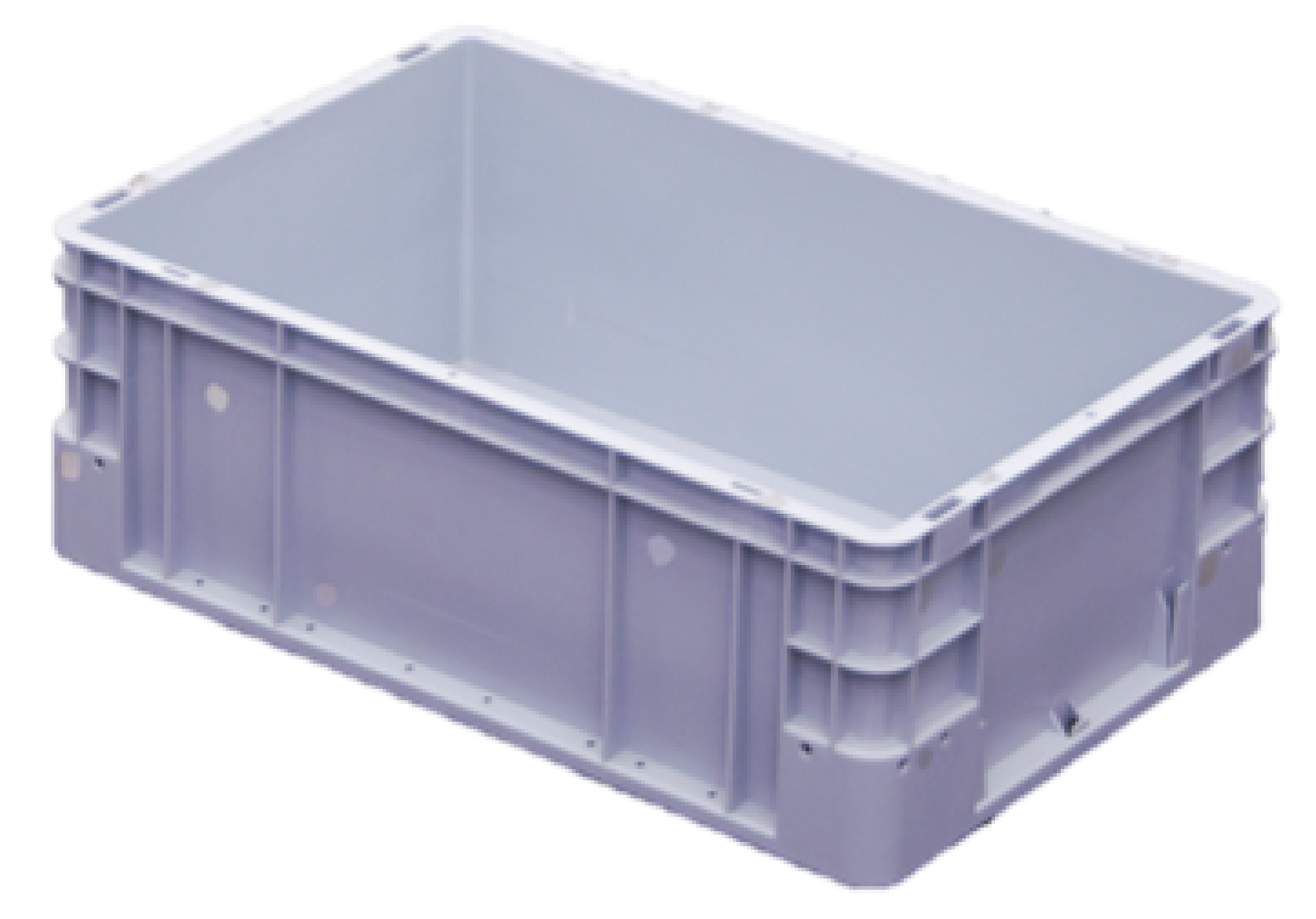}
        \caption{}
    \end{subfigure}
    \begin{subfigure}{0.2\linewidth}
        \includegraphics[width=\linewidth]{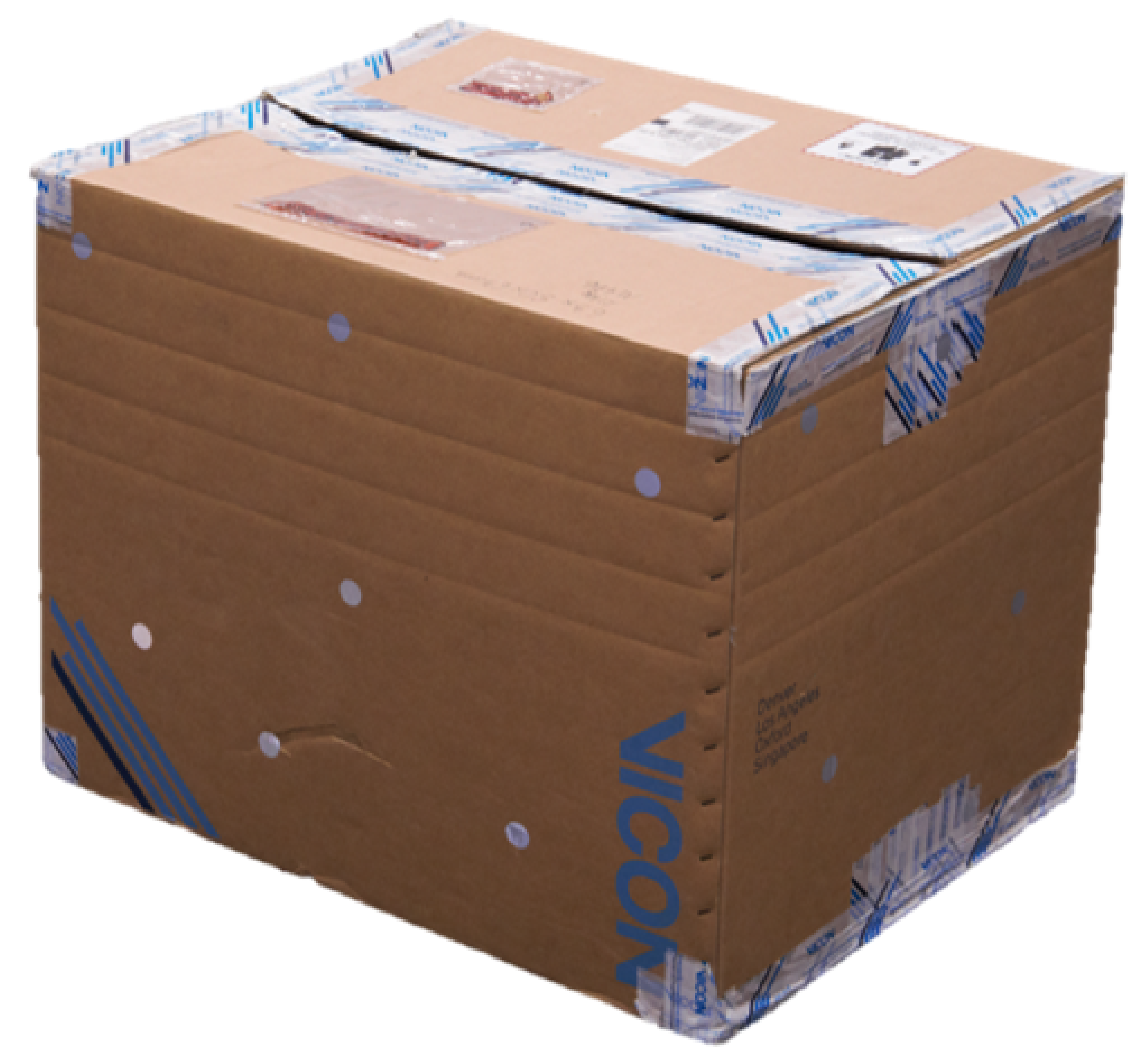}
        \caption{}
    \end{subfigure}
        \begin{subfigure}{0.2\linewidth}
        \includegraphics[width=\linewidth]{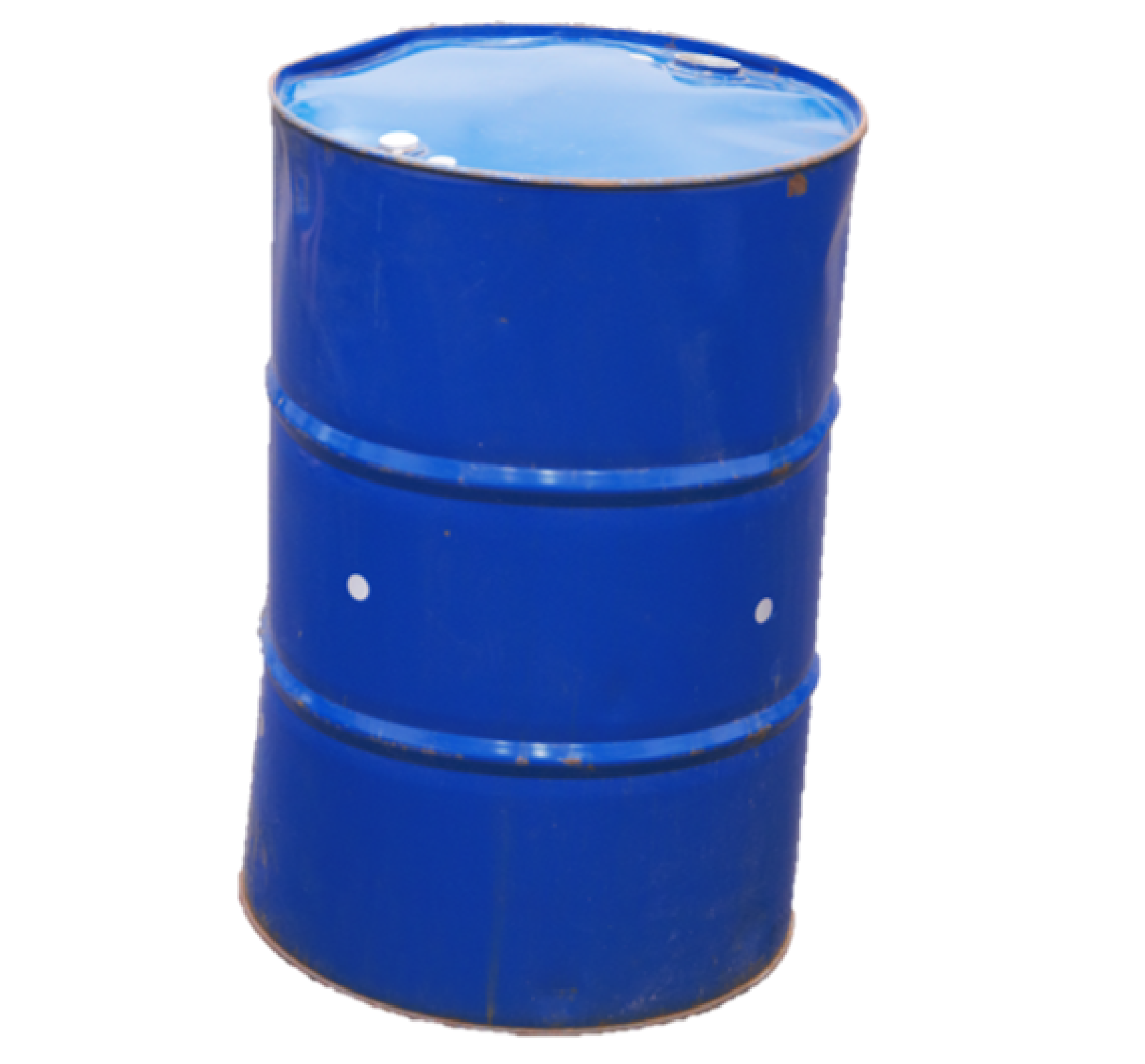}
        \caption{}
    \end{subfigure}
        \begin{subfigure}{0.2\linewidth}
        \includegraphics[width=\linewidth]{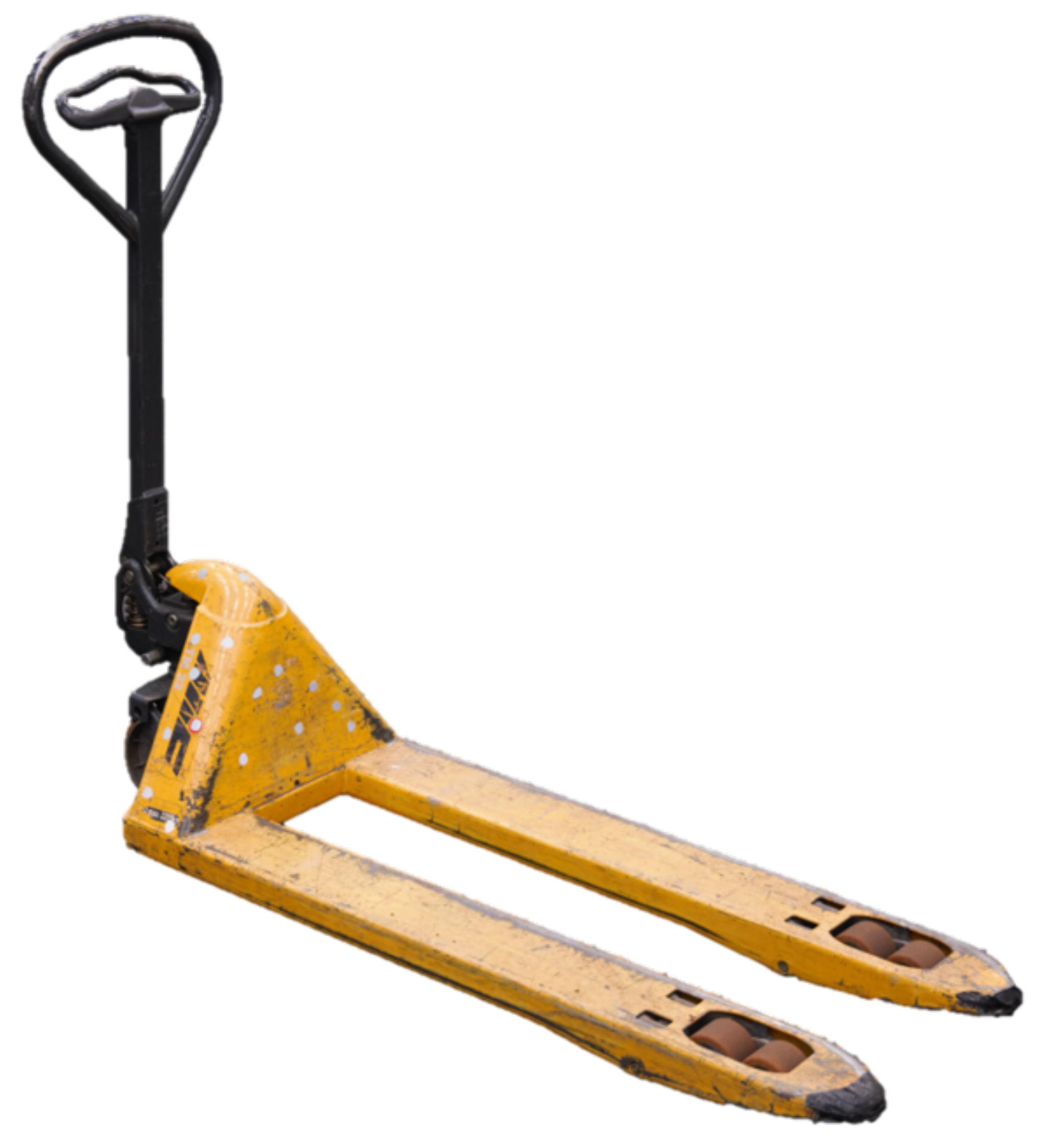}
        \caption{}
    \end{subfigure}
    \caption{Entities used in our recordings: (a) Euro pallet, (b) CHEP pallet, (c) Hygiene pallet, (d) Mesh box, (e) Red small load carrier, (f) Gray small load carrier, (g) Cardboard, (h) Barrel, (i) Forklift}
    \label{fig:entities}
\end{figure}

We derive two situations from the warehousing sector that represent processes occurring in actual industrial use cases, namely a goods reception scenario and a block storage scenario. 
In order to ensure realistic circumstances, two different loading degrees of the pallets were recorded. 
In the first stage, only empty pallets are moved. The second stage involves fully loaded pallets.  
As to ensure a realistic environment, we use six different industrial entities (small load carriers, pallets, barrels, cardboard boxes, forklifts, and a mesh box, as shown in Fig. \ref{fig:entities}). 
Pallets of different types were used, including Euro pallets, CHEP pallets, and hygiene pallets. The entities were handled with two manual pallet trucks. The selection of entities is inspired by DIN $55405$ and DIN EN $13698-1$ \cite{DIN_55405,DIN_EN_13698-1}. 

\begin{figure}[htbp]
    \centering  
        \includegraphics[width=0.8\linewidth]{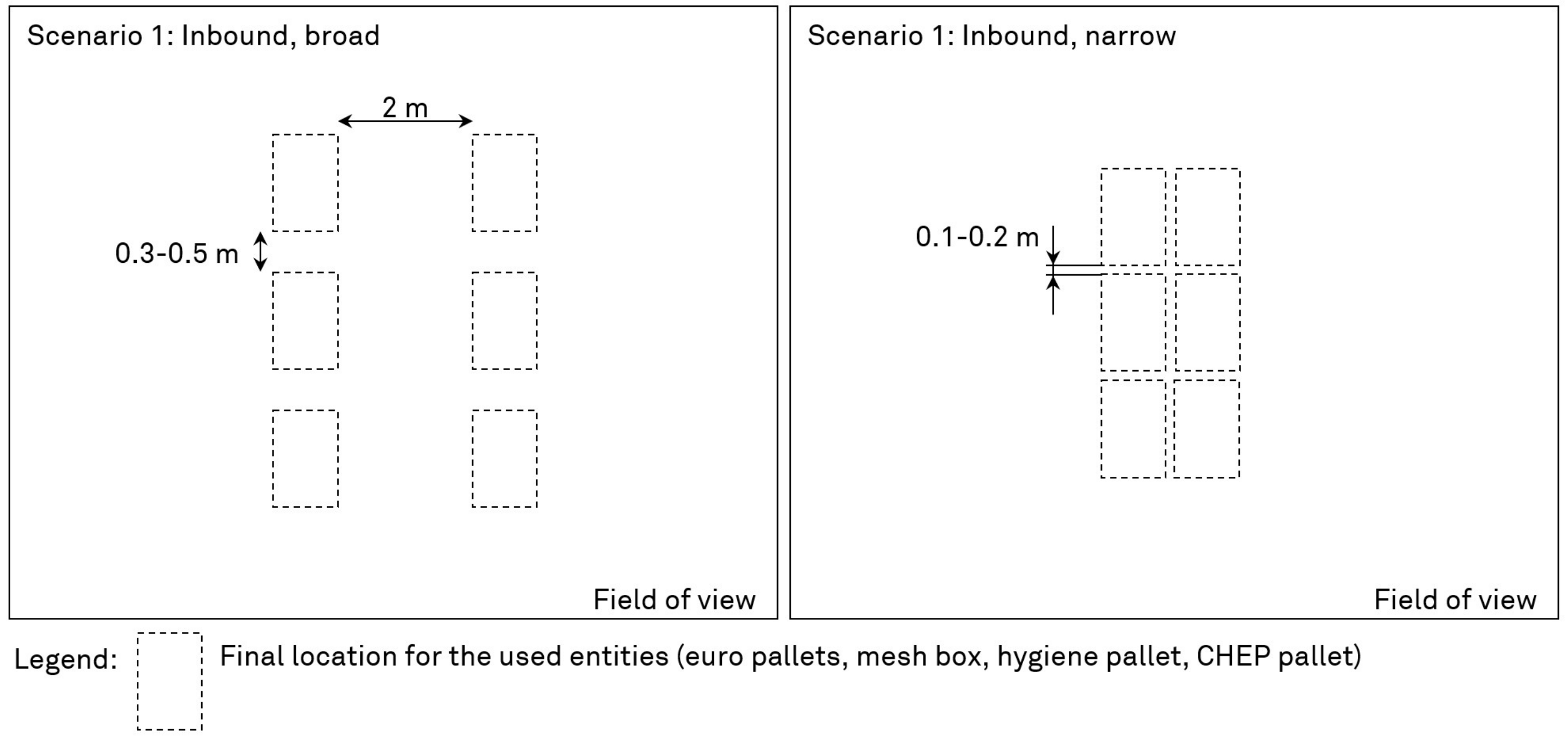}
        \caption{Schematic illustration of scenario $1$ with its two degrees of pallet proximity.}
        \label{fig:scenario1}
\end{figure}

\begin{figure}[htbp]
    \centering  
        \includegraphics[width=0.8\linewidth]{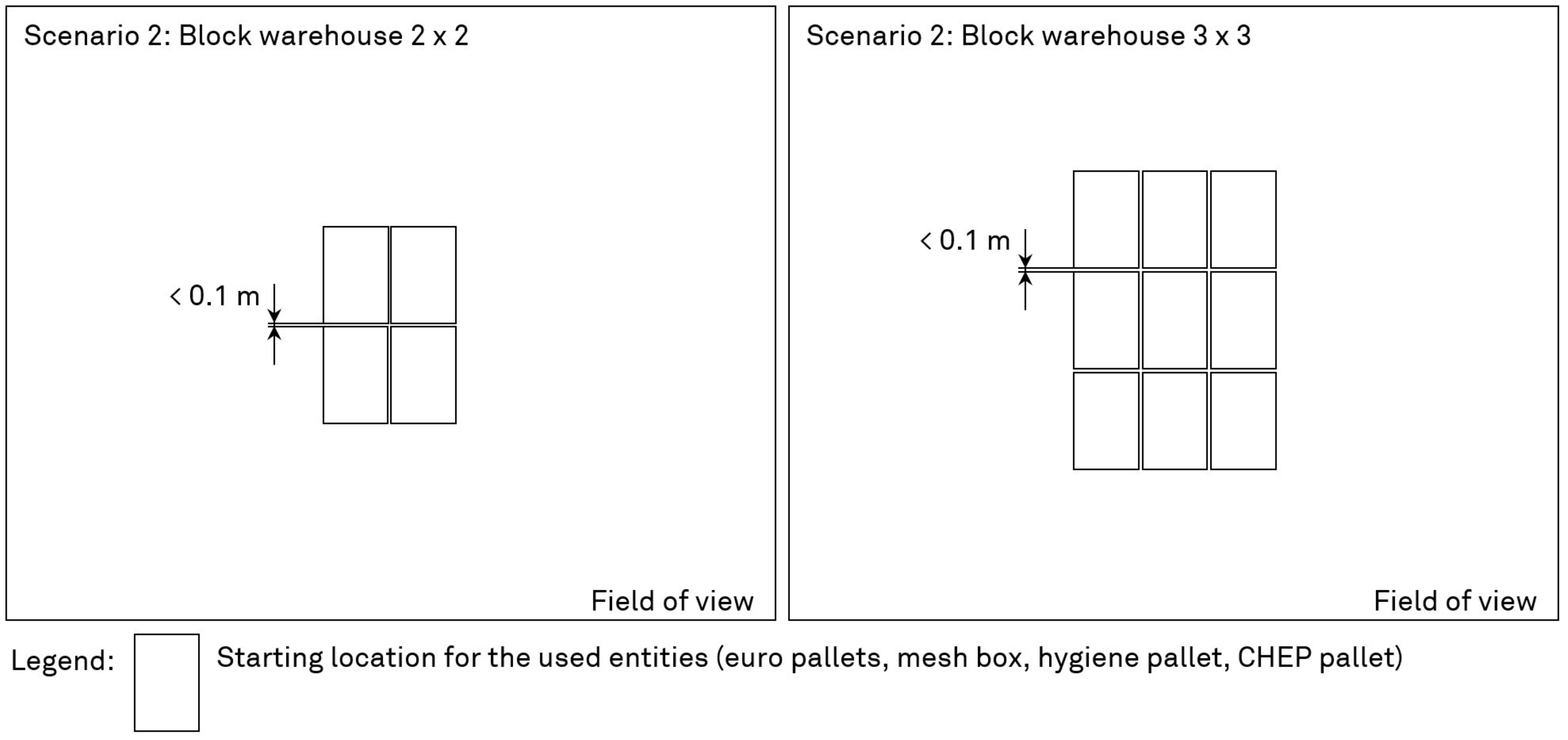}
        \caption{Schematic illustration of scenario $2$ with its two block warehouse pallet ordering structures.}
        \label{fig:scenario2}
\end{figure}

We define a pallet to be fully loaded if it is stacked with three layers of small load carriers on top of one another. 
In addition, entities such as barrels and cardboard boxes have been used and were not stacked. 
The first scenario, shown in Fig. \ref{fig:scenario1}, mimics an inbound material flow scenario that starts with an empty loading area, with the pallets set up to fill said area along the process.
The dotted lines represent the spots that the pallets are placed in during this scenario.
In the first stage, they are placed apart from one another while in the second scenario, they are placed more closely together. 
In the block warehouse scenario, shown in Fig. \ref{fig:scenario2}, the recordings being with a block of pallets that is already set up.
Subsequently, individual pallets are pulled out and moved outside of the field of view of the cameras. 
For this scenario, a $2$~×~$2$ block of pallets has been used in the first stage, and a $3$~×~$3$ one in the second stage. 

\begin{figure}[hbtp]
    \centering
    \begin{subfigure}{0.3\linewidth}
        \includegraphics[width=\linewidth]{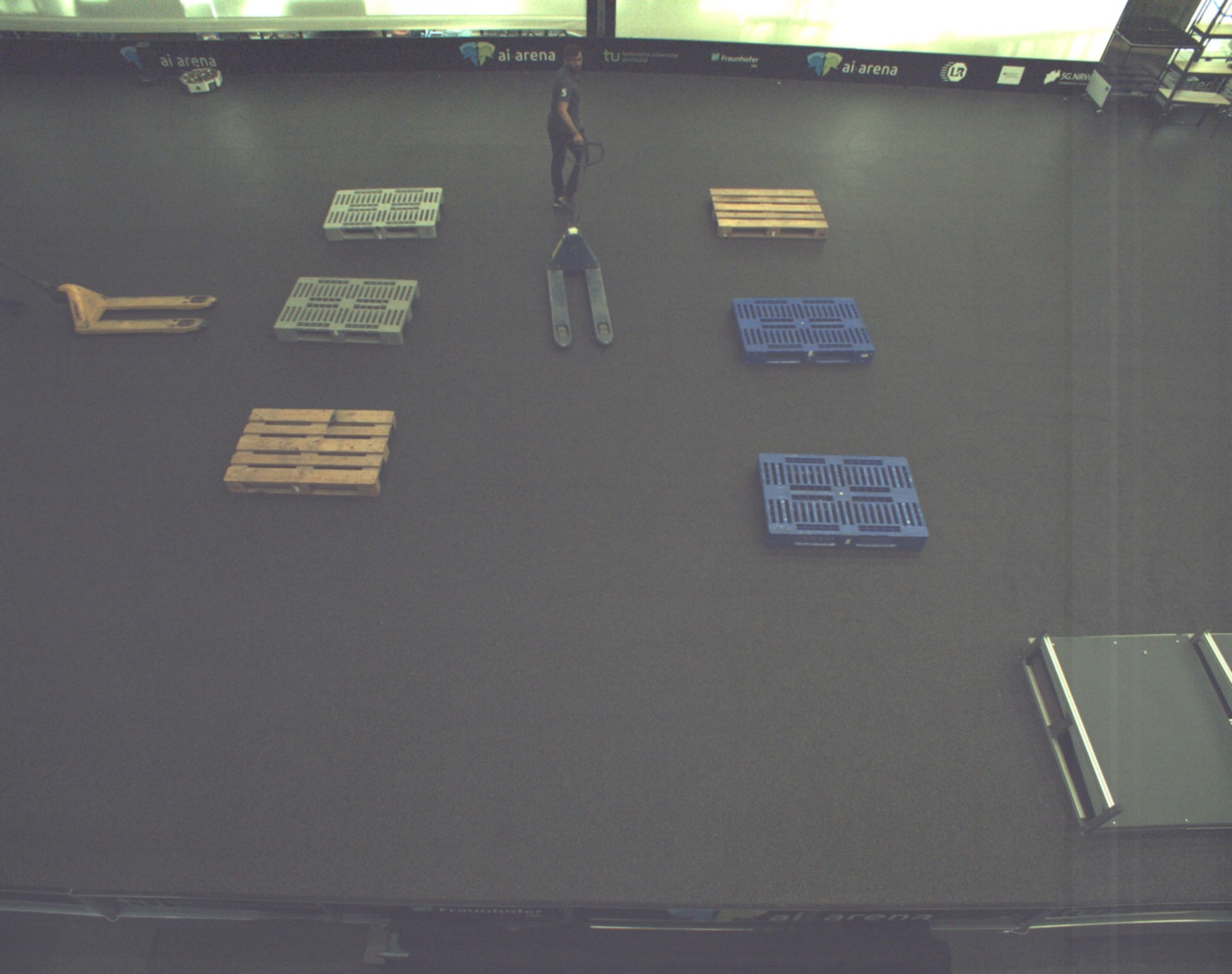}
        \caption{Scenario 1 - Broad, Stage 1}
        \label{fig:sc_1-Diff_1-broad}
    \end{subfigure}
        \begin{subfigure}{0.3\linewidth}
        \includegraphics[width=\linewidth]{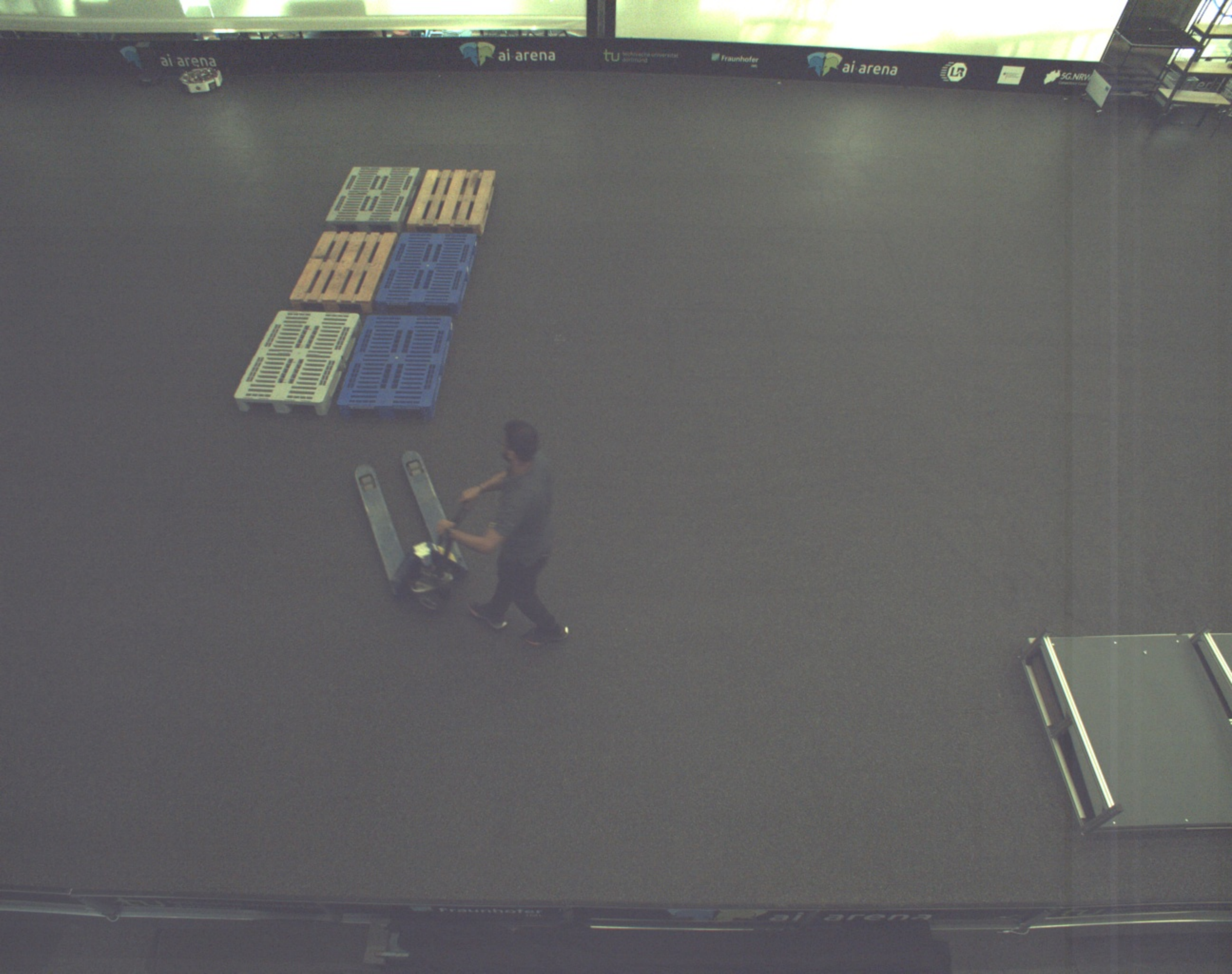}
        \caption{Scenario 1 - Narrow, Stage 1}
        \label{fig:sc_1-Diff_1-narrow}
    \end{subfigure}
    \begin{subfigure}{0.3\linewidth}
        \includegraphics[width=\linewidth]{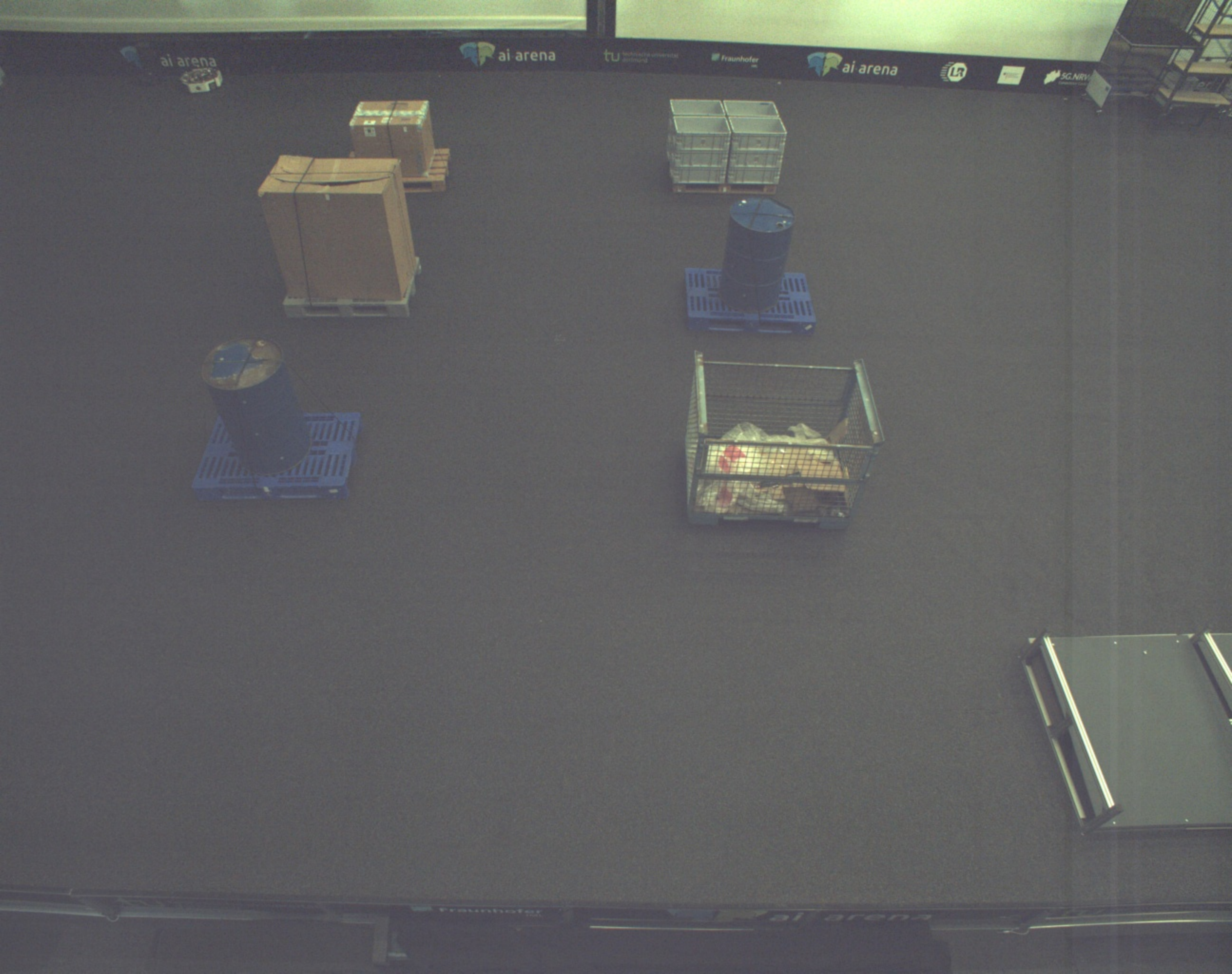}
        \caption{Scenario 1 - Broad, Stage 2}
        \label{fig:sc_1-Diff_2-broad}
    \end{subfigure}
    \begin{subfigure}{0.3\linewidth}
        \includegraphics[width=\linewidth]{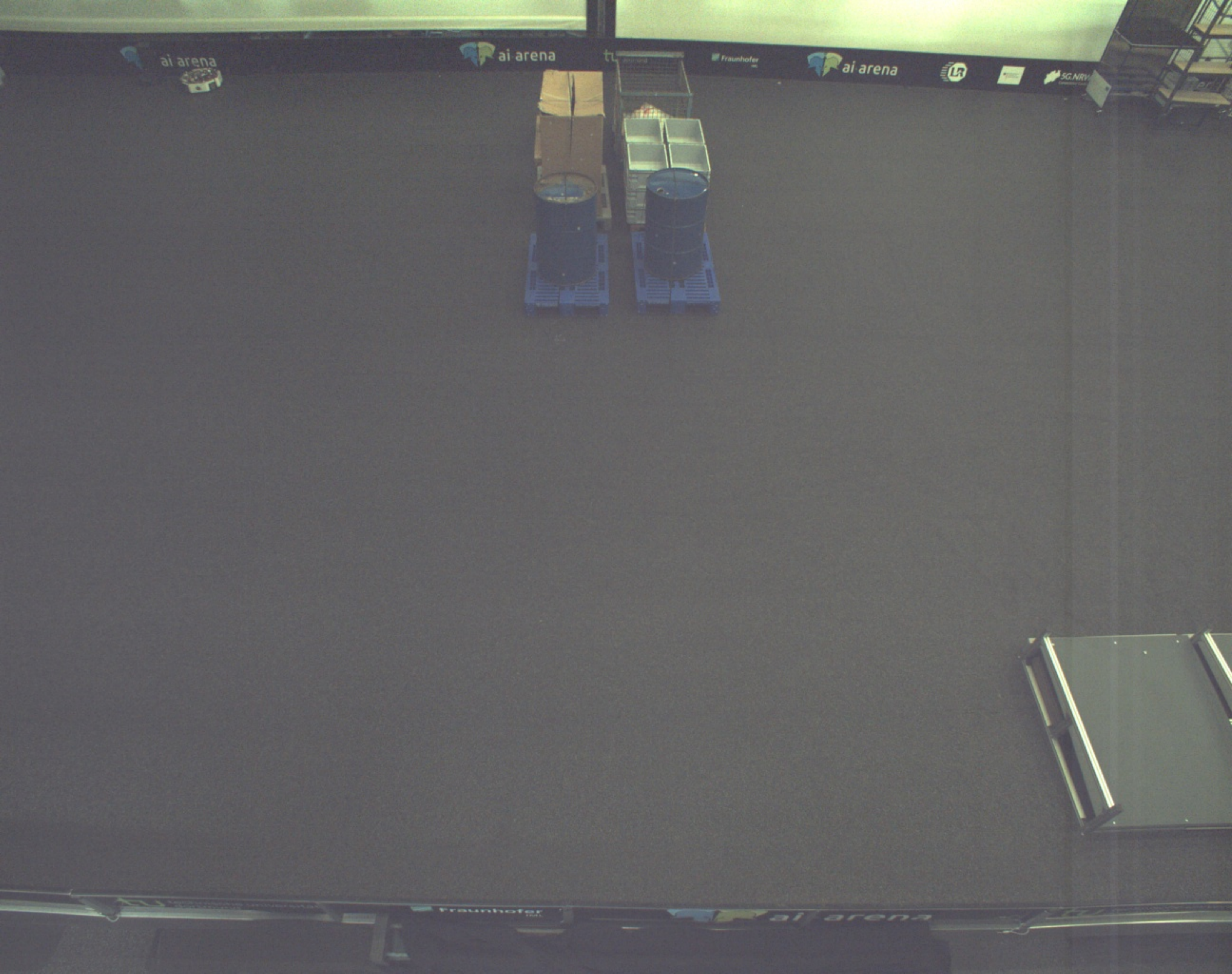}
        \caption{Scenario 1 - Narrow, Stage 2}
        \label{fig:sc_1-Diff_2-narrow}
    \end{subfigure}
    \begin{subfigure}{0.3\linewidth}
        \includegraphics[width=\linewidth]{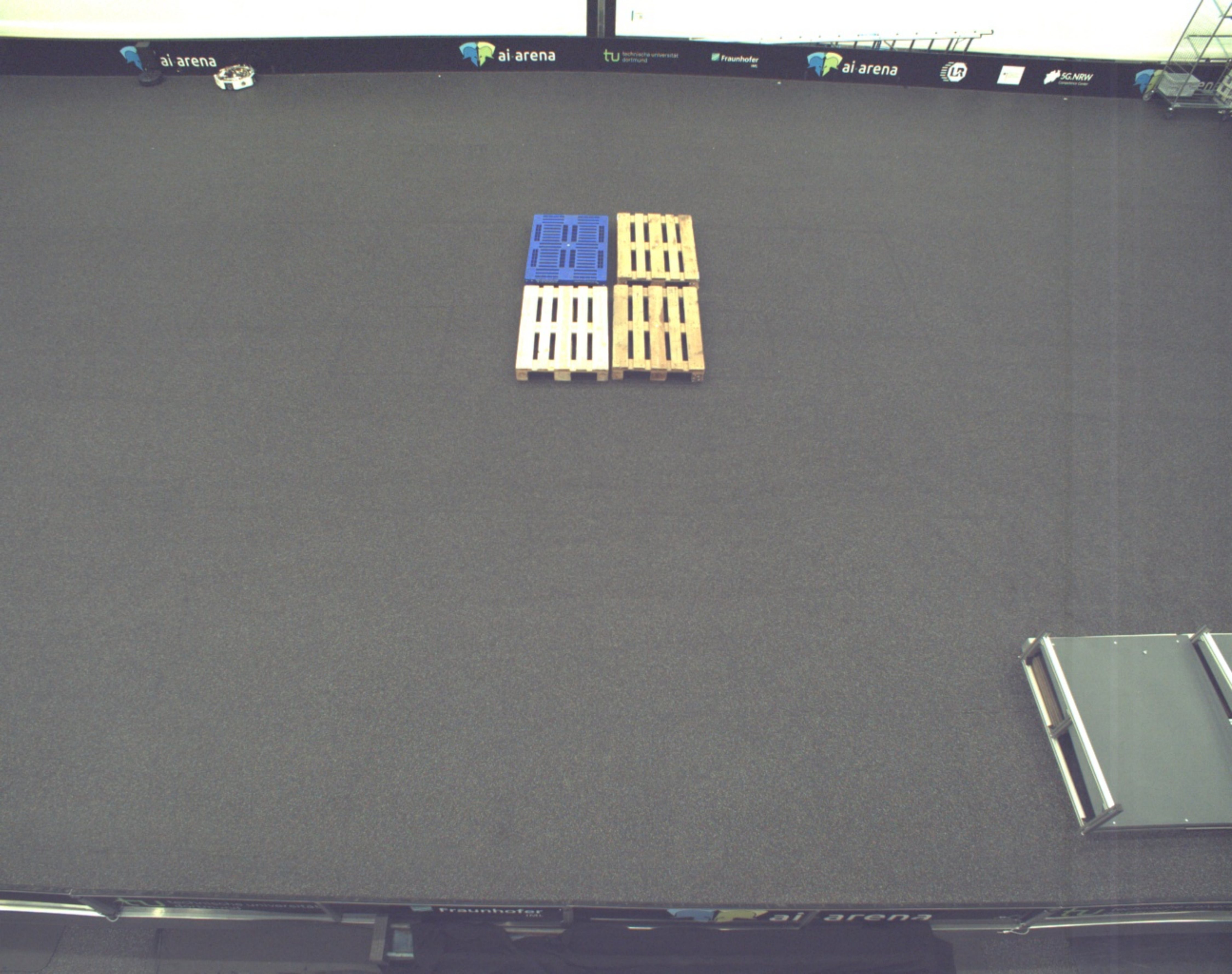}
        \caption{Scenario 2 - 2 x 2, Stage 1}
        \label{fig:sc_2-Diff_1-2x2}
    \end{subfigure}
    \begin{subfigure}{0.3\linewidth}
        \includegraphics[width=\linewidth]{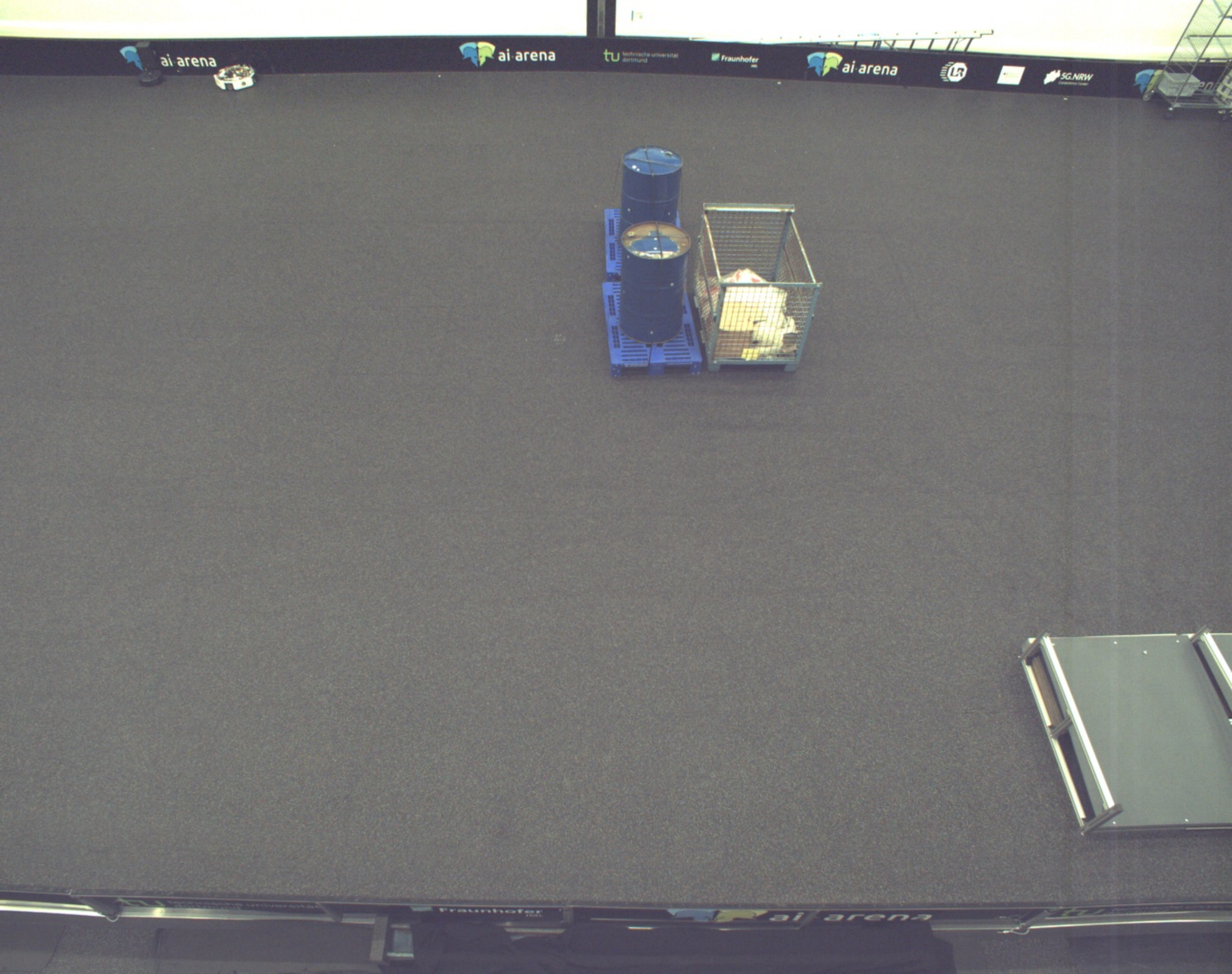}
        \caption{Scenario 2 - 2 x 2, Stage 2}
        \label{fig:sc_2-Diff_2-2x2}
    \end{subfigure}
    \begin{subfigure}{0.3\linewidth}
        \includegraphics[width=\linewidth]{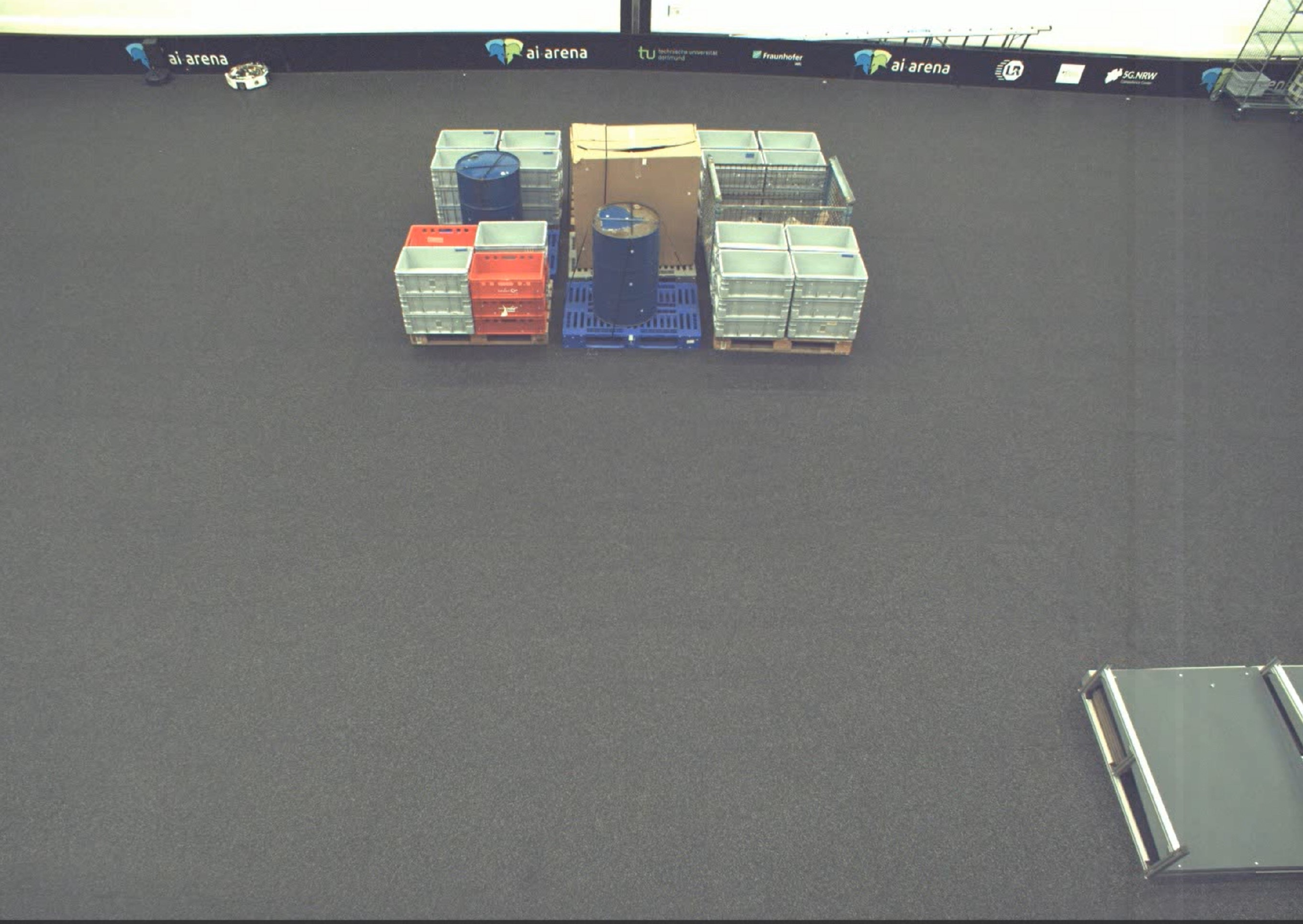}
        \caption{Scenario 2 - 3 x 3, Stage 2}
        \label{fig:sc_2-Diff_2-3x3}
    \end{subfigure}
    \caption{Frames taken from the two scenarios and their respective stages, used for our recordings.}
    \label{fig:scenrios}
\end{figure}

In total seven recordings are performed, as shown in Fig. \ref{fig:scenrios}. Fig. \ref{fig:sc_1-Diff_1-broad} shows scenario $1$, stage $1$, during which the pallets are arranged with a considerable distance between them. 
The inspiration for this scenario is that the two lanes that are built in this way could be found in the goods-receiving area of a warehouse, e.g., as to unload trucks. 
The pallets are then unloaded, e.g., from a truck and are placed far apart to allow warehouse workers to inspect the newly arrived goods. 
In Fig. \ref{fig:sc_1-Diff_1-narrow}, scenario $1$, stage $1$ with the closely placed pallets is shown. 
This scenario mirrors the loading process as it could be expected to be performed to load a truck.
Fig. \ref{fig:sc_1-Diff_2-broad} and Fig. \ref{fig:sc_1-Diff_2-narrow} show the first scenario in their second stage, i.e., with loaded pallets. 
Lastly, Fig. \ref{fig:sc_2-Diff_1-2x2}, \ref{fig:sc_2-Diff_2-2x2} and \ref{fig:sc_2-Diff_2-3x3} show the second scenario, which mimics a block warehouse, in the above mentioned stages. 
During the recording of these scenarios, varying lighting conditions were used.

\subsection{Setup and Data Collection}
\begin{figure}[hbtp]
    \centering
    \begin{subfigure}{0.8\linewidth}
        \begin{subfigure}{0.325\linewidth}
            \includegraphics[width=\linewidth]{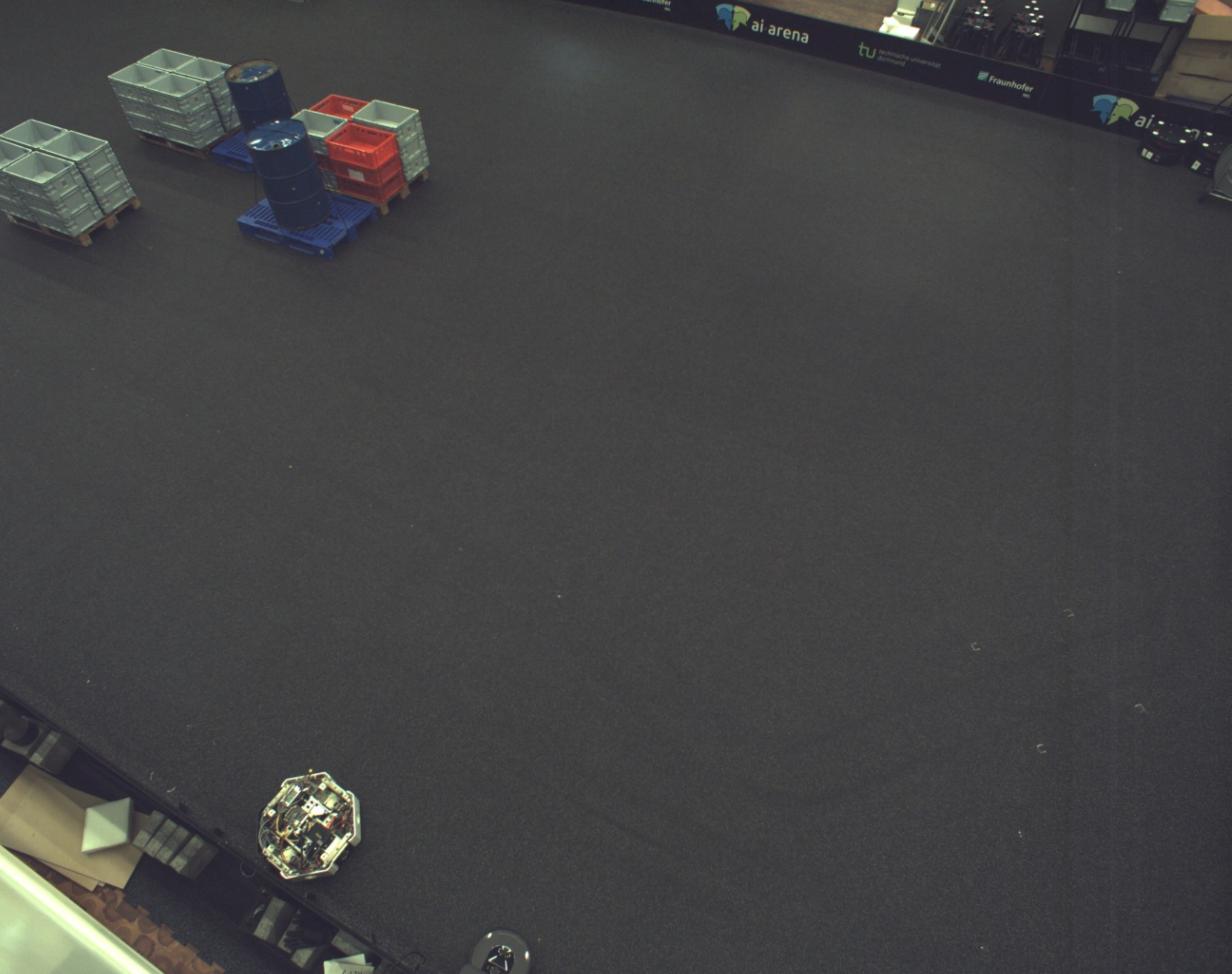}
        \end{subfigure}
        \begin{subfigure}{0.325\linewidth}
            \includegraphics[width=\linewidth]{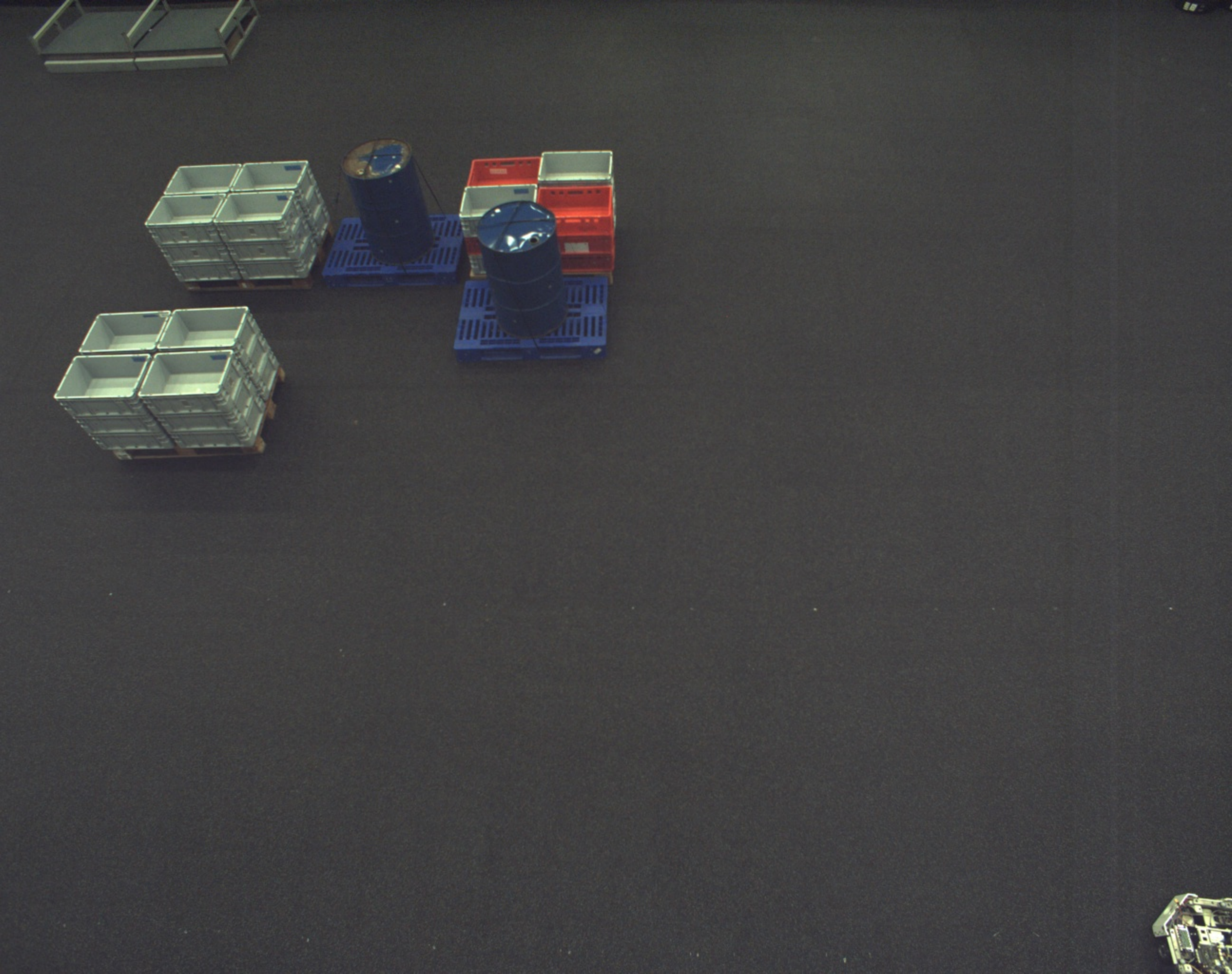}
        \end{subfigure}
        \begin{subfigure}{0.325\linewidth}
            \includegraphics[width=\linewidth]{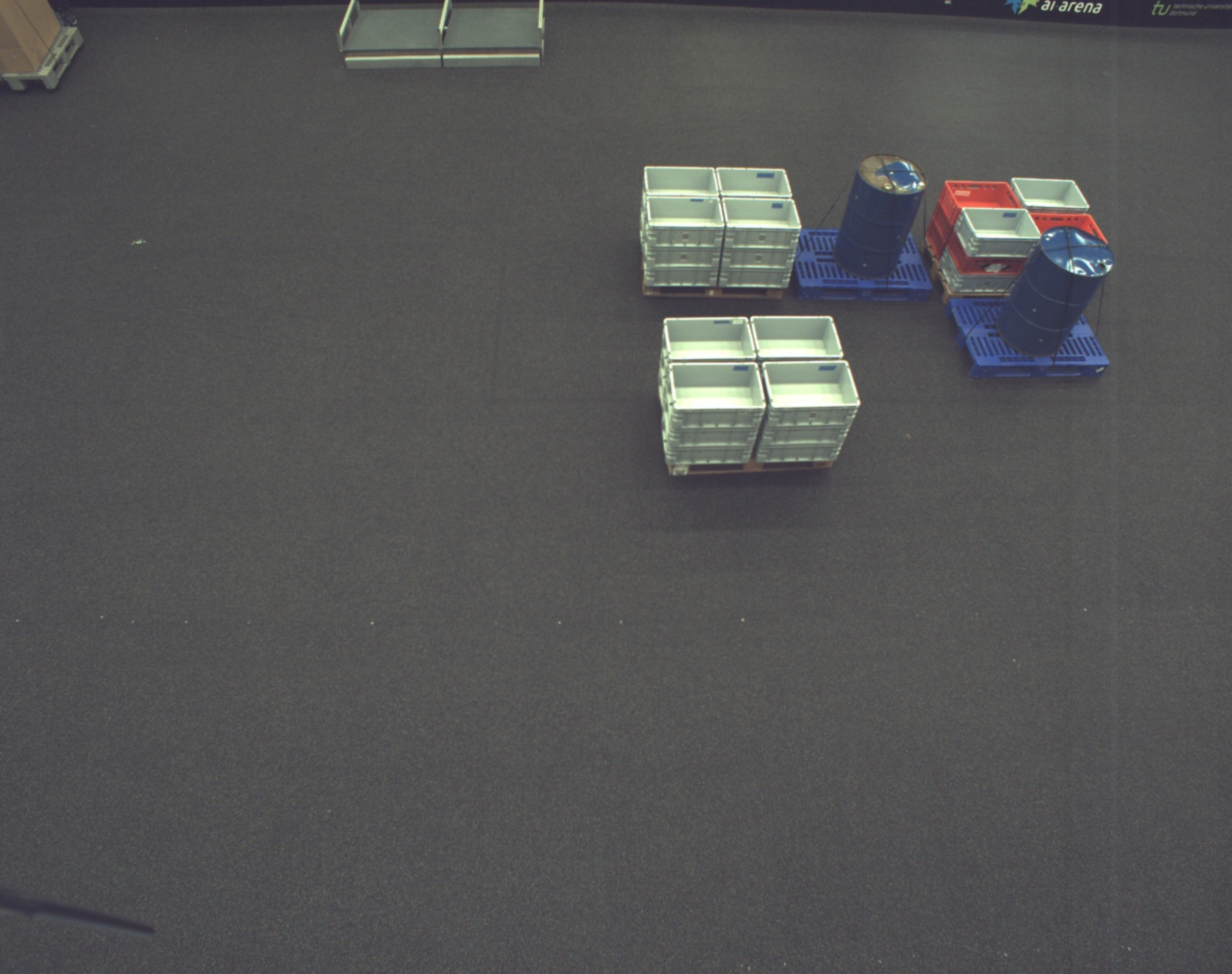}
        \end{subfigure}
        \begin{subfigure}{0.325\linewidth}
            \includegraphics[width=\linewidth]{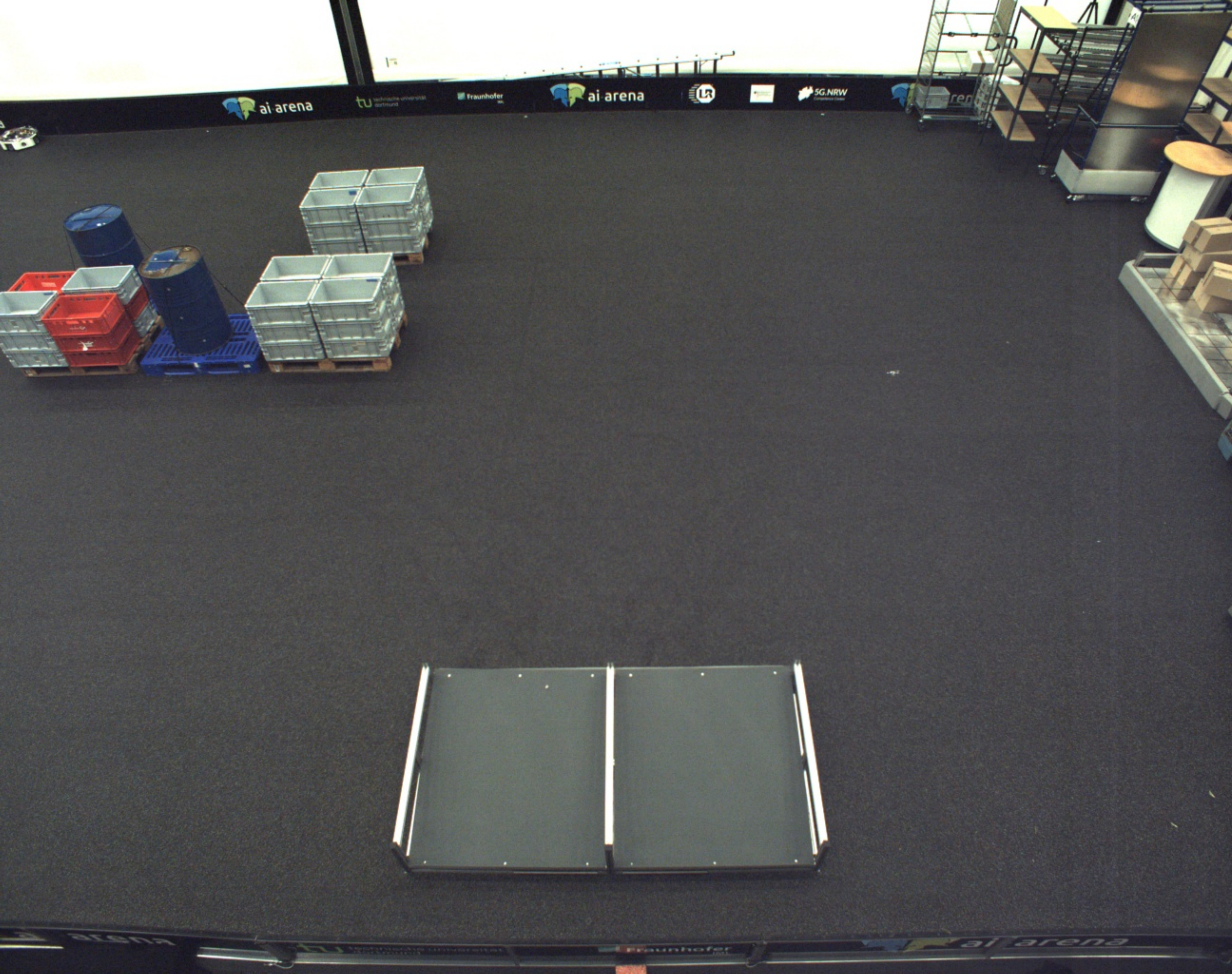}
        \end{subfigure}
        \begin{subfigure}{0.325\linewidth}
            \includegraphics[width=\linewidth]{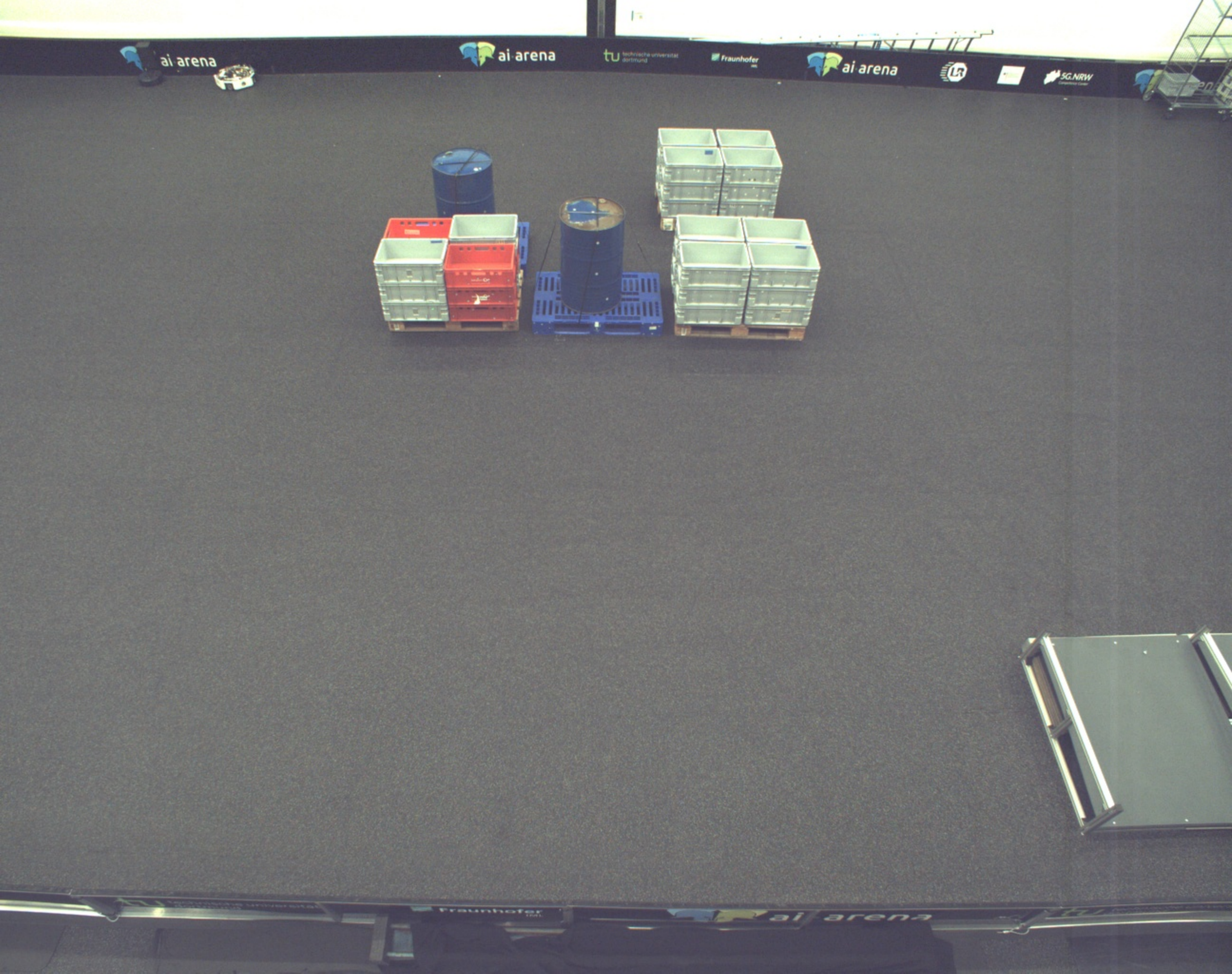}
        \end{subfigure}
        \begin{subfigure}{0.325\linewidth}
            \includegraphics[width=\linewidth]{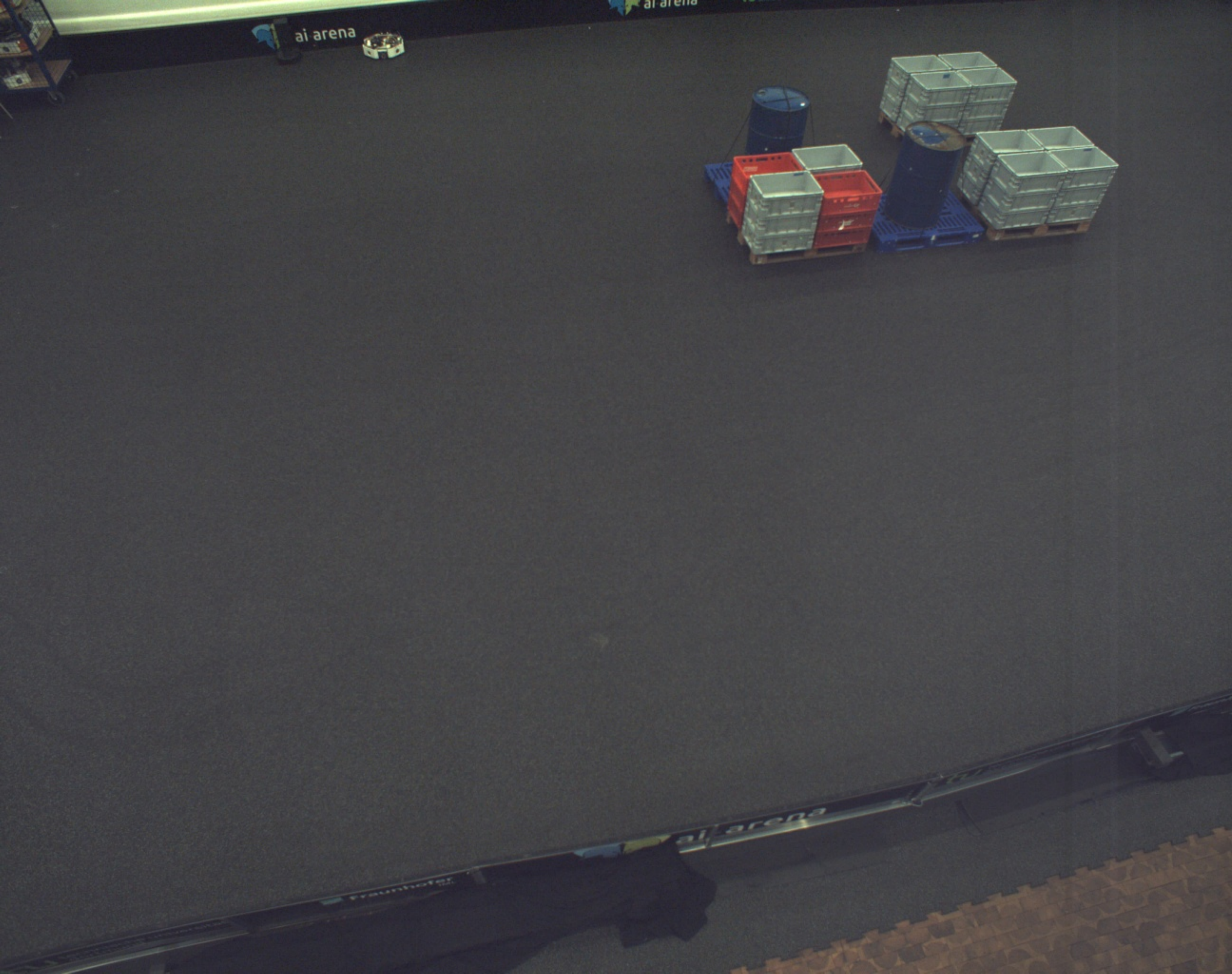}
        \end{subfigure}
        \caption{}
    \end{subfigure}
    \begin{subfigure}{0.8\linewidth}
        \includegraphics[width=\linewidth]{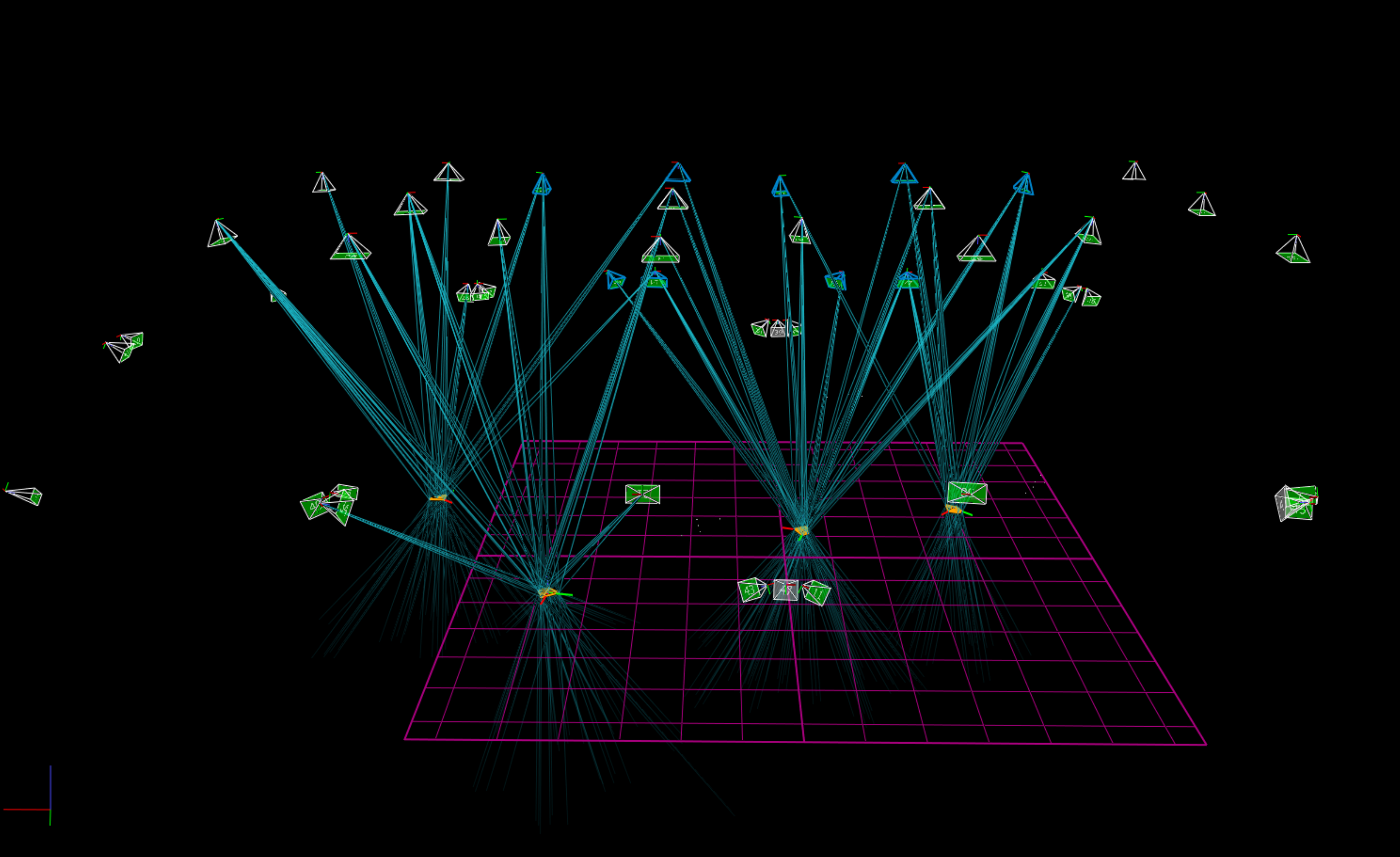}
        \caption{}
    \end{subfigure}
    \caption{Data collection setup. (a) shows the RGB images of the same scene as viewed from the six camera, (b) shows entities as perceived in the motion capturing system (obtained for a different scene). Rays show the detected retro-reflective markers by the system.}
    \label{fig:setup}
\end{figure}

 The area that is used to record the data proposed in this work is a former warehouse that has been transformed into an applied research facility.
 Its recording space is covered by six monocular RGB cameras providing parallel video streams.  
 The area is also covered by a marker-based motion capture system \cite{s20154083} comprised of $52$ infrared cameras.
 These cameras provide accurate poses of the tracked entities with respect to a common reference frame.
 This setup is shown in Fig. \ref{fig:setup}.

The dataset is collected by deploying industrial entities within the recording space, according to the configuration of the scenarios mentioned in \cref{sec:scenarios}. 
The entities are moved around by human operators to simulate inbound and outbound operations, again according to the previously described scenarios. 
While doing so, a video stream is captured through the RGB cameras. 
Simultaneously, the ground truth pose information for all tracked entities is acquired through the motion capture system.

\subsection{Data Processing}

The data collected by the motion capture system and the RGB camera system are processed on separate computers. The aim is to reduce the processing time necessary to request pose frames from the motion capture system and thus to increase the frames per second (FPS) of the streamed images from the RGB camera system. The frames from each of the six RGB cameras are collected on one computer along with their timestamps. The second computer collects information on entity IDs, entity poses, and timestamps from the motion capture system. The start and stop of collection from each of the systems are triggered manually. Each system's streams are synchronized in a post-processing phase.

In terms of hardware, six Genie Nano C2590 RGB cameras with $2$ MP resolution are used. The cameras are fitted with a Kowa LM8HC-SW lens with a $79.4$~×~$63.0$ field angle. All six cameras are connected to $10$ Gigabit Ethernet switch, which passes the streamed data to a data collection computer via an optical fiber network connection. The motion capture system consisting of $52$ cameras uses a mixture of Vicon Vero and Vicon Vantage cameras that are mounted on the ceiling and at different elevations in our research facility.

RGB camera settings such as brightness and white balance values were allowed to update periodically throughout the recordings. Illumination in the recording space was kept constant throughout each individual recording, changing in between recordings, and there was no significant color hue variation from the scene. Images were stored in raw \emph{bmp} format and distortion was preserved.

\subsubsection{Synchronization} \label{sec:synch}

The recording of video streams is event-triggered for each camera. However, to guarantee an equal number of retrieved images from all cameras, simultaneous capturing is necessary. Synchronized, simultaneous capturing also has the advantage of preserving the instantaneous state of the scene. Recording in such a manner can facilitate performing hand-offs between the different perspectives for multi-camera tracking algorithms. This also has the advantage of enabling more accurate re-identification of entities from different viewpoints.

Simultaneous capturing is done for all cameras by triggering a single image capture on each camera followed by trigger locking to prevent further capturing. The software lock is released on all cameras simultaneously only when image retrieval on all cameras has ended. Thus, for each capturing trigger, the slowest camera determines the overall FPS of the system. An average of approximately $20$ FPS per scenario is achieved.

Beyond achieving synchronization amongst the RGB cameras, it is necessary to synchronize between the RGB camera system and the motion capture system due to data capturing rate differences. During our experiments, the motion capture system had a fixed pose update rate of $200$ Hz. We match image frames to their respective poses based on the smallest timestamp difference between both instances. Since entities in the scene move at less than $1 \frac{m}{s}$ and due to the high update rate of the motion capture system, pose differences between consecutive frames are insignificant. The synchronization between both streams is accomplished as a post-processing step.

\subsubsection{Data Structure}
Since the currently available datasets for object tracking lack the combination of systems used in this work, we collect our data and process it into a custom data structure. The final annotation data structure of our custom dataset is shown in Table \ref{table:data}.
\begin{table}[htbp]
\centering
 \caption{Sample entries in post-processed annotated data.}
    \begin{tabular}{|c c c|}
    \hline
        Image Path & Entity Name & Position\\  
        \hline\hline
        {...} & {...} & {...}\\
        camera\_6/images/3.jpg & Pallet\_9 & [-10672.35, 1815.89, 85.49]\\ 
        camera\_6/images/769.jpg & Forklift\_2 & [-3142.96, -1409.38, 239.16]\\
        {...} & {...} & {...}\\
        \hline
    \end{tabular}

    \begin{tabular}{|c c c c|}
        {...}Orientation & Delta Time & Bounding Box & Visible\\ 
        \hline\hline
        {...} & {...} & {...} & {...}\\ [0.5ex]
        {...}[0.0037, 0.0019, -1.5481] & -0.00088 & [-1, -1, -1, -1] & 0 \\ [0.5ex]
        {...}[-0.0035, -0.0036, -0.0014] & -0.0037 & [293, 0, 215, 339] & 1 \\
        {...} & {...} & {...} & {...}\\[1ex]
        \hline
 \end{tabular}

 \label{table:data}
\end{table}

The \emph{Image Path} refers to the relative image path with respect to each camera view. Images are converted to \emph{jpg} format for efficient storage. \emph{Entity Name} refers to the entity ID as retrieved via the motion capture system. It is worth noting that initially an entry is preserved for all entities in each captured image, regardless of their existence in the captured scene. During the annotation phase, as discussed in \cref{sec:annot}, invalid projections of the entities' 3D models are removed. \emph{Position} and \emph{Orientation} are $3$~×~$1$ vectors defining the relative pose of the entities in 3D space with respect to each camera. Position data are provided in mm and orientation data are provided in radians in intrinsic $XYZ$ Euler format. The position is obtained with respect to the motion capture system's global reference frame. The reference frames of the motion capture system and the RGB camera system are unified to enable the calculation of the transformation chain generating the entity's relative pose. The entry \emph{Delta Time} is the smallest calculated time offset between the capturing time of the RGB image and its corresponding pose. The \emph{Bounding Box} is the $4$~×~$1$ vector defining the pixel coordinates of the top left $x$ and $y$ coordinates, along with the width and height of the box. The \emph{Visible} flag indicates whether an entity is perceived in the field of view of the respective camera. The flag is generated automatically as part of the post-processing step of the annotation pipeline used. This is accomplished by disregarding entity 3D model projections when rendered at their ground truth pose, as discussed in \cref{sec:annot}. Bounding boxes that correspond to entities that are invisible in the relevant camera view are denoted with coordinates of $-1$. Invalid data from the motion capture system, such as those obtained when an entity is outside the system's region of operation, are filtered out in a post-processing step.

\subsection{Annotation} \label{sec:annot}

To maximize image capturing throughput, we separate the data collection phase from the annotation phase. In the annotation phase, we generate image annotations in an automated fashion by leveraging the 3D models' projection at the ground truth poses collected from the motion capture system. The annotation pipeline fits bounding boxes to the 2D image projections of the 3D models at their obtained poses in the scene relative to the camera of interest. 
 
The annotation pipeline is comprised of different phases. Initially, the RGB images and motion capture system poses are collected simultaneously. Then the reference frames of the motion capturing system and the RGB camera system are unified, and incoming streams from both systems are synchronized. This is followed by the main phase during which the relative transformations are calculated between the tracked entities of interest and each camera. Finally, the 3D models are projected at their calculated relative transformations where they are fitted with bounding boxes to generate the final image annotations.

\section{Results}\label{sec4}

The herein presented TOMIE dataset includes a total of $112,860$ images and $640,936$ entity instances.
In comparison to similar datasets, the number of captured images outnumbers the biggest dataset \cite{MOT16} by a factor of $4$, while the number of captured entity instances is approximately $25$\% smaller.

\begin{figure}[htbp]
    \centering
    \begin{subfigure}{0.4\linewidth}
        \includegraphics[width=\linewidth]{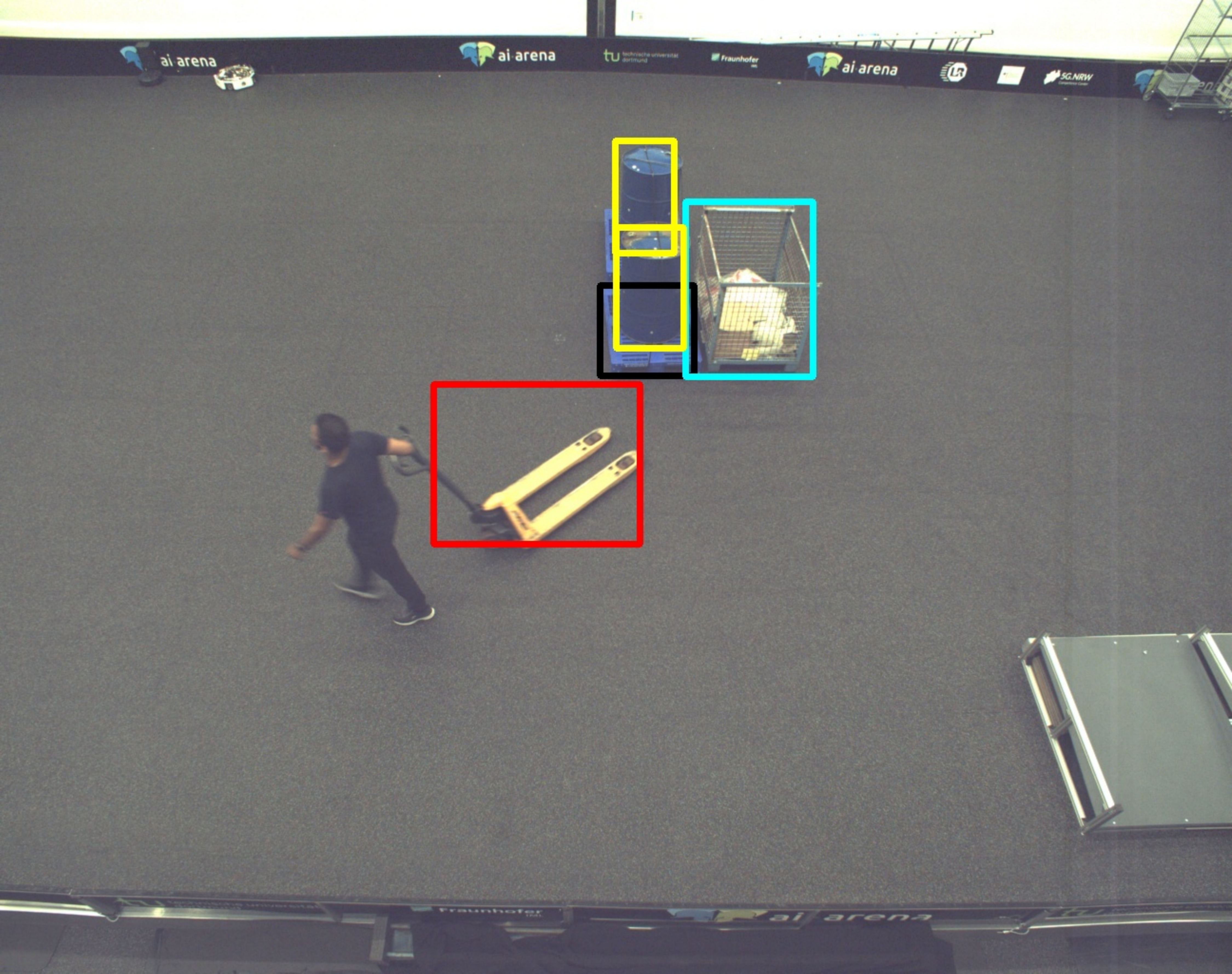}
    \end{subfigure}
        \begin{subfigure}{0.4\linewidth}
        \includegraphics[width=\linewidth]{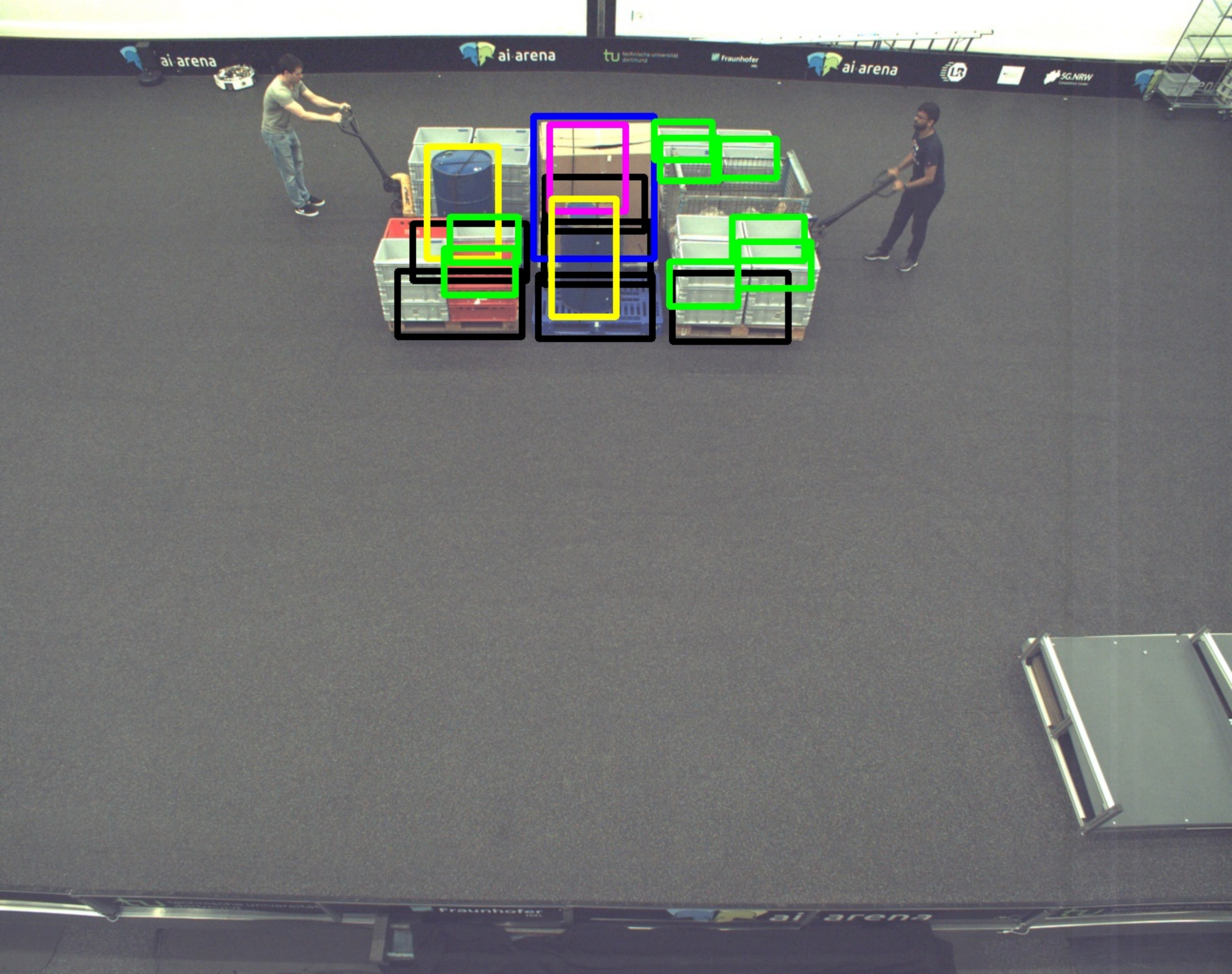}
    \end{subfigure}
    \begin{subfigure}{0.4\linewidth}
        \includegraphics[width=\linewidth]{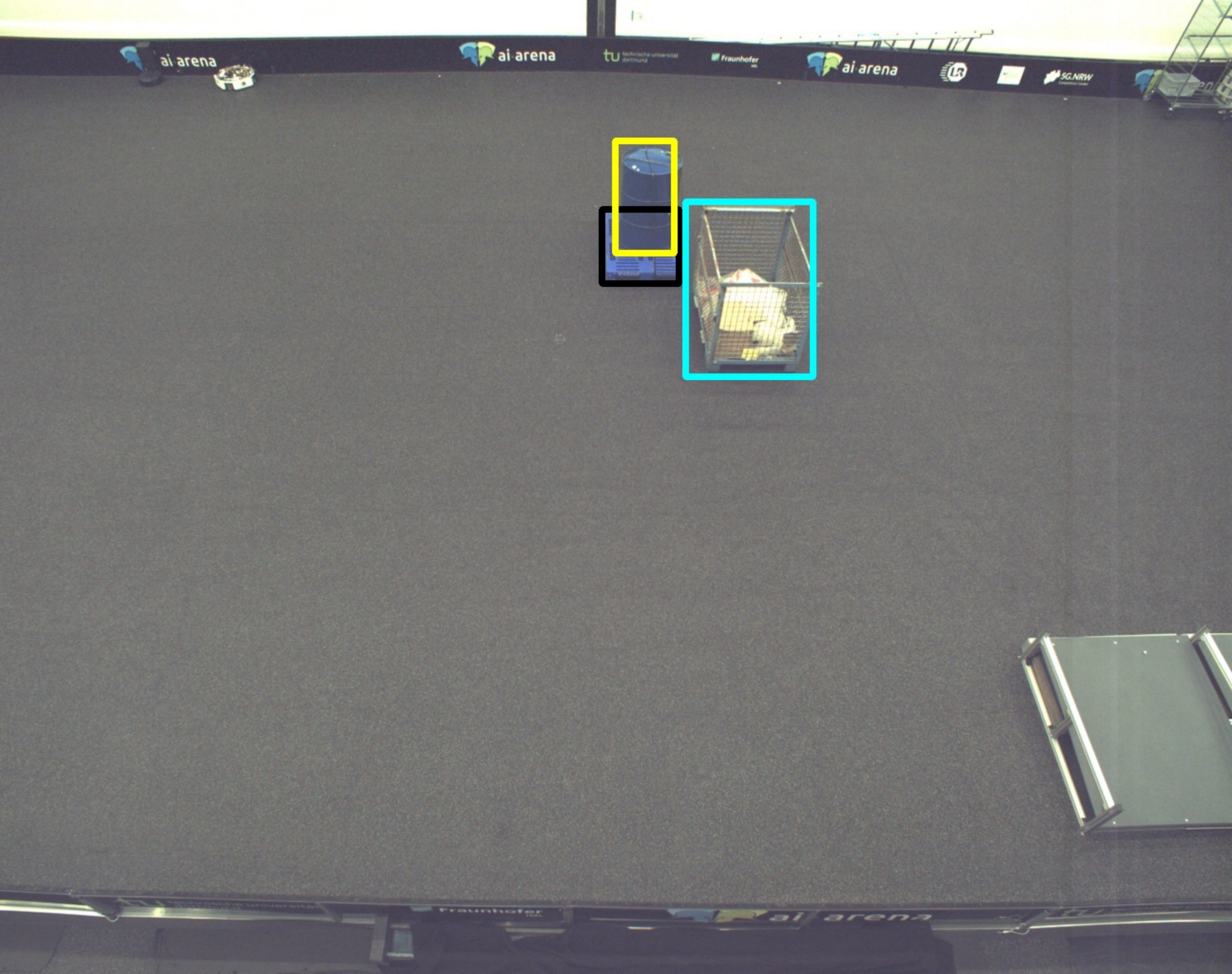}
    \end{subfigure}
    \begin{subfigure}{0.4\linewidth}
        \includegraphics[width=\linewidth]{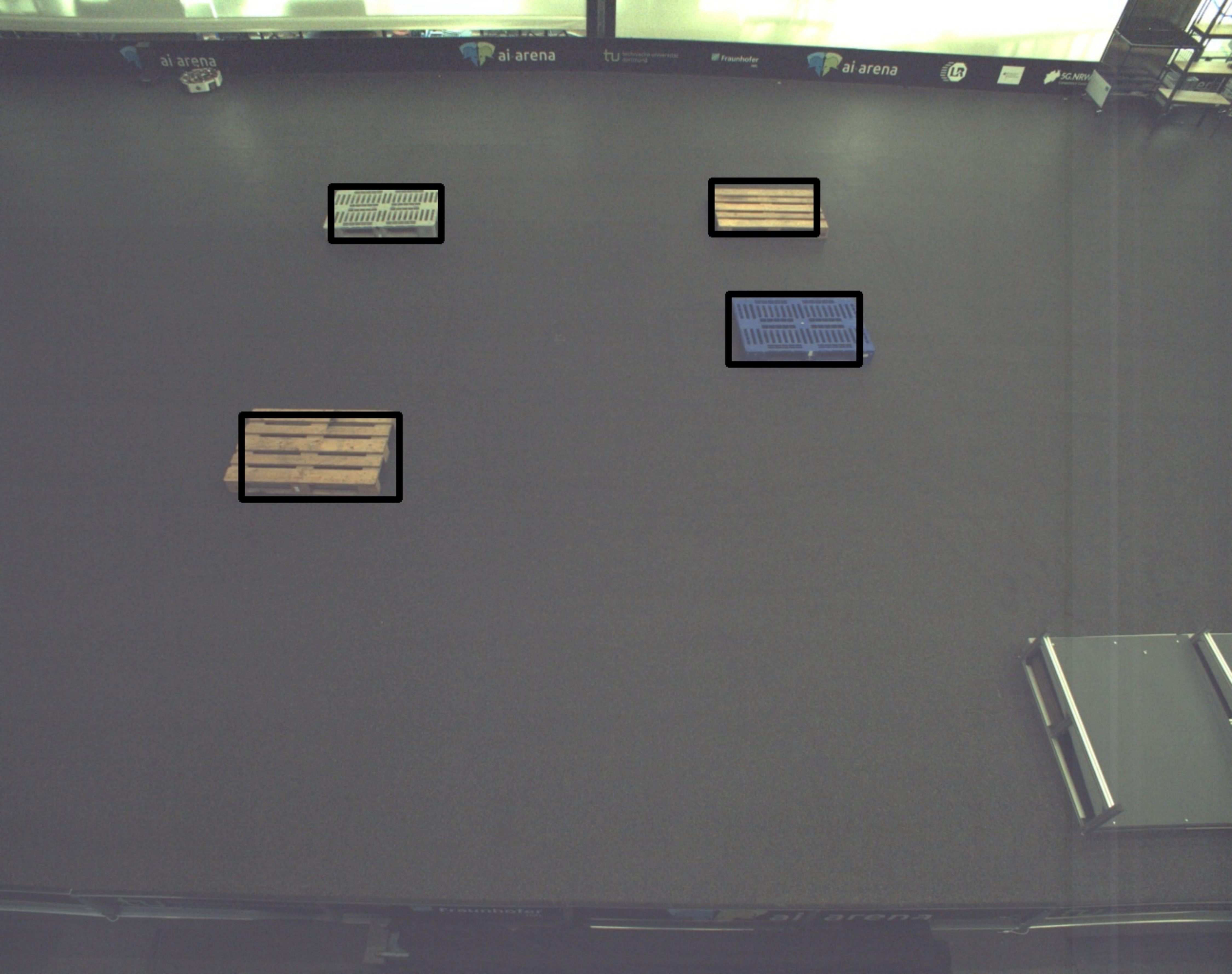}
    \end{subfigure}
    \caption{Samples of annotated images from a single view and different scenarios from our custom dataset. Bounding box colors are unique to each entity class.}
    \label{fig:samples}
\end{figure}

The annotations were generated using a computer equipped with an Intel Core i9 that possesses $28$ cores and $128$ GB of RAM. The renderer deployed, VisPy \cite{luke_campagnola_2022_5974509}, uses the onboard Nvidia Titan Xp GPU with $12$ GB of VRAM throughout the annotation process. Samples of annotated images are shown in Fig. \ref{fig:samples}. We provide the source code for our automated annotation pipeline \footnote{https://anonymous.4open.science/r/TOMIE-Dataset-0AE5}
for public usage as well as the source code for our data collection phase\footnote{https://anonymous.4open.science/r/RGB-Camera-System-BD98}.
\clearpage

During the annotation process, an average of $1.5$ s was spent on each object instance in the recording.
This amounts on average to $9$ s spent per image for the annotation of all visible entities.
The annotation speed achieved through the use of automated annotation is significantly higher than comparable manual annotation, like the one described in \cite{Adhikari18}. Dataset statistics per camera and per entity are shown in Table \ref{tab:stat_cam_1} and Table \ref{tab:stat_cam_2}. 

\begin{table}[ht]
    \centering
    \caption{Dataset statistics per camera.}
    \begin{tabular}{l|| r r r r r r}
        \textbf{Sequence}       &   I       &   II      &   III     &   IV      &   V       &   VI       \\ \hline
        \# instances            &   64,430   &   55,136   &   76,904   &   208,134  &   51,364   &   184,968  \\
        \# frames               &   14,825   &   19,141   &   20,767   &   23,359   &   12,651   &   22,117  \\
        Annotation time (min)   &   1,618    &   1,388    &   1,926    &   5,209    &   1,285    &   4,637    \\
    \end{tabular}
    \label{tab:stat_cam_1}
\end{table}

\begin{table}[ht]
    \centering
    \caption{Dataset statistics per entity class.}
    \setlength{\tabcolsep}{0,145cm}
    \begin{tabular}{l|| r r r r r r}
        \textbf{Entity}    &    Barrel  &   Forklift    &   Pallet      &   Mesh Box   &   Cardboard Box     &   Load carrier    \\ \hline
        \# instances        &   55,492   &   87,914       &   305,498      &   33,452   &   57,672   &   100,908          \\
    \end{tabular}
    
    \label{tab:stat_cam_2}
\end{table}

\begin{figure}[htbp]
    \centering
    \begin{subfigure}{0.4\linewidth}
        \includegraphics[width=\linewidth]{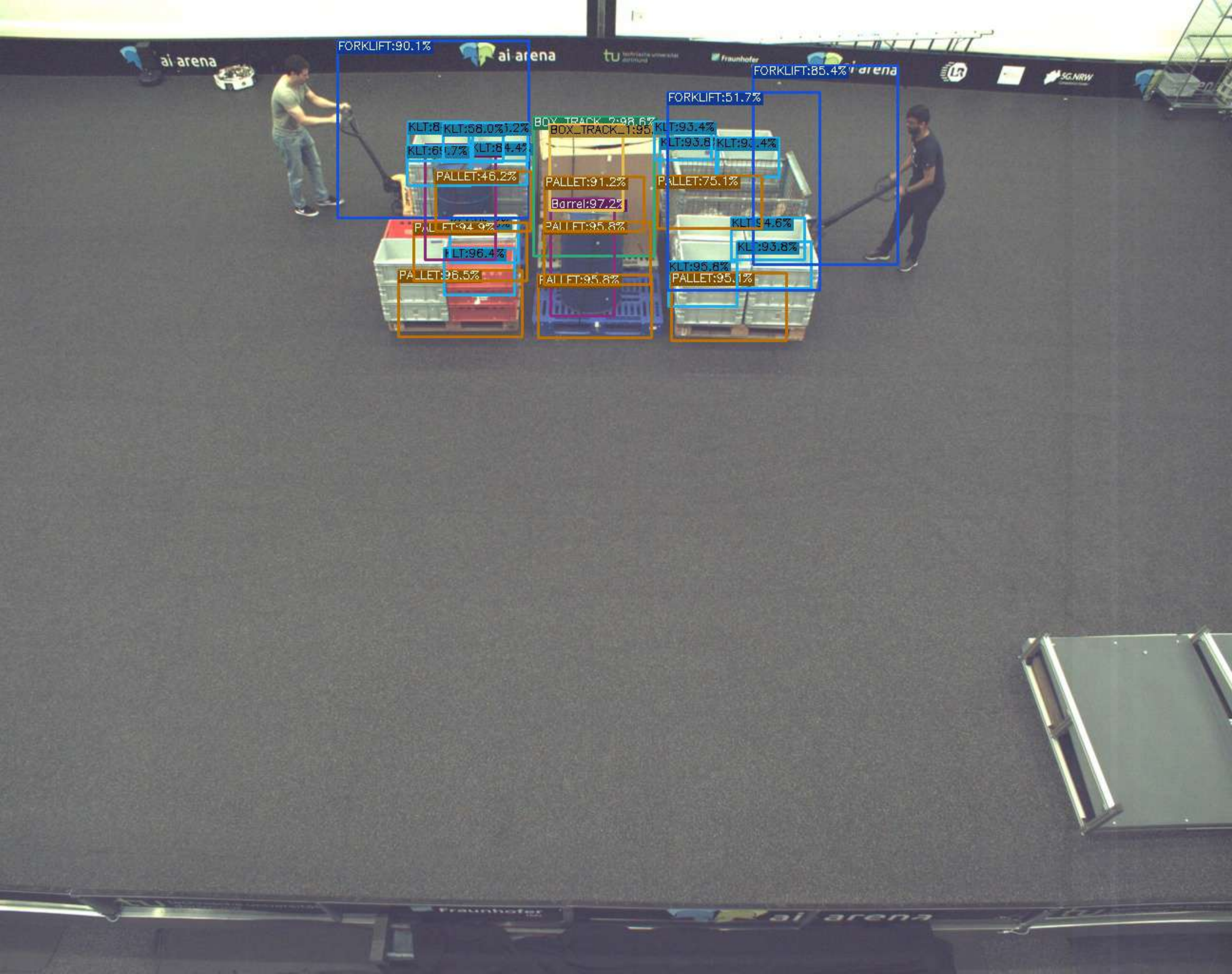}
    \end{subfigure}
    \begin{subfigure}{0.4\linewidth}
        \includegraphics[width=\linewidth]{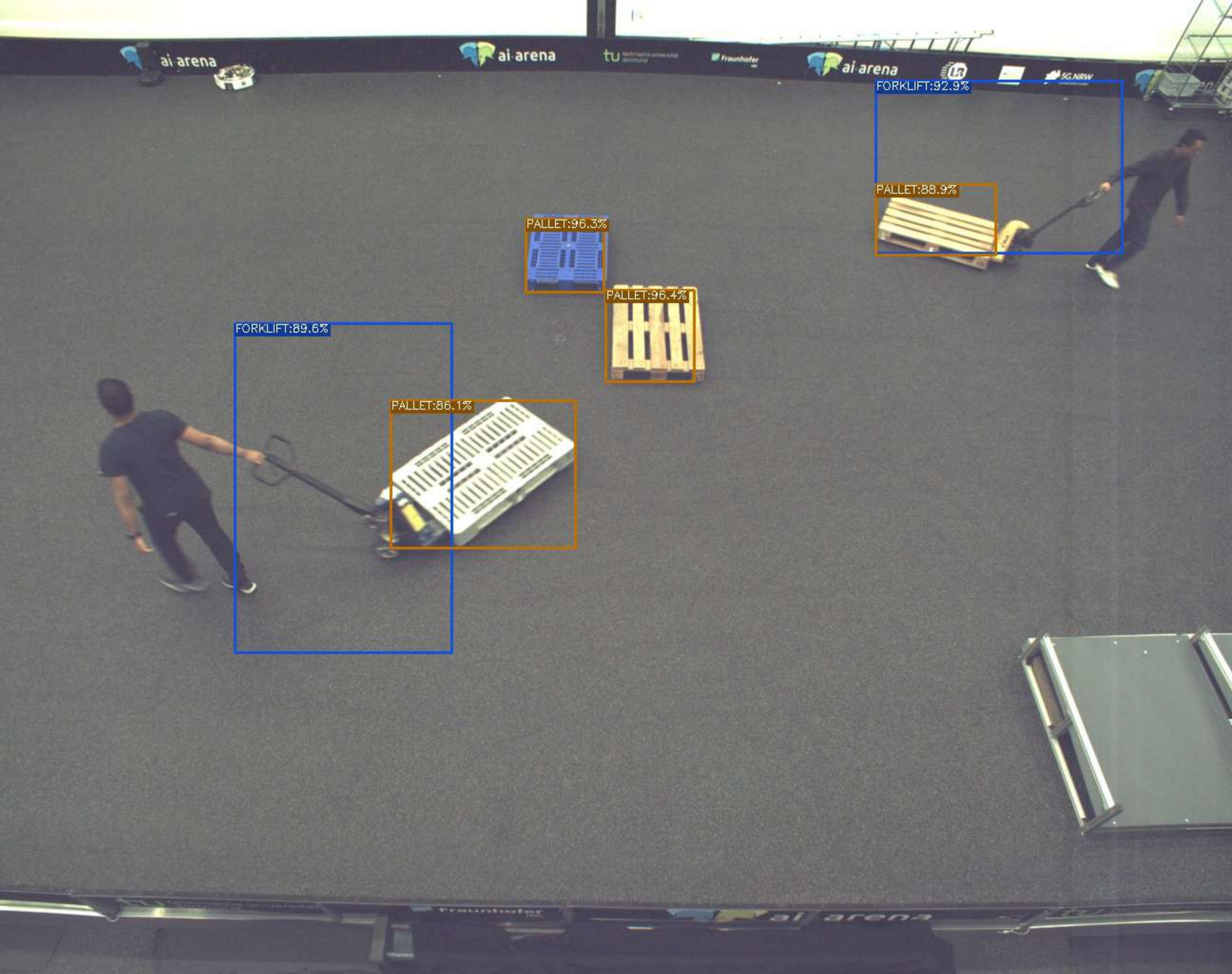}
    \end{subfigure}
    \begin{subfigure}{0.4\linewidth}
        \includegraphics[width=\linewidth]{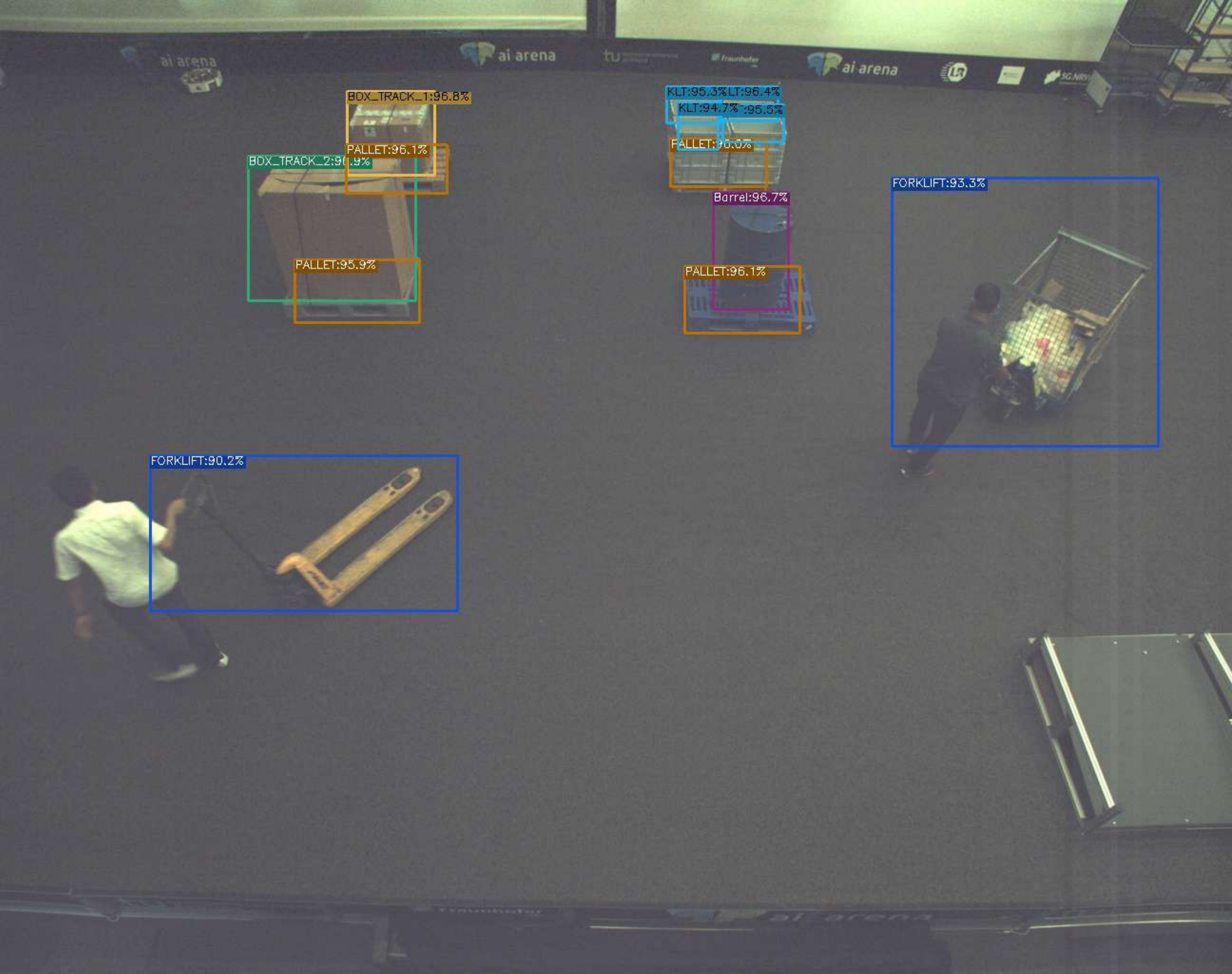}
    \end{subfigure}
    \begin{subfigure}{0.4\linewidth}
        \includegraphics[width=\linewidth]{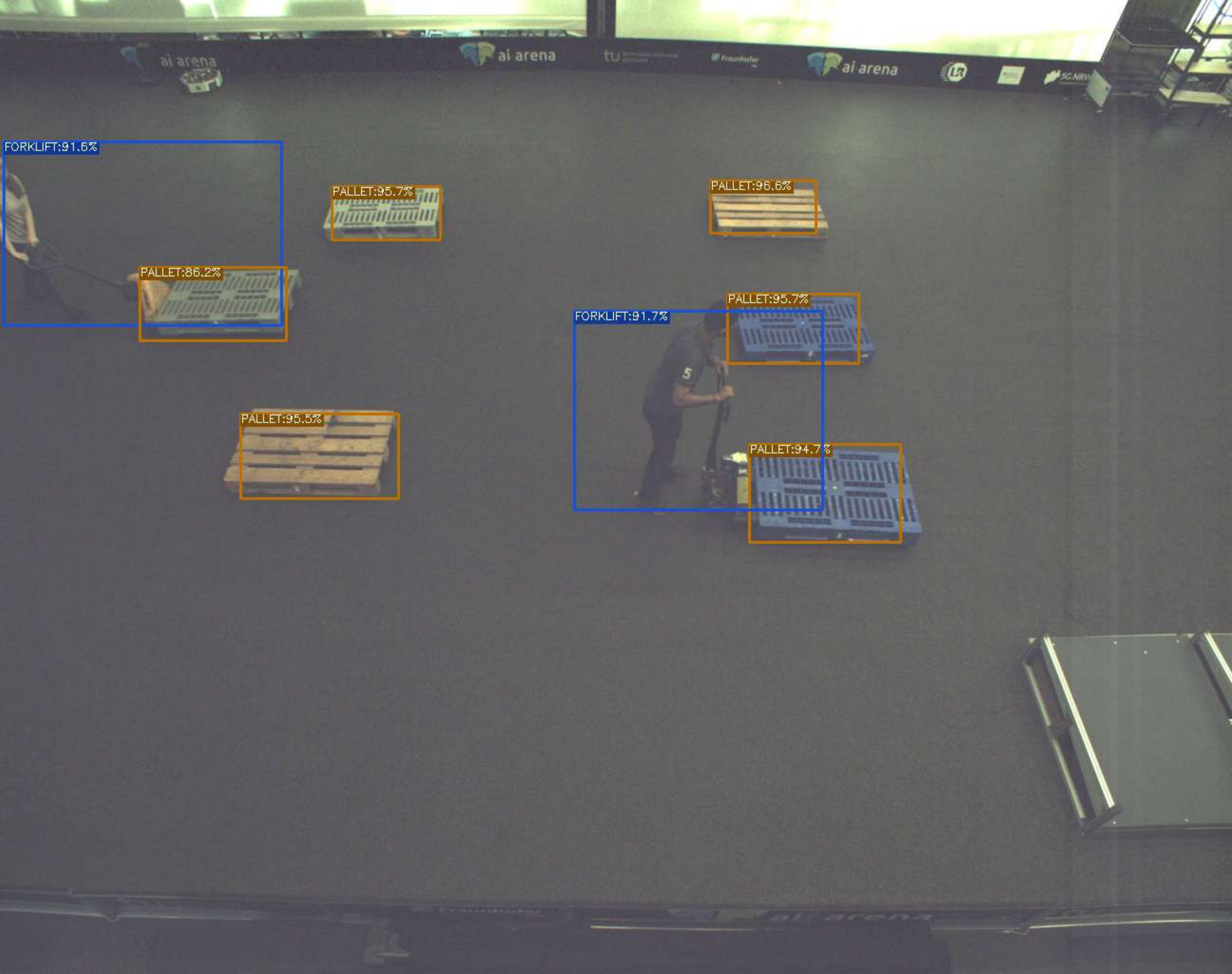}
    \end{subfigure}
    \caption{Samples of annotated images by the chosen object detector of different scenarios from our custom dataset.}
    \label{fig:annotation_yolox}
\end{figure}

\clearpage
To evaluate how far our custom dataset can be used for training classifiers that achieve a performance sufficient for industrial applications, multiple experiments were conducted. For these experiments, three of the currently best-performing models for the MOT20 \cite{MOT20} dataset, namely \textit{ByteTrack} \cite{Aharon22}, \textit{SiamMot} \cite{Siamese_Track_2020}, and \textit{Bot-SORT} \cite{Aharon22} were chosen. Publically available and official implementations for all models were used during the evaluation. 
The \textit{ByteTrack} and \textit{Bot-SORT} models rely on YoloX \cite{Megvii21} as a backbone for object detection. To this end, one YoloX model was pre-trained on our custom dataset to be used for both evaluation models. The average precision and recall of the resulting model were measured and are shown in Table \ref{tab:ap_ar_results}. The resulting object detection results are visualized on some samples of our custom dataset in Fig. \ref{fig:annotation_yolox} as well.

All models were trained and evaluated on our custom dataset in accordance with their respective work. For evaluation, the CLEAR metrics \cite{Bernardin08}, including MOTA, as well as IDF1, and HOTA were used. These metrics evaluate different aspects of the detection and tracking performance. The results are displayed in  Table \ref{tab:reid_val_results}.

While the TOMIE dataset is composed of more data, the results show that the performance of the tracking algorithms does not match those of similar datasets.
This deficit could be the result of the change in observed entities compared to MOT20, as well as limitations in the dataset itself.

\begin{table}[ht]
    \centering
    \caption{Average precision (AP) and average recall (AR) for bounding box estimation of industrial entities}
    \begin{tabular}{l|| r r r r r r}
        \textbf{Metric} & $AP^{val}$    & $AR^{val}$    & $F1-Score$    & $AP_{.50}^{val}$  & $AP_{.75}^{val}$  & $AP_{.50:.95}^{val}$   \\ \hline
        Result & 0.80          &   0.83        &   0.815       &   0.92            &   0.87            &   0.83
    \end{tabular}
    
    \label{tab:ap_ar_results}
\end{table}

\begin{table}[ht]
    \centering
    \caption{Results on our validation data.}
    \begin{tabular}{l || r r r r}
        \textbf{Method}      &   \textbf{MOTA}    &   \textbf{IDF1}    &   \textbf{HOTA}    &   \textbf{IDs} \\
        \hline
        BYTE-TRACK  &   0.654        &   0.641    &   0.564    &   778  \\
        BoT-SORT    &   0.672        &   0.667    &   0.58    &   569  \\
        SiamMOT     &   0.575        &   0.594    &   0.503    &   928  \\
    \end{tabular}
    
    \label{tab:reid_val_results}
\end{table}

\section{Conclusion and Outlook}\label{sec5}
In this contribution, a novel framework and approach for the efficient computer vision based tracking of multiple industrial entities was presented.
Using a space of approximately $16$ x $8$ sqm in a warehousing environment, $52$ infrared cameras and six RGB cameras mounted on the ceiling and railings of this warehouse, a tracking space was defined.
In this space, six industrial entities, including small load carriers, pallets, barrels, cardboard boxes, forklifts, and a mesh box were tracked using reflective markers and tracking software using infrared tracking hardware.
With this tracking setup, the herein presented TOMIE dataset was recorded, including $112,860$ frames worth of RGB images and annotation files that contain  approximately $16$ min of recordings, after data synchronization and filtration.
The recordings were subdivided into distinct logistical scenarios, drawn from industrial applications (e.g., setting up pallets in lanes, to be loaded into trucks).
Three commonly used tracking algorithms, namely \textit{ByteTrack}, \textit{SiamMot}, and \textit{Bot-Sort}, were applied to the herein developed dataset, performing overall worse than on comparable state-of-the-art datasets.

While developing the recording setup, during the process of recording itself, and while evaluating the resulting data and its use,  additional limitations and challenges were encountered.

\subsection{Limitations and Challenges}
While setting up the camera network for recording, a major challenge arose while trying to mark the industrial entities in a way, in which they would be detectable and distinguishable for the infrared cameras.
As previously described, the marking tape needed to be distributed along the faces of the entities in such a unique way, that they would be distinguishable by virtue of the resulting point cloud.
When working with a limited amount of entities, that have large surface areas, this does typically not cause any trouble.
However, applying the same approach to a multitude of entities, especially smaller ones (e.g., the small load carriers in our dataset), causes the infrared cameras to yield suboptimal tracking results.

In addition, the proximity of the entities that ought to be tracked to one another further complicated the tracking process.
When the markers on the edges of one entity came too close to those of another, one or both entities tended to disappear in the tracking software, resulting in frames that provide users with no positional ground truth.
However, both the ground truth and the realistic positioning of the entities in a way that resembles industrial applications is of importance.

Furthermore, the software used in the herein presented tracking setup does not enable the tracking of human motion.
The operators in the recorded tracking scenarios were therefore not tracked and come with no labeled ground truth in our dataset.
The addition of such data might be of interest for researchers in the field of human activity recognition or person re-identification.

Once recorded, the data proved challenging to be interpreted for the purpose of frame-wise object detection, due to the use of multiple RGB cameras and the underlying ground truth being infrared camera based.
This is because the ground truth is calculated based on the markers on the given entity in combination with its 3D rendered model.
Using this set of data, no information is given on visual occlusion by other entities present in the recording.
This results in the creation of 2D bounding boxes as a ground truth that are accurate in free space but would result in poor IoU results, when used with common object detection algorithms, which would only detect the non-occluded parts of the entities.
In addition, when using more than one RGB camera, the notion of the term occlusion becomes even more complicated to deal with, as an entity that is occluded in one perspective might be entirely visible in another.
This results in bounding boxes being created for entities that are entirely occluded in some perspectives, which would lead to an IoU of $0$\%, if the data were to be put to a test.

Subsequently, once an industrial entity were to be detected, the interest would lie in the classification and identification of said entity.
While classification is in part feasible with the herein presented recording setup, the identification of specific entities, analogous to the work presented in \cite{Rutinowski}, would necessitate an altered sensor use.
More specifically, this would entail the use of cameras at a level close to the ground and closer to the recorded entities, as to capture their surface structure in more detail.
This however, might lead to further occlusions, due to camera positioning.

Looking back at the vision for a tracking system that was established in the beginning of this contribution, some limitations still persist.
One such limitation of the above mentioned occlusions, that do occur in industrial scenarios that are uncontrollable.
In addition, since this work was conducted in only a single recording environment, it is yet to be evaluated, whether the selected algorithms would perform similarly in another environment.

Finally, while handling the recorded data, synchronization problems occurred, in which the RGB and infrared frames were not overlapping as they should.
The reason for this has yet to be further explored.
Additionally, the volume of the data that is generated using this recording setup is not to be underestimated.
An efficient way of handling such large amounts of data is also of great importance, as to increase the efficiency and applicability of our recording approach.

\subsection{Follow-up Research}

Taking the limitations mentioned in the previous subsection and our results in general into account, we identified the following ways in which our contribution could be expanded upon:

The scenarios that were recorded could be expanded upon in terms of their diversity (i.e., different versions of the same scenarios or more scenarios to begin with) and their duration.
Furthermore, the complexity of the scenarios could be increased by including a greater amount of industrial entities and a greater amount of entity classes, including human operators.

The way in which the industrial entities are marked with reflective tape could be analyzed once more, creating a system that would allow for a more reliable marking of a larger amount of entities.
In doing so, reproducibility and result quality could be enhanced.

Finally, the tracking software that was used thus far could be replaced by a self-developed one, which could be tailored for a multi camera setup.
This tracking software might then be able to not only provide bounding boxes that would take occlusions into account but might also provide 3D bounding boxes, including information on the entity's orientation in space.
The use of depth information (e.g., by virtue of RGBD cameras) might be necessary to accomplish this task.

\backmatter

\section*{Acknowledgments}

This work is part of the project “Silicon Economy Logistics Ecosystem” which is funded by the German Federal Ministry of Transport and Digital Infrastructure.

This work is part of the research of the Lamarr Institute for Machine Learning and Artificial Intelligence which is funded by the German Ministry of Education and Research.

In our experiments, we adopt real-world scenarios from the warehousing sector. 
As to ensure the validity of the herein presented scenarios, we had our colleagues at the Fraunhofer Institute for Material Flow and Logistics in Dortmund, Germany, evaluate them.  We would specifically like to thank Jennifer Beuth, head of the department of warehousing logistics and IT planning at the Fraunhofer Institute for her support.

\bibliography{main.bib}


\begin{thebibliography}{53}
\ifx \bisbn   \undefined \def \bisbn  #1{ISBN #1}\fi
\ifx \binits  \undefined \def \binits#1{#1}\fi
\ifx \bauthor  \undefined \def \bauthor#1{#1}\fi
\ifx \batitle  \undefined \def \batitle#1{#1}\fi
\ifx \bjtitle  \undefined \def \bjtitle#1{#1}\fi
\ifx \bvolume  \undefined \def \bvolume#1{\textbf{#1}}\fi
\ifx \byear  \undefined \def \byear#1{#1}\fi
\ifx \bissue  \undefined \def \bissue#1{#1}\fi
\ifx \bfpage  \undefined \def \bfpage#1{#1}\fi
\ifx \blpage  \undefined \def \blpage #1{#1}\fi
\ifx \burl  \undefined \def \burl#1{\textsf{#1}}\fi
\ifx \doiurl  \undefined \def \doiurl#1{\url{https://doi.org/#1}}\fi
\ifx \betal  \undefined \def \betal{\textit{et al.}}\fi
\ifx \binstitute  \undefined \def \binstitute#1{#1}\fi
\ifx \binstitutionaled  \undefined \def \binstitutionaled#1{#1}\fi
\ifx \bctitle  \undefined \def \bctitle#1{#1}\fi
\ifx \beditor  \undefined \def \beditor#1{#1}\fi
\ifx \bpublisher  \undefined \def \bpublisher#1{#1}\fi
\ifx \bbtitle  \undefined \def \bbtitle#1{#1}\fi
\ifx \bedition  \undefined \def \bedition#1{#1}\fi
\ifx \bseriesno  \undefined \def \bseriesno#1{#1}\fi
\ifx \blocation  \undefined \def \blocation#1{#1}\fi
\ifx \bsertitle  \undefined \def \bsertitle#1{#1}\fi
\ifx \bsnm \undefined \def \bsnm#1{#1}\fi
\ifx \bsuffix \undefined \def \bsuffix#1{#1}\fi
\ifx \bparticle \undefined \def \bparticle#1{#1}\fi
\ifx \barticle \undefined \def \barticle#1{#1}\fi
\bibcommenthead
\ifx \bconfdate \undefined \def \bconfdate #1{#1}\fi
\ifx \botherref \undefined \def \botherref #1{#1}\fi
\ifx \url \undefined \def \url#1{\textsf{#1}}\fi
\ifx \bchapter \undefined \def \bchapter#1{#1}\fi
\ifx \bbook \undefined \def \bbook#1{#1}\fi
\ifx \bcomment \undefined \def \bcomment#1{#1}\fi
\ifx \oauthor \undefined \def \oauthor#1{#1}\fi
\ifx \citeauthoryear \undefined \def \citeauthoryear#1{#1}\fi
\ifx \endbibitem  \undefined \def \endbibitem {}\fi
\ifx \bconflocation  \undefined \def \bconflocation#1{#1}\fi
\ifx \arxivurl  \undefined \def \arxivurl#1{\textsf{#1}}\fi
\csname PreBibitemsHook\endcsname

\bibitem{franko_reliable_2020}
\begin{barticle}
\bauthor{\bsnm{Frankó}, \binits{A.}},
\bauthor{\bsnm{Vida}, \binits{G.}},
\bauthor{\bsnm{Varga}, \binits{P.}}:
\batitle{Reliable identification schemes for asset and production tracking in
  industry 4.0}.
\bjtitle{Sensors}
\bvolume{20},
\bfpage{3709}
(\byear{2020}).
\doiurl{10.3390/s20133709}
\end{barticle}
\endbibitem

\bibitem{anuj2017multiple}
\begin{bchapter}
\bauthor{\bsnm{Anuj}, \binits{L.}},
\bauthor{\bsnm{Krishna}, \binits{M.G.}}:
\bctitle{Multiple camera based multiple object tracking under occlusion: A
  survey}.
In: \bbtitle{International Conference on Innovative Mechanisms for Industry
  Applications (ICIMIA)},
pp. \bfpage{432}--\blpage{437}
(\byear{2017}).
\doiurl{10.1109/ICIMIA.2017.7975652}
\end{bchapter}
\endbibitem

\bibitem{liu_recent_2023}
\begin{barticle}
\bauthor{\bsnm{Liu}, \binits{W.}},
\bauthor{\bsnm{Bao}, \binits{Q.}},
\bauthor{\bsnm{Sun}, \binits{Y.}},
\bauthor{\bsnm{Mei}, \binits{T.}}:
\batitle{Recent advances of monocular {2D} and {3D} human pose estimation: {A}
  deep learning perspective}.
\bjtitle{ACM Computing Surveys}
\bvolume{55},
\bfpage{1}--\blpage{41}
(\byear{2023}).
\doiurl{10.1145/3524497}
\end{barticle}
\endbibitem

\bibitem{zhan_ray3d_2022}
\begin{bchapter}
\bauthor{\bsnm{Zhan}, \binits{Y.}},
\bauthor{\bsnm{Li}, \binits{F.}},
\bauthor{\bsnm{Weng}, \binits{R.}},
\bauthor{\bsnm{Choi}, \binits{W.}}:
\bctitle{{Ray3D}: ray-based {3D} human pose estimation for monocular absolute
  {3D} localization}.
In: \bbtitle{{Computer} {Vision} and {Pattern} {Recognition} ({CVPR})},
pp. \bfpage{13106}--\blpage{13115}
(\byear{2022}).
\doiurl{10.1109/CVPR52688.2022.01277}
\end{bchapter}
\endbibitem

\bibitem{wang_deep_2021}
\begin{barticle}
\bauthor{\bsnm{Wang}, \binits{J.}},
\bauthor{\bsnm{Tan}, \binits{S.}},
\bauthor{\bsnm{Zhen}, \binits{X.}},
\bauthor{\bsnm{Xu}, \binits{S.}},
\bauthor{\bsnm{Zheng}, \binits{F.}},
\bauthor{\bsnm{He}, \binits{Z.}},
\bauthor{\bsnm{Shao}, \binits{L.}}:
\batitle{Deep {3D} human pose estimation: {A} review}.
\bjtitle{Computer Vision and Image Understanding}
\bvolume{210},
\bfpage{103225}
(\byear{2021}).
\doiurl{10.1016/j.cviu.2021.103225}
\end{barticle}
\endbibitem

\bibitem{Ciaparrone_2019}
\begin{barticle}
\bauthor{\bsnm{Ciaparrone}, \binits{G.}},
\bauthor{\bsnm{Luque~S\'{a}nchez}, \binits{F.}},
\bauthor{\bsnm{Tabik}, \binits{S.}},
\bauthor{\bsnm{Troiano}, \binits{L.}},
\bauthor{\bsnm{Tagliaferri}, \binits{R.}},
\bauthor{\bsnm{Herrera}, \binits{F.}}:
\batitle{Deep learning in video multi-object tracking: A survey}.
\bjtitle{Neurocomputing}
\bvolume{381},
\bfpage{61}--\blpage{88}
(\byear{2020}).
\doiurl{10.1016/j.neucom.2019.11.023}
\end{barticle}
\endbibitem

\bibitem{dendorfer2021motchallenge}
\begin{barticle}
\bauthor{\bsnm{Dendorfer}, \binits{P.}},
\bauthor{\bsnm{Osep}, \binits{A.}},
\bauthor{\bsnm{Milan}, \binits{A.}},
\bauthor{\bsnm{Schindler}, \binits{K.}},
\bauthor{\bsnm{Cremers}, \binits{D.}},
\bauthor{\bsnm{Reid}, \binits{I.}},
\bauthor{\bsnm{Roth}, \binits{S.}},
\bauthor{\bsnm{Leal-Taix{\'e}}, \binits{L.}}:
\batitle{Motchalllenge: A benchmark for single-camera multiple target
  tracking}.
\bjtitle{International Journal of Computer Vision}
\bvolume{129},
\bfpage{845}--\blpage{881}
(\byear{2021}).
\doiurl{10.1007/s11263-020-01393-0}
\end{barticle}
\endbibitem

\bibitem{yu_poi_2016}
\begin{bchapter}
\bauthor{\bsnm{Yu}, \binits{F.}},
\bauthor{\bsnm{Li}, \binits{W.}},
\bauthor{\bsnm{Li}, \binits{Q.}},
\bauthor{\bsnm{Liu}, \binits{Y.}},
\bauthor{\bsnm{Shi}, \binits{X.}},
\bauthor{\bsnm{Yan}, \binits{J.}}:
\bctitle{{POI}: {M}ultiple object tracking with high performance detection and
  appearance feature}.
In: \bbtitle{European Conference on Computer Vision (ECCV) Workshops},
pp. \bfpage{36}--\blpage{42}
(\byear{2016}).
\doiurl{10.1007/978-3-319-48881-3_3}
\end{bchapter}
\endbibitem

\bibitem{Hilke_2018}
\begin{bchapter}
\bauthor{\bsnm{Kieritz}, \binits{H.}},
\bauthor{\bsnm{Hübner}, \binits{W.}},
\bauthor{\bsnm{Arens}, \binits{M.}}:
\bctitle{Joint detection and online multi-object tracking}.
In: \bbtitle{Computer Vision and Pattern Recognition Workshops (CVPRW)},
pp. \bfpage{1540}--\blpage{15408}
(\byear{2018}).
\doiurl{10.1109/CVPRW.2018.00195}
\end{bchapter}
\endbibitem

\bibitem{Zhao2018MultiObjectTW}
\begin{botherref}
\oauthor{\bsnm{Zhao}, \binits{D.}},
\oauthor{\bsnm{Fu}, \binits{H.}},
\oauthor{\bsnm{Xiao}, \binits{L.}},
\oauthor{\bsnm{Wu}, \binits{T.}},
\oauthor{\bsnm{Dai}, \binits{B.}}:
Multi-object tracking with correlation filter for autonomous vehicle.
Sensors
\textbf{18}
(2018).
\doiurl{10.3390/s18072004}
\end{botherref}
\endbibitem

\bibitem{GoogLeNet_2015}
\begin{bchapter}
\bauthor{\bsnm{Szegedy}, \binits{C.}},
\bauthor{\bsnm{Liu}, \binits{W.}},
\bauthor{\bsnm{Jia}, \binits{Y.}},
\bauthor{\bsnm{Sermanet}, \binits{P.}},
\bauthor{\bsnm{Reed}, \binits{S.}},
\bauthor{\bsnm{Anguelov}, \binits{D.}},
\bauthor{\bsnm{Erhan}, \binits{D.}},
\bauthor{\bsnm{Vanhoucke}, \binits{V.}},
\bauthor{\bsnm{Rabinovich}, \binits{A.}}:
\bctitle{Going deeper with convolutions}.
In: \bbtitle{Computer Vision and Pattern Recognition (CVPR)},
pp. \bfpage{1}--\blpage{9}
(\byear{2015}).
\doiurl{10.1109/CVPR.2015.7298594}
\end{bchapter}
\endbibitem

\bibitem{MOT15}
\begin{botherref}
\oauthor{\bsnm{Leal-Taixé}, \binits{L.}},
\oauthor{\bsnm{Milan}, \binits{A.}},
\oauthor{\bsnm{Reid}, \binits{I.}},
\oauthor{\bsnm{Roth}, \binits{S.}}:
Motchallenge 2015: Towards a benchmark for multi-target tracking
(2015).
\doiurl{10.48550/arXiv.1504.01942}
\end{botherref}
\endbibitem

\bibitem{MOT16}
\begin{botherref}
\oauthor{\bsnm{Milan}, \binits{A.}},
\oauthor{\bsnm{Leal-Taixe}, \binits{L.}},
\oauthor{\bsnm{Reid}, \binits{I.}},
\oauthor{\bsnm{Roth}, \binits{S.}},
\oauthor{\bsnm{Schindler}, \binits{K.}}:
MOT16: A Benchmark for Multi-Object Tracking.
arXiv
(2016).
\doiurl{10.48550/arXiv.1603.00831}
\end{botherref}
\endbibitem

\bibitem{Tang2017MultiplePT}
\begin{botherref}
\oauthor{\bsnm{Tang}, \binits{S.}},
\oauthor{\bsnm{Andriluka}, \binits{M.}},
\oauthor{\bsnm{Andres}, \binits{B.}},
\oauthor{\bsnm{Schiele}, \binits{B.}}:
Multiple people tracking by lifted multicut and person re-identification.
Conference on Computer Vision and Pattern Recognition (CVPR),
3701--3710
(2017).
\doiurl{10.1109/CVPR.2017.394}
\end{botherref}
\endbibitem

\bibitem{Chen2017OnlineMT}
\begin{botherref}
\oauthor{\bsnm{Chen}, \binits{L.}},
\oauthor{\bsnm{Ai}, \binits{H.}},
\oauthor{\bsnm{Shang}, \binits{C.}},
\oauthor{\bsnm{Zhuang}, \binits{Z.}},
\oauthor{\bsnm{Bai}, \binits{B.}}:
Online multi-object tracking with convolutional neural networks.
International Conference on Image Processing (ICIP),
645--649
(2017).
\doiurl{10.1109/ICIP.2017.8296360}
\end{botherref}
\endbibitem

\bibitem{Ma2018CustomizedMT}
\begin{bchapter}
\bauthor{\bsnm{Ma}, \binits{L.}},
\bauthor{\bsnm{Tang}, \binits{S.}},
\bauthor{\bsnm{Black}, \binits{M.J.}},
\bauthor{\bsnm{Gool}, \binits{L.V.}}:
\bctitle{Customized multi-person tracker}.
In: \bbtitle{Asian Conference on Computer Vision (ACCV)}
(\byear{2018})
\end{bchapter}
\endbibitem

\bibitem{RNN_assoc_2018}
\begin{bchapter}
\bauthor{\bsnm{Ma}, \binits{C.}},
\bauthor{\bsnm{Yang}, \binits{C.}},
\bauthor{\bsnm{Yang}, \binits{F.}},
\bauthor{\bsnm{Zhuang}, \binits{Y.}},
\bauthor{\bsnm{Zhang}, \binits{Z.}},
\bauthor{\bsnm{Jia}, \binits{H.}},
\bauthor{\bsnm{Xie}, \binits{X.}}:
\bctitle{Trajectory factory: Tracklet cleaving and re-connection by deep
  siamese bi-gru for multiple object tracking}.
In: \bbtitle{International Conference on Multimedia and Expo (ICME)},
pp. \bfpage{1}--\blpage{6}
(\byear{2018}).
\doiurl{10.1109/ICME.2018.8486454}
\end{bchapter}
\endbibitem

\bibitem{Ren_2018}
\begin{bchapter}
\bauthor{\bsnm{Ren}, \binits{L.}},
\bauthor{\bsnm{Lu}, \binits{J.}},
\bauthor{\bsnm{Wang}, \binits{Z.}},
\bauthor{\bsnm{Tian}, \binits{Q.}},
\bauthor{\bsnm{Zhou}, \binits{J.}}:
\bctitle{Collaborative deep reinforcement learning for multi-object tracking}.
In: \bbtitle{European Conference on Computer Vision (ECCV)},
pp. \bfpage{605}--\blpage{621}
(\byear{2018}).
\doiurl{10.1007/978-3-030-01219-9_36}
\end{bchapter}
\endbibitem

\bibitem{SORT_2016}
\begin{bchapter}
\bauthor{\bsnm{Bewley}, \binits{A.}},
\bauthor{\bsnm{Ge}, \binits{Z.}},
\bauthor{\bsnm{Ott}, \binits{L.}},
\bauthor{\bsnm{Ramos}, \binits{F.}},
\bauthor{\bsnm{Upcroft}, \binits{B.}}:
\bctitle{Simple online and realtime tracking}.
In: \bbtitle{International Conference on Image Processing (ICIP)},
pp. \bfpage{3464}--\blpage{3468}
(\byear{2016}).
\doiurl{10.1109/ICIP.2016.7533003}
\end{bchapter}
\endbibitem

\bibitem{Aharon22}
\begin{botherref}
\oauthor{\bsnm{Aharon}, \binits{N.}},
\oauthor{\bsnm{Orfaig}, \binits{R.}},
\oauthor{\bsnm{Bobrovsky}, \binits{B.-Z.}}:
BoT-SORT: Robust Associations Multi-Pedestrian Tracking.
arXiv
(2022).
\doiurl{10.48550/ARXIV.2206.14651}
\end{botherref}
\endbibitem

\bibitem{zhang_bytetrack_2022}
\begin{botherref}
\oauthor{\bsnm{Zhang}, \binits{Y.}},
\oauthor{\bsnm{Sun}, \binits{P.}},
\oauthor{\bsnm{Jiang}, \binits{Y.}},
\oauthor{\bsnm{Yu}, \binits{D.}},
\oauthor{\bsnm{Weng}, \binits{F.}},
\oauthor{\bsnm{Yuan}, \binits{Z.}},
\oauthor{\bsnm{Luo}, \binits{P.}},
\oauthor{\bsnm{Liu}, \binits{W.}},
\oauthor{\bsnm{Wang}, \binits{X.}}:
Bytetrack: Multi-object tracking by associating every detection box
(2022).
\doiurl{10.1007/978-3-031-20047-2_1}
\end{botherref}
\endbibitem

\bibitem{Yoon2015BayesianMT}
\begin{botherref}
\oauthor{\bsnm{Yoon}, \binits{J.H.}},
\oauthor{\bsnm{Yang}, \binits{M.-H.}},
\oauthor{\bsnm{Lim}, \binits{J.}},
\oauthor{\bsnm{Yoon}, \binits{K.-j.}}:
Bayesian multi-object tracking using motion context from multiple objects.
Winter Conference on Applications of Computer Vision (WACV),
33--40
(2015).
\doiurl{10.1109/WACV.2015.12}
\end{botherref}
\endbibitem

\bibitem{Tiwari2017ARO}
\begin{barticle}
\bauthor{\bsnm{Tiwari}, \binits{M.}},
\bauthor{\bsnm{Singhai}, \binits{R.}}:
\batitle{A review of detection and tracking of object from image and video
  sequences}.
\bjtitle{International Journal of Computational Intelligence Research}
\bvolume{13},
\bfpage{745}--\blpage{765}
(\byear{2017})
\end{barticle}
\endbibitem

\bibitem{Analysis_DL_2021}
\begin{botherref}
\oauthor{\bsnm{Kalake}, \binits{L.}},
\oauthor{\bsnm{Wan}, \binits{W.}},
\oauthor{\bsnm{Hou}, \binits{L.}}:
Analysis based on recent deep learning approaches applied in real-time
  multi-object tracking: A review
\textbf{9},
32650--32671
(2021).
\doiurl{10.1109/ACCESS.2021.3060821}
\end{botherref}
\endbibitem

\bibitem{Bredereck_2012}
\begin{botherref}
\oauthor{\bsnm{Bredereck}, \binits{M.}},
\oauthor{\bsnm{Jiang}, \binits{X.}},
\oauthor{\bsnm{K{\"o}rner}, \binits{M.}},
\oauthor{\bsnm{Denzler}, \binits{J.}}:
Data association for multi-object tracking-by-detection in multi-camera
  networks.
International Conference on Distributed Smart Cameras (ICDSC),
1--6
(2012)
\end{botherref}
\endbibitem

\bibitem{Wang2013IntelligentMV}
\begin{barticle}
\bauthor{\bsnm{Wang}, \binits{X.}}:
\batitle{Intelligent multi-camera video surveillance: A review}.
\bjtitle{Pattern Recognition Letters}
\bvolume{34},
\bfpage{3}--\blpage{19}
(\byear{2013}).
\doiurl{10.1016/j.patrec.2012.07.005}
\end{barticle}
\endbibitem

\bibitem{zhang2015camera}
\begin{bchapter}
\bauthor{\bsnm{Zhang}, \binits{S.}},
\bauthor{\bsnm{Staudt}, \binits{E.}},
\bauthor{\bsnm{Faltemier}, \binits{T.}},
\bauthor{\bsnm{Roy-Chowdhury}, \binits{A.K.}}:
\bctitle{A camera network tracking ({CamNeT}) dataset and performance
  baseline}.
In: \bbtitle{Winter Conference on Applications of Computer Vision},
pp. \bfpage{365}--\blpage{372}
(\byear{2015}).
\doiurl{10.1109/WACV.2015.55}
\end{bchapter}
\endbibitem

\bibitem{An_Occlussion_2021}
\begin{bchapter}
\bauthor{\bsnm{Specker}, \binits{A.}},
\bauthor{\bsnm{Stadler}, \binits{D.}},
\bauthor{\bsnm{Florin}, \binits{L.}},
\bauthor{\bsnm{Beyerer}, \binits{J.}}:
\bctitle{An occlusion-aware multi-target multi-camera tracking system}.
In: \bbtitle{Computer Vision and Pattern Recognition Workshops (CVPRW)},
pp. \bfpage{4168}--\blpage{4177}
(\byear{2021}).
\doiurl{10.1109/CVPRW53098.2021.00471}
\end{bchapter}
\endbibitem

\bibitem{Improving_Multi_Cam_2022}
\begin{bchapter}
\bauthor{\bsnm{Specker}, \binits{A.}},
\bauthor{\bsnm{Florin}, \binits{L.}},
\bauthor{\bsnm{Cormier}, \binits{M.}},
\bauthor{\bsnm{Beyerer}, \binits{J.}}:
\bctitle{Improving multi-target multi-camera tracking by track refinement and
  completion}.
In: \bbtitle{Computer Vision and Pattern Recognition Workshops (CVPRW)},
pp. \bfpage{3198}--\blpage{3208}
(\byear{2022}).
\doiurl{10.1109/CVPRW56347.2022.00361}
\end{bchapter}
\endbibitem

\bibitem{liu2021city}
\begin{bchapter}
\bauthor{\bsnm{Liu}, \binits{C.}},
\bauthor{\bsnm{Zhang}, \binits{Y.}},
\bauthor{\bsnm{Luo}, \binits{H.}},
\bauthor{\bsnm{Tang}, \binits{J.}},
\bauthor{\bsnm{Chen}, \binits{W.}},
\bauthor{\bsnm{Xu}, \binits{X.}},
\bauthor{\bsnm{Wang}, \binits{F.}},
\bauthor{\bsnm{Li}, \binits{H.}},
\bauthor{\bsnm{Shen}, \binits{Y.-D.}}:
\bctitle{{City-Scale} multi-camera vehicle tracking guided by crossroad zones}.
In: \bbtitle{Computer Vision and Pattern Recognition Workshops (CVPRW)},
vol. \bseriesno{3},
pp. \bfpage{4124}--\blpage{4132}
(\byear{2021}).
\doiurl{10.1109/CVPRW53098.2021.00466}
\end{bchapter}
\endbibitem

\bibitem{he2020multi}
\begin{bchapter}
\bauthor{\bsnm{He}, \binits{S.}},
\bauthor{\bsnm{Luo}, \binits{H.}},
\bauthor{\bsnm{Chen}, \binits{W.}},
\bauthor{\bsnm{Zhang}, \binits{M.}},
\bauthor{\bsnm{Zhang}, \binits{Y.}},
\bauthor{\bsnm{Wang}, \binits{F.}},
\bauthor{\bsnm{Li}, \binits{H.}},
\bauthor{\bsnm{Jiang}, \binits{W.}}:
\bctitle{Multi-domain learning and identity mining for vehicle
  re-identification}.
In: \bbtitle{Computer Vision and Pattern Recognition Workshops (CVPRW)},
pp. \bfpage{582}--\blpage{583}
(\byear{2020}).
\doiurl{10.1109/CVPRW50498.2020.00299}
\end{bchapter}
\endbibitem

\bibitem{hsu2019multi}
\begin{bchapter}
\bauthor{\bsnm{Hsu}, \binits{H.-M.}},
\bauthor{\bsnm{Huang}, \binits{T.-W.}},
\bauthor{\bsnm{Wang}, \binits{G.}},
\bauthor{\bsnm{Cai}, \binits{J.}},
\bauthor{\bsnm{Lei}, \binits{Z.}},
\bauthor{\bsnm{Hwang}, \binits{J.-N.}}:
\bctitle{Multi-camera tracking of vehicles based on deep features {Re-ID} and
  trajectory-based camera link models}.
In: \bbtitle{Computer Vision and Pattern Recognition Workshops (CVPRW)},
pp. \bfpage{416}--\blpage{424}
(\byear{2019})
\end{bchapter}
\endbibitem

\bibitem{kohl2020mta}
\begin{bchapter}
\bauthor{\bsnm{Kohl}, \binits{P.}},
\bauthor{\bsnm{Specker}, \binits{A.}},
\bauthor{\bsnm{Schumann}, \binits{A.}},
\bauthor{\bsnm{Beyerer}, \binits{J.}}:
\bctitle{The {MTA} dataset for multi-target multi-camera pedestrian tracking by
  weighted distance aggregation}.
In: \bbtitle{Computer Vision and Pattern Recognition Workshops (CVPRW)},
pp. \bfpage{1042}--\blpage{1043}
(\byear{2020}).
\doiurl{10.1109/CVPRW50498.2020.00529}
\end{bchapter}
\endbibitem

\bibitem{mayershofer2020loco}
\begin{bchapter}
\bauthor{\bsnm{Mayershofer}, \binits{C.}},
\bauthor{\bsnm{Holm}, \binits{D.-M.}},
\bauthor{\bsnm{Molter}, \binits{B.}},
\bauthor{\bsnm{Fottner}, \binits{J.}}:
\bctitle{Loco: Logistics objects in context}.
In: \bbtitle{International Conference on Machine Learning and Applications
  (ICMLA)},
pp. \bfpage{612}--\blpage{617}
(\byear{2020}).
\doiurl{10.1109/ICMLA51294.2020.00102}
\end{bchapter}
\endbibitem

\bibitem{MOT20}
\begin{botherref}
\oauthor{\bsnm{Dendorfer}, \binits{P.}},
\oauthor{\bsnm{Rezatofighi}, \binits{H.}},
\oauthor{\bsnm{Milan}, \binits{A.}},
\oauthor{\bsnm{Shi}, \binits{J.Q.}},
\oauthor{\bsnm{Cremers}, \binits{D.}},
\oauthor{\bsnm{Reid}, \binits{I.D.}},
\oauthor{\bsnm{Roth}, \binits{S.}},
\oauthor{\bsnm{Schindler}, \binits{K.}},
\oauthor{\bsnm{Leal-Taix'e}, \binits{L.}}:
Mot20: A benchmark for multi object tracking in crowded scenes
(2020).
\doiurl{10.48550/ARXIV.2003.09003}
\end{botherref}
\endbibitem

\bibitem{geiger2013vision}
\begin{barticle}
\bauthor{\bsnm{Geiger}, \binits{A.}},
\bauthor{\bsnm{Lenz}, \binits{P.}},
\bauthor{\bsnm{Stiller}, \binits{C.}},
\bauthor{\bsnm{Urtasun}, \binits{R.}}:
\batitle{Vision meets robotics: The kitti dataset}.
\bjtitle{The International Journal of Robotics Research}
\bvolume{32},
\bfpage{1231}--\blpage{1237}
(\byear{2013}).
\doiurl{10.1177/0278364913491297}
\end{barticle}
\endbibitem

\bibitem{lin_microsoft_2014}
\begin{bchapter}
\bauthor{\bsnm{Lin}, \binits{T.-Y.}},
\bauthor{\bsnm{Maire}, \binits{M.}},
\bauthor{\bsnm{Belongie}, \binits{S.}},
\bauthor{\bsnm{Hays}, \binits{J.}},
\bauthor{\bsnm{Perona}, \binits{P.}},
\bauthor{\bsnm{Ramanan}, \binits{D.}},
\bauthor{\bsnm{Dollár}, \binits{P.}},
\bauthor{\bsnm{Zitnick}, \binits{C.L.}}:
\bctitle{Microsoft {COCO}: {Common} {Objects} in {Context}}.
In: \bbtitle{European Conference on Computer Vision (ECCV)},
pp. \bfpage{740}--\blpage{755}
(\byear{2014})
\end{bchapter}
\endbibitem

\bibitem{Bernardin08}
\begin{barticle}
\bauthor{\bsnm{Bernardin}, \binits{K.}},
\bauthor{\bsnm{Stiefelhagen}, \binits{R.}}:
\batitle{Evaluating multiple object tracking performance: the {CLEAR} {MOT}
  metrics}.
\bjtitle{EURASIP Journal on Image and Video Processing}
(\byear{2008}).
\doiurl{10.1155/2008/246309}
\end{barticle}
\endbibitem

\bibitem{Luiten20}
\begin{botherref}
\oauthor{\bsnm{Luiten}, \binits{J.}},
\oauthor{\bsnm{Osep}, \binits{A.}},
\oauthor{\bsnm{Dendorfer}, \binits{P.}},
\oauthor{\bsnm{Torr}, \binits{P.H.S.}},
\oauthor{\bsnm{Geiger}, \binits{A.}},
\oauthor{\bsnm{Leal{-}Taix{\'{e}}}, \binits{L.}},
\oauthor{\bsnm{Leibe}, \binits{B.}}:
{HOTA:} {A} higher order metric for evaluating multi-object tracking.
International Journal of Computer Vision
\textbf{129}
(2020).
\doiurl{10.1007/s11263-020-01375-2}
\end{botherref}
\endbibitem

\bibitem{Ristani16}
\begin{bchapter}
\bauthor{\bsnm{Ristani}, \binits{E.}},
\bauthor{\bsnm{Solera}, \binits{F.}},
\bauthor{\bsnm{Zou}, \binits{R.}},
\bauthor{\bsnm{Cucchiara}, \binits{R.}},
\bauthor{\bsnm{Tomasi}, \binits{C.}}:
\bctitle{Performance measures and a data set for multi-target, multi-camera
  tracking}.
In: \bbtitle{European Conference on Computer Vision (ECCV) Workshops},
pp. \bfpage{17}--\blpage{35}
(\byear{2016}).
\doiurl{10.1007/978-3-319-48881-3_2}
\end{bchapter}
\endbibitem

\bibitem{drost2017introducing}
\begin{bchapter}
\bauthor{\bsnm{Drost}, \binits{B.}},
\bauthor{\bsnm{Ulrich}, \binits{M.}},
\bauthor{\bsnm{Bergmann}, \binits{P.}},
\bauthor{\bsnm{Hartinger}, \binits{P.}},
\bauthor{\bsnm{Steger}, \binits{C.}}:
\bctitle{Introducing {MVT}ec {ITODD}-a dataset for {3D} object recognition in
  industry}.
In: \bbtitle{Iinternational Conference on Computer Vision Workshops (ICCVW)},
pp. \bfpage{2200}--\blpage{2208}
(\byear{2017}).
\doiurl{10.1109/ICCVW.2017.257}
\end{bchapter}
\endbibitem

\bibitem{luo2019benchmark}
\begin{barticle}
\bauthor{\bsnm{Luo}, \binits{C.}},
\bauthor{\bsnm{Yu}, \binits{L.}},
\bauthor{\bsnm{Yang}, \binits{E.}},
\bauthor{\bsnm{Zhou}, \binits{H.}},
\bauthor{\bsnm{Ren}, \binits{P.}}:
\batitle{A benchmark image dataset for industrial tools}.
\bjtitle{Pattern Recognition Letters}
\bvolume{125},
\bfpage{341}--\blpage{348}
(\byear{2019}).
\doiurl{10.1016/j.patrec.2019.05.011}
\end{barticle}
\endbibitem

\bibitem{de2022dataset}
\begin{botherref}
\oauthor{\bsnm{De~Roovere}, \binits{P.}},
\oauthor{\bsnm{Moonen}, \binits{S.}},
\oauthor{\bsnm{Michiels}, \binits{N.}}, et al.:
Dataset of industrial metal objects
(2022).
\doiurl{10.48550/ARXIV.2208.04052}
\end{botherref}
\endbibitem

\bibitem{abou2022synthetic}
\begin{bchapter}
\bauthor{\bsnm{Abou~Akar}, \binits{C.}},
\bauthor{\bsnm{Tekli}, \binits{J.}},
\bauthor{\bsnm{Jess}, \binits{D.}},
\bauthor{\bsnm{Khoury}, \binits{M.}},
\bauthor{\bsnm{Kamradt}, \binits{M.}},
\bauthor{\bsnm{Guthe}, \binits{M.}}:
\bctitle{Synthetic object recognition dataset for industries}.
In: \bbtitle{SIBGRAPI Conference on Graphics, Patterns and Images (SIBGRAPI)},
vol. \bseriesno{1},
pp. \bfpage{150}--\blpage{155}
(\byear{2022}).
\doiurl{10.1109/SIBGRAPI55357.2022.9991784}
\end{bchapter}
\endbibitem

\bibitem{s20154083}
\begin{botherref}
\oauthor{\bsnm{Niemann}, \binits{F.}},
\oauthor{\bsnm{Reining}, \binits{C.}},
\oauthor{\bsnm{Moya~Rueda}, \binits{F.}},
\oauthor{\bsnm{Nair}, \binits{N.R.}},
\oauthor{\bsnm{Steffens}, \binits{J.A.}},
\oauthor{\bsnm{Fink}, \binits{G.A.}},
\oauthor{\bparticle{ten} \bsnm{Hompel}, \binits{M.}}:
{LARa}: Creating a dataset for human activity recognition in logistics using
  semantic attributes.
Sensors
\textbf{20}(15)
(2020).
\doiurl{10.3390/s20154083}
\end{botherref}
\endbibitem

\bibitem{Rutinowski}
\begin{bchapter}
\bauthor{\bsnm{Rutinowski}, \binits{J.}},
\bauthor{\bsnm{Chilla}, \binits{T.}},
\bauthor{\bsnm{Pionzewski}, \binits{C.}},
\bauthor{\bsnm{Reining}, \binits{C.}},
\bauthor{\bparticle{ten} \bsnm{Hompel}, \binits{M.}}:
\bctitle{Towards re-identification for warehousing entities -- a
  work-in-progress study}.
In: \bbtitle{Emerging Technologies in Factory Automation (ETFA)},
pp. \bfpage{501}--\blpage{504}
(\byear{2021}).
\doiurl{10.1109/ETFA45728.2021.9613250}
\end{bchapter}
\endbibitem

\bibitem{Rutinowskia}
\begin{bchapter}
\bauthor{\bsnm{Rutinowski}, \binits{J.}},
\bauthor{\bsnm{Pionzewski}, \binits{C.}},
\bauthor{\bsnm{Chilla}, \binits{T.}},
\bauthor{\bsnm{Reining}, \binits{C.}},
\bauthor{\bparticle{ten} \bsnm{Hompel}, \binits{M.}}:
\bctitle{{Deep Learning Based Re-identification of Wooden Euro-pallets}}.
In: \bbtitle{International Conference on Machine Learning and Applications
  (ICMLA)}
(\byear{2022})
\end{bchapter}
\endbibitem

\bibitem{DIN_55405}
\begin{botherref}
\oauthor{\bsnm{DIN}}:
DIN 55405:2014-12, Packaging - Terminology - Terms and definitions
(2014)
\end{botherref}
\endbibitem

\bibitem{DIN_EN_13698-1}
\begin{botherref}
\oauthor{\bsnm{DIN}}:
DIN EN 13698-1:2004-01, Pallet production specification - Part 1: Construction
  specification for 800 mm × 1200 mm flat wooden pallets
(2004)
\end{botherref}
\endbibitem

\bibitem{luke_campagnola_2022_5974509}
\begin{botherref}
\oauthor{\bsnm{Campagnola}, \binits{L.}},
\oauthor{\bsnm{Larson}, \binits{E.}},
\oauthor{\bsnm{Klein}, \binits{A.}},
\oauthor{\bsnm{Hoese}, \binits{D.}},
\oauthor{\bsnm{Siddharth}},
\oauthor{\bsnm{Rossant}, \binits{C.}},
\oauthor{\bsnm{Griffiths}, \binits{A.}},
\oauthor{\bsnm{Rougier}, \binits{N.P.}},
\oauthor{\bparticle{van} \bsnm{Dijk}, \binits{L.}},
\oauthor{\bsnm{Mühlbauer}, \binits{K.}},
\oauthor{\bparticle{et} \bsnm{al.}}:
vispy/vispy: Version 0.9.5.
Zenodo
(2022).
\doiurl{10.5281/zenodo.5974509}
\end{botherref}
\endbibitem

\bibitem{Adhikari18}
\begin{barticle}
\bauthor{\bsnm{Adhikari}, \binits{B.}},
\bauthor{\bsnm{Peltom{\"{a}}ki}, \binits{J.}},
\bauthor{\bsnm{Puura}, \binits{J.}},
\bauthor{\bsnm{Huttunen}, \binits{H.}}:
\batitle{Faster bounding box annotation for object detection in indoor scenes}.
\bjtitle{European Workshop on Visual Information Processing (EUVIP)}
(\byear{2018}).
\doiurl{10.1109/EUVIP.2018.8611732}
\end{barticle}
\endbibitem

\bibitem{Siamese_Track_2020}
\begin{botherref}
\oauthor{\bsnm{Shuai}, \binits{B.}},
\oauthor{\bsnm{Berneshawi}, \binits{A.G.}},
\oauthor{\bsnm{Modolo}, \binits{D.}},
\oauthor{\bsnm{Tighe}, \binits{J.}}:
Multi-object tracking with siamese {Track-RCNN}
(2020).
\doiurl{10.48550/ARXIV.2004.07786}
\end{botherref}
\endbibitem

\bibitem{Megvii21}
\begin{botherref}
\oauthor{\bsnm{Zhang}, \binits{S.}},
\oauthor{\bsnm{Wang}, \binits{F.}},
\oauthor{\bsnm{Songtao}, \binits{L.}},
\oauthor{\bsnm{Zheng}, \binits{G.}}:
{YOLOX}: Exceeding yolo series in 2021
(2021).
\doiurl{10.48550/arXiv.2107.08430}
\end{botherref}
\endbibitem

\end{thebibliography}

\end{document}